\pdfoutput=1
\documentclass[11pt]{article}

\usepackage[margin=0.85in]{geometry}
\usepackage{times}
\usepackage[mono,scaled=0.9]{inconsolata}
\usepackage[numbers,sort&compress]{natbib}
\usepackage{authblk}
\usepackage{soul}
\usepackage{url}
\usepackage[colorlinks=true,
            linkcolor=blue,
            citecolor=blue,
            urlcolor=blue]{hyperref}
\usepackage[utf8]{inputenc}
\usepackage[small]{caption}
\usepackage{subcaption}
\usepackage{graphicx}
\usepackage{needspace}

\usepackage{amsmath}
\usepackage{amsthm}
\usepackage{booktabs}
\usepackage{bbm}
\makeatletter
    \DeclareRobustCommand*{\escapeus}[1]{%
        \begingroup\@activeus\scantokens{#1\endinput}\endgroup}
    \begingroup\lccode`\~=`\_\relax
        \lowercase{\endgroup\def\@activeus{\catcode`\_=\active \let~\_}}
\makeatother
\usepackage[linesnumbered,ruled,vlined]{algorithm2e}
\SetFuncSty{textbf}
\usepackage{xcolor}
\definecolor{halfgray}{gray}{0.55}
\usepackage{makecell}
\renewcommand{\arraystretch}{1.5}
\usepackage{etoolbox}
\makeatletter
\newcommand*\NoIndentAfterEnv[1]{%
  \AfterEndEnvironment{#1}{\par\@afterindentfalse\@afterheading}}
\makeatother

\usepackage[most]{tcolorbox}
\usepackage{amssymb}
\usepackage{mathtools}
\usepackage{amsfonts}
\usepackage{stmaryrd}
\usepackage{complexity}
\usepackage{csquotes}
\usepackage{multirow}
\usepackage[capitalise]{cleveref}
\crefname{appendix}{}{}
\usetikzlibrary{arrows.meta,positioning,decorations.pathreplacing}
\usepackage{threeparttable}
\usepackage{tabularx,array}

\tcbset{colback=white}
\newtcbtheorem[auto counter]{problemwithtitle}{Problem}{code={\edef\@currentlabelname{#2}},top=-0.5pt,bottom=-0.5pt,left=0pt,right=0pt}{prob}
\newtcbtheorem[auto counter]{tcbproblem}{Problem}{
    top=-15pt,bottom=-5pt,
    fontupper=\itshape,
    notitle,
    enhanced,
    frame hidden, 
    breakable,
    opacityfill=0,
}{prob}
\crefname{tcb@cnt@problemwithtitle}{Problem}{Problems}

\tcbuselibrary{listings, skins, breakable}
\definecolor{figurenavy}{HTML}{173B57}
\definecolor{codeblue}{RGB}{46,78,110}
\definecolor{codeframe}{gray}{0.78}
\lstdefinelanguage{cofola}{
    sensitive=true,
    morekeywords={
        set, bag, choose, choose_replace, tuple, sequence,
        choose_tuple, choose_sequence, choose_replace_tuple,
        choose_replace_sequence, partition, compose,
        supp, count, dedup_count, for, in, not, and, or,
        subset, disjoint, together, next_to
    },
    morecomment=[l]{\#},
    morestring=[b]",
    morestring=[b]'
}
\lstdefinestyle{verbatim}{
    basicstyle=\ttfamily,
    literate=
    *{\\}{{\textbackslash{}}}{1},
    keepspaces,
}
\lstdefinestyle{paper-code}{
    basicstyle=\ttfamily\small\mdseries\upshape\color{black},
    columns=fullflexible,
    keywordstyle=\mdseries\color{codeblue},
    stringstyle=\mdseries\upshape\color{black},
    commentstyle=\mdseries\upshape\color{black!60},
    numbers=none,
    numberstyle=\tiny\color{halfgray},
    firstnumber=1,
    breaklines=true,
    showspaces=false,
    showstringspaces=false,
    tabsize=1
}
\lstdefinestyle{inline-code}{
    language={},
    basicstyle=\ttfamily\small\mdseries\upshape\color{black},
    columns=fullflexible,
    keywordstyle=\color{black},
    stringstyle=\color{black},
    commentstyle=\color{black},
    identifierstyle=\color{black},
    emphstyle=\color{black},
    mathescape=false
}
\newtcblisting[auto counter]{cofolacode}[1]{%
    enhanced,
    frame hidden, 
    breakable,
    borderline west = {0.4pt}{0pt}{codeframe},
    opacityfill=0,
    top=-5pt,bottom=-5pt,
    fonttitle=\bfseries,
    listing only,
    listing options={style=paper-code,language=cofola,#1}
}
\newtcblisting[auto counter]{cofolacodeblock}[1]{%
    enhanced,
    frame hidden, 
    borderline west = {0.4pt}{0pt}{codeframe},
    opacityfill=0,
    top=-5pt,bottom=-5pt,
    fonttitle=\bfseries,
    listing only,
    listing options={style=paper-code,language=cofola,#1}
}
\lstdefinelanguage{CofolaASP}{
    morekeywords={not},
    morecomment=[l]{\%},
    morestring=[b]"
}
\lstdefinelanguage{CofolaEssence}{
    morekeywords={letting,be,domain,int,find,set,mset,sequence,size,maxSize,maxOccur,such,that,forAll,subsetEq,union,intersect,toInt,freq,sum},
    morecomment=[l]{$},
    morestring=[b]"
}
\newtcblisting{backendcode}[1]{%
    enhanced,
    frame hidden,
    breakable,
    borderline west = {0.4pt}{0pt}{codeframe},
    opacityfill=0,
    top=-5pt,bottom=-5pt,
    listing only,
    listing options={
        style=paper-code,
        language=#1,
        mathescape=false
    }
}
\newtcblisting[auto counter]{problemwithcode}[2]{%
    label=#1,
    enhanced,
    colback=white,
    frame hidden,
    borderline west = {0.4pt}{0pt}{codeframe},
    breakable,
    bottom=-5pt,left=0pt,right=0pt, middle=1pt,
    title={Code Example \thetcbcounter},
    fonttitle=\bfseries,
    comment and listing,
    comment={\textbf{Problem:} #2},
    after={\par\noindent},
    listing options={style=paper-code,language=cofola}
}

\newtheorem{definition}{Definition}
\newtheorem{theorem}{Theorem}
\theoremstyle{definition}
\newtheorem{problem}{P\ignorespaces}
\crefformat{problem}{#2\textbf{P#1}#3}
\newcommand{\code}[1]{\relax\ifmmode\text{\lstinline[style=inline-code]@#1@}\else\lstinline[style=inline-code]@#1@\fi}
\renewcommand{\sharp}{\code{\#}}
\newcommand{\myand}{\code{\&}}
\newcommand{\ourlang}{{Cofola}}
\newcommand{\cofolato}{\rightsquigarrow}

\SetKwFunction{encode}{encode}
\SetKwFunction{decode}{decode}
\newcommand{\overcount}{\mathit{overcount}}

\newcommand{\mln}{\Phi}
\newcommand{\sentence}{\Gamma}
\newcommand{\generalsentence}{{\widehat{\sentence}}}
\newcommand{\recursivesentence}{{\widetilde{\sentence}}}
\newcommand{\fotwoformula}{\psi}
\newcommand{\formula}{\alpha}
\newcommand{\formulas}{\Psi}
\newcommand{\weight}{w}
\newcommand{\vecweight}{\mathbf{\weight}}
\newcommand{\negweight}{\bar{w}}
\newcommand{\negvecweight}{i\mathbf{\negweight}}
\newcommand{\World}{\Omega}
\newcommand{\world}{\omega}
\newcommand{\wfomc}{WFOMC}
\newcommand{\fastwfomc}{FastWFOMC}
\newcommand{\fomc}{FOMC}
\newcommand{\symwfomc}{\ensuremath{\mathsf{WFOMC}}}
\newcommand{\symfomc}{\ensuremath{\mathsf{FOMC}}}
\newcommand{\wmc}{WMC}
\newcommand{\symwmc}{\ensuremath{\mathsf{WMC}}}
\newcommand{\fo}{\ensuremath{\mathbf{FO}}}
\newcommand{\fotwo}{\ensuremath{\mathbf{FO}^2}}
\newcommand{\ctwo}{\ensuremath{\mathbf{C}^2}}
\newcommand{\sctwo}{\ensuremath{\mathbf{SC}^2}}
\newcommand{\setcell}{\mathbf{C}}
\newcommand{\config}{C}
\newcommand{\indicator}{\mathbbm{1}}
\newcommand{\indices}{\mathbf{N}}
\newcommand{\cardinality}{\mathbf{N}}
\newcommand{\domain}{\Delta}
\newcommand{\repdomain}{\tilde{\domain}}
\newcommand{\dftdomain}{D}
\newcommand{\pent}{\mathsf{pEnt}}
\newcommand{\maxsize}{\mathsf{maxSize}}
\newcommand{\exactsize}{\mathsf{size}}
\newcommand{\mult}{\mathsf{mult}}
\newcommand{\disent}{\mathsf{Dis}}
\newcommand{\indisent}{\mathsf{Indis}}
\newcommand{\singletons}{\mathcal{S}}
\newcommand{\dftvec}{\mathbf{M}}
\newcommand{\vecn}{\mathbf{n}}
\newcommand{\veck}{\mathbf{k}}
\newcommand{\vecg}{\mathbf{g}}
\newcommand{\vech}{\mathbf{h}}
\newcommand{\vecj}{\mathbf{j}}
\newcommand{\vecy}{\mathbf{y}}
\newcommand{\vecx}{\mathbf{x}}
\newcommand{\vecu}{\mathbf{u}}
\newcommand{\vecd}{\mathbf{d}}
\newcommand{\innerprod}[2]{\langle#1,#2\rangle}
\newcommand{\nat}{\mathbb{N}}
\newcommand{\real}{\mathbb{R}}
\newcommand{\complex}{\mathbb{C}}
\newcommand{\dftmln}{\mln_{\textrm{DFT}}}
\newcommand{\generaor}{\mathcal{G}}
\newcommand{\pro}{\mathbb{P}}
\newcommand{\extformula}{\varphi}
\newcommand{\rejsentence}{\sentence_{\rm{reject}}}
\newcommand{\rejlineage}{\eta_{\rm{reject}}}
\newcommand{\set}[1]{\{#1\}}
\newcommand{\multiset}[1]{\{#1\}}
\newcommand{\constraint}{C}
\newcommand{\fomodels}[2]{\mathcal{M}_{#1, #2}}
\newcommand{\tree}{\ensuremath{\mathsf{T}}}
\newcommand{\treesum}{\ensuremath{\mathsf{TS}}}
\newcommand{\ufotwo}{\ensuremath{\mathbf{UFO}^2}}
\newcommand{\distinctpair}{\Omega}
\newcommand{\onetypes}{\mathcal{U}}
\newcommand{\twotypes}{\mathcal{B}}
\newcommand{\structure}{\mathcal{A}}
\newcommand{\typeweight}[1]{\langle \weight, \negweight\rangle(#1)}
\newcommand{\generalonetype}{\widetilde{\tau}}
\newcommand{\sumweight}{\mathcal{W}}
\newcommand{\relaxcelltype}[2]{{#1}\downarrow{{#2}}}
\newcommand{\relaxcelltypeb}[2]{({#1}\downarrow{{#2}})}
\newcommand{\proj}[2]{\langle#1\rangle_{#2}}
\newcommand{\repeatable}{\ensuremath{\mathcal{R}}}

\newcommand{\twoheadrightarrowtail}{\mathrel{\mathrlap{\rightarrowtail}}\mathrel{\mkern2mu\twoheadrightarrow}}
\newcommand{\disentity}{{\hat{e}}}
\newcommand{\indisentity}{{\bar{e}}}
\newcommand{\csp}{\textrm{CSP}}
\newcommand{\sharpcsp}{\ensuremath{\#}CSP}
\newcommand{\symsharpcsp}[1]{\ensuremath{\mathsf{\#\langle #1 \rangle}}}
\newcommand{\sem}[1]{\ensuremath{\llbracket #1 \rrbracket}}

\usetikzlibrary{arrows.meta, positioning, shapes.geometric,calc}
\newcommand{\support}[1]{\ensuremath{\text{Supp}(#1)}}
\newcommand{\partition}[2]{\ensuremath{\text{Par}_{#1}(#2)}}
\newcommand{\composition}[2]{\ensuremath{\text{Comp}_{#1}(#2)}}
\newcommand{\mylist}[1]{\ensuremath{\text{List}(#1)}}
\newcommand{\together}[2]{\ensuremath{\text{Together}_{#2}(#1)}}
\newcommand{\evidence}{\ensuremath{\mathcal{E}}}
\newcommand{\cardconst}{\ensuremath{\mathcal{C}}}
\newcommand{\axiom}{\ensuremath{\mathcal{A}}}
\newcommand{\setclass}{\ensuremath{\mathsf{s}}}
\newcommand{\multisetclass}{\ensuremath{\mathsf{ms}}}
\newcommand{\orderedlistclass}{\ensuremath{\mathsf{ol}}}
\newcommand{\setsofset}{\ensuremath{\mathsf{ss}}}
\newcommand{\setsofmultiset}{\ensuremath{\mathsf{sms}}}
\newcommand{\orderedlistsofset}{\ensuremath{\mathsf{ols}}}
\newcommand{\orderedlistsofmultiset}{\ensuremath{\mathsf{olms}}}
\newcommand{\cspv}[1]{\ensuremath{v_{\code{#1}}}}
\newcommand{\cspd}[1]{\ensuremath{D_{\code{#1}}}}
\newcommand{\cspc}[1]{\ensuremath{c_{\code{#1}}}}
\newcommand{\lv}[2]{\ensuremath{\cspv{#1}^{#2}}}
\newcommand{\closedevidence}[3]{\ensuremath{{#1}[{#3}, {#2}]}}
\newcommand{\comcount}[1]{\ensuremath{\#\langle {#1} \rangle}}

\newtheorem{proposition}{Proposition}
\newtheorem{lemma}{Lemma}

\newtheorem{example}{Example}

\newcommand{\restateheading}{}
\newtheorem*{restatebody}{\restateheading}

\newcommand{\lnote}[1]{\textcolor{orange}{[Lucien: #1]}}

\title{Solving Combinatorial Counting Problems with Weighted First-Order Model Counting}

\renewcommand\Authfont{\normalsize}
\renewcommand\Affilfont{\small}

\author[1]{Yuanhong Wang\thanks{Part of the work was done when the author was with Beihang University.}}
\author[2,3]{Juhua Pu}
\author[1]{Yuxu Zhou}
\author[4,5]{Yuyi Wang}
\author[6]{Ond\v{r}ej Ku\v{z}elka\thanks{Corresponding author (\texttt{ondrej.kuzelka@fel.cvut.cz}).}}

\affil[1]{School of Artificial Intelligence, Jilin University, Changchun, China}
\affil[2]{State Key Laboratory of Complex \& Critical Software Environment, Beihang University, China}
\affil[3]{National Research Center for Educational Materials, China}
\affil[4]{CRRC Zhuzhou Institute, Zhuzhou, China}
\affil[5]{Tengen Intelligence Institute, China}
\affil[6]{Czech Technical University in Prague, Prague, Czech Republic}

\date{}

\begin{document}

\maketitle

\begin{abstract}
Combinatorial counting asks how many structures satisfy a specification of dependent choices and constraints, underpinning tasks from assignment counting to probabilistic inference. Manual derivations are fragile when specifications change, and large language models offer no guarantee of correctness, motivating declarative methods that compute exact counts automatically. Weighted first-order model counting (WFOMC), originally developed for statistical relational inference, provides a natural foundation by expressing relational constraints uniformly in first-order logic. Its lifted algorithms exploit exchangeable domain elements to count without full grounding, running in polynomial time in the domain size for fixed sentences in tractable fragments. Bridging high-level specifications to these reusable WFOMC algorithms, however, requires a count-preserving translation handling multiplicities, redundant representations, and dependent choices. We introduce Cofola, a declarative language for combinatorial counting over sets, bags, tuples, sequences, partitions, and compositions, with backend-independent joint counting semantics. We compile Cofola into symbolic WFOMC via relational encodings, polynomial weights, and decoding rules, proving exact count preservation on a supported fragment under exact symbolic computation. For multiset scalability, we introduce an optional exchangeable bag factorization that shares multiplicity symbols across interchangeable entities while provably preserving counts. Across 695 real, synthetic, and growing-domain instances, our WFOMC pipeline solves 675 within a 100-second limit per run, outperforming four alternative counting pipelines. In a controlled ablation, bag factorization raises coverage from 13 to all 32 expanded multi-type bag instances.
\end{abstract}



\section{Introduction}

Combinatorial counting asks how many structures satisfy a specification of objects and constraints.
Exact counts underpin tasks ranging from counting feasible assignments to evaluating discrete probabilistic models~\citep{chavira_probabilistic_2008-1}.
Even a concise problem statement can define a vast space of structures through familiar operations such as selecting elements, arranging them in order, or partitioning them into groups.
Counting such spaces accurately requires a precise account of how choices depend on one another and when two structures are considered distinct.
Two problems adapted from MATH~\citep{hendrycksmath2021} illustrate these requirements.
\begin{problem}
    \label{prob:water_polo}
    A water polo team has 15 members. In how many ways can seven starting players be selected and one of them designated as the goalie?
\end{problem}
\begin{problem}
    \label{prob:mathcounts}
    The word ``MATHCOUNTS'' is spelled out with 10 letter magnets, including two indistinguishable copies of T. How many distinct collections of four letters can be selected if each collection must contain fewer vowels than consonants?
\end{problem}
\noindent In \cref{prob:water_polo}, a solution comprises a starting team and a goalie selected from that team.
The second choice depends on the outcome of the first, and altering either choice yields a distinct solution.
In \cref{prob:mathcounts}, each collection is a multiset, or \emph{bag}, in which elements may repeat.
Swapping the two copies of T leaves a collection unchanged, whereas selecting one copy versus two produces distinct collections.
The vowel condition further restricts which multiplicities may co-occur.

Classical combinatorial methods handle such distinctions through product rules, recurrences, generating functions, and case analysis~\citep{Stanley_2011}.
For instance, selecting the team and then its goalie yields $\binom{15}{7}\cdot 7=45{,}045$ solutions to \cref{prob:water_polo}.
These derivations are tailored to individual problems and must be reworked whenever dependencies or constraints change.
Large language models can tackle such problems directly from natural language and frequently succeed, yet they offer no guarantee that any particular answer is correct.
In a concurrent benchmark study~\citep{zhou2026combeval} built on \ourlang{}, even the strongest evaluated model still answers about $6\%$ of $1{,}500$ instances incorrectly, with or without code execution.
Formal methods therefore play a complementary role by computing exact answers automatically once the objects, dependencies, and constraints have been specified, whether by a person or by a language model.

Declarative modelling languages such as MiniZinc~\citep{nethercote2007minizinc} and \textsc{Essence}~\citep{frisch_essence_2008} decouple problem specification from solving, with \textsc{Essence} natively supporting high-level types such as multisets and partitions.
For counting, however, the representation passed to a solver must preserve the exact \emph{number} of solutions rather than merely their existence.
Conjure achieves this guarantee for \textsc{Essence} by eliminating redundant representations during refinement~\citep{akgun_conjure_2022}.
Propositional encodings for model counters such as \textsc{sharpSAT}~\citep{thurley2006sharpsat} must likewise account for how many Boolean assignments correspond to each combinatorial structure, while grounding interchangeable entities into separate variables.
Although component caching and symmetry-aware techniques can recover part of this regularity~\citep{van_bremen_symmetric_2021-1}, compact reasoning in terms of group sizes and multiplicities is not exposed automatically.
CoLa and its solver CoSo address this limitation using high-level combinatorial abstractions paired with specialized counting rules~\citep{totis_lifted_2023}.
Our objective is to retain high-level abstractions while compiling joint choices over multiple dependent objects into a common logical counting backend.

First-order logic offers a relational representation in which predicates and quantified constraints describe many entities uniformly.
\emph{Weighted first-order model counting} (\wfomc{}) turns this representation into a counting task and emerged as a unifying formulation for inference in statistical relational models~\citep{van_den_broeck_lifted_2011,van_den_broeck_skolemization_2014}.
Given a first-order sentence, a finite domain, and weights for the true and false ground atoms of each predicate, \wfomc{} sums the weights of all satisfying interpretations; when all weights equal $1$, it counts them directly.
To encode \cref{prob:water_polo}, we take the 15 team members as the domain and introduce unary predicates $player$ and $goalie$ for selected players and goalies.
Their dependency is expressed by
\begin{equation}\label{eq:water_polo_fomc}
    \Phi = \forall x:\; \bigl(goalie(x) \rightarrow player(x)\bigr),
\end{equation}
accompanied by the \emph{cardinality constraints} $|player|=7$ and $|goalie|=1$, where the cardinality of a predicate counts its true ground atoms~\citep{kuzelka_weighted_2021-1}.
Because satisfying interpretations stand in one-to-one correspondence with valid team--goalie pairs, the model count of $\Phi$ under these constraints yields the exact answer.

This encoding also illustrates the regularity exploited by \emph{lifted inference}~\citep{van_den_broeck_introduction_2021,tanya_braun_lifted_2016-1,kersting_lifted_2012-1,singla_lifted_2008-1}, which aggregates exchangeable domain elements without fully grounding the sentence.
In \cref{prob:water_polo}, each member falls into one of three roles, and the constraints require eight unselected members, six non-goalie players, and one goalie.
Since all members share the same admissible roles, lifted inference aggregates assignments by role sizes while still respecting individual identities.
On several first-order fragments, lifted \wfomc{} algorithms run in time polynomial in the domain size and the bit length of the weights for a fixed sentence~\citep{van_den_broeck_lifted_2011,kuzelka_weighted_2021-1,toth_lifted_2022-1,zou_faster_2025}.
Prior applications to $k$-regular graphs, $k$-matchings~\citep{timothy_van_bremen_faster_2021}, and stable matchings~\citep{kuula2026stable} highlight this connection to combinatorial counting.
Extending these algorithms to general high-level combinatorial specifications, however, requires a systematic translation of combinatorial objects and their dependencies.

To provide this translation, we introduce \emph{\ourlang{}} (COmbinatorial counting LAnguage with First-Order logic).
It supports six object types spanning unordered collections (sets and bags), ordered collections (tuples and sequences), and unordered or ordered groupings (partitions and compositions).
A program can combine multiple dependent choices subject to constraints on their contents, sizes, and positions.
Each subsequent choice operates on the realized value of its source, which may itself be the outcome of an earlier choice.
The semantics counts all \emph{joint assignments} to the declared objects that satisfy both declarations and constraints.
Distinct intermediate choices are therefore counted separately even when they yield identical values for downstream objects.
For instance, the \ourlang{} program for \cref{prob:water_polo} can be found in \cref{ex:water_polo_cofola}.

While the water-polo encoding enjoys a one-to-one correspondence between interpretations and structures, richer object types break this correspondence.
A count-preserving translation must simultaneously link derived choices to realized sources, track bag multiplicities without creating duplicate domain elements (which we encode as exponents of \emph{polynomial} weights~\citep{kimmig_algebraic_2017-1}), and compensate for representation redundancy when linear orders or labelled partition slots map a single structure to multiple logical models.
\ourlang{} resolves these issues in a unified pipeline that compiles objects and dependencies into a symbolic \wfomc{} query and decodes the resulting polynomial by extracting valid coefficients and dividing out overcounting factors.
For a stated supported fragment under exact symbolic backend computation, we prove that this pipeline returns the exact number of satisfying structures (\cref{sec:count_preservation}).

Assigning separate multiplicity symbols to many distinct entities can inflate the cost of symbolic computation.
To alleviate this bottleneck, we introduce an optional \emph{exchangeable bag factorization} that shares symbols across entities playing identical roles while preserving dependencies between their multiplicities across bags.
We formulate this optimization for components consisting solely of sets and bags derived from initialized objects under the supported operations and exchangeability conditions of \cref{sec:bag_factorization}.
Positive bag containment and equality are enforced locally, size constraints are tracked through shared exponents, and every entity--object coordinate outside a factorized class retains its ordinary encoding.

We implement this approach and evaluate it against four alternative pipelines across real, paired synthetic, and growing-domain benchmarks.
Within a $100$-second time limit per run, our \wfomc{}-based pipeline returns the exact reference count across all three repetitions on $246$ of $247$ manually modelled MATH problems, $269$ of $288$ synthetic programs, and all $160$ growing-domain instances.
In a separate ablation comparing the ordinary and factorized encodings under lifted inference, both solve all $60$ eligible original MATH programs, with factorization adding $1.4\%$ to geometric mean runtime.
When the number of distinct entity types is scaled up in eight MATH bag programs, however, factorization lifts coverage from $13$ of the $32$ expanded instances to all $32$.

The contributions of this paper are as follows.
\begin{itemize}
    \item \textbf{A language with joint counting semantics.}
    We define six combinatorial object types, their dependencies and constraints, and backend-independent joint counting semantics for the structures to be counted (\cref{sec:cofola}).

    \item \textbf{Count-preserving symbolic compilation.}
    We develop relational encodings, polynomial weights, and decoding rules (\cref{sec:cofola_solution,sec:encoding_rules}), and prove that the supported pipeline preserves joint solution counts under exact symbolic backend computation (\cref{sec:count_preservation}).

    \item \textbf{Exchangeable bag factorization.}
    We establish sufficient conditions allowing interchangeable entities to share multiplicity symbols while retaining intermediate choices and cross-bag dependencies, and prove that compatible replacements with exact local factors preserve the decoded count (\cref{sec:bag_factorization}).

    \item \textbf{An implementation and empirical evaluation.}
    We implement \ourlang{}-\wfomc{} and evaluate it against CoLa--CoSo, ASP--Clingo, ESSENCE--Conjure, and grounded CNF--Ganak on $695$ benchmarks, solving $675$ instances and confirming through a controlled ablation that bag factorization restores full coverage ($32/32$) on expanded multi-type bag instances (\cref{sec:experiments}).
\end{itemize}

The rest of the paper follows the architecture of this compilation approach.
After reviewing the required \wfomc{} background in \cref{sec:preliminaries}, we formalize \ourlang{} and its backend-independent counting semantics in \cref{sec:cofola}.
We then reduce \ourlang{} programs to symbolic \wfomc{} queries by presenting the overall compilation pipeline in \cref{sec:cofola_solution} and the ordinary encoding rules in \cref{sec:encoding_rules}, before proving end-to-end count preservation in \cref{sec:count_preservation}.
To address multiset scaling, \cref{sec:bag_factorization} develops exchangeable bag factorization and proves its correctness.
Finally, \cref{sec:experiments} evaluates the implementation against alternative pipelines, \cref{sec:related} positions our approach relative to prior work, and \cref{sec:conclusion} summarizes limitations and open directions.

\section{Preliminaries}
\label{sec:preliminaries}

This section introduces the combinatorial notation and the symbolic weighted first-order model-counting interface used throughout our compilation pipeline.

\subsection{Combinatorial Notation}
\label{subsec:notations}

We write $[n]=\{1,\ldots,n\}$ and $[n]_0=\{0,\ldots,n-1\}$.
A \emph{bag} (multiset) $M=\multiset{e_1:k_1,\ldots,e_r:k_r}$ with $k_i\in\nat$ assigns multiplicity $M(e_i)=k_i$ to each listed entity and multiplicity $0$ to all other entities.
Its support is $\support{M}=\{e:M(e)>0\}$ and its cardinality is $|M|=\sum_e M(e)$.
Membership $e\in M$ holds when $M(e)>0$, subbag inclusion $M\subseteq_m N$ holds when $M(e)\leq N(e)$ for every entity $e$, and two bags are equal when all of their multiplicities coincide.
Bag union, intersection, and difference take the element-wise maximum, minimum, and truncated difference $\max(M(e)-N(e),0)$ of multiplicities, respectively, while additive union $\uplus$ sums them.
For a set $S$, $2^S$ denotes its power set, and for a bag $M$, $2^M=\{N:N\subseteq_m M\}$ denotes the set of its subbags.

An \emph{ordered list} $L=[e_0,\ldots,e_{n-1}]$ may contain repeated entities, with positions indexed by $[n]_0$ so that $L[i]=e_i$, and $i\lessdot j$ indicates immediate succession ($j=i+1$).
Membership $e\in L$ holds when $e=L[i]$ for some $i\in[n]_0$.
A \emph{partition} is an unordered multiset of parts, whereas a \emph{composition} is an ordered list of parts.
For a set source, the parts are pairwise disjoint and their union equals the source; for a bag source, the additive union of the parts equals the source.
Parts may be empty unless restricted by an explicit constraint.

\subsection{Symbolic Weighted First-Order Model Counting}
\label{sub:wfomc}

We work in function-free first-order logic over a finite domain $\domain$.
An interpretation $\world$ assigns each predicate $R/k$ of arity $k$ a relation $\world_R\subseteq\domain^k$.
We write $\world\models\sentence$ when $\world$ satisfies a sentence $\sentence$ under standard first-order semantics, and denote the set of all models of $\sentence$ over $\domain$ by $\fomodels{\sentence}{\domain}$.
The two-variable fragment with counting quantifiers, $\ctwo{}$, restricts formulas to at most two logical variables (which may be reused under quantification) and permits counting quantifiers such as $\exists_{\leq k}x:\phi$, which holds when at most $k$ domain elements satisfy $\phi$.

A weighting $(\weight,\negweight)$ assigns to each predicate $R$ a weight $\weight(R)$ for its true ground atoms and a weight $\negweight(R)$ for its false ground atoms, with unspecified weights defaulting to $1$.
In this paper, weights may be polynomials in $\real[\mathbf{u}]=\real[u_1,\ldots,u_s]$.
\begin{definition}[Weighted first-order model counting]
For a sentence $\sentence$, domain $\domain$, and weights $(\weight,\negweight)$,
\begin{align*}
    W(\world;\weight,\negweight)
        &=\prod_{R/k\in\mathcal{P}_{\sentence}}
          \weight(R)^{|\world_R|}
          \negweight(R)^{|\domain|^k-|\world_R|},\\
    \symwfomc(\sentence,\domain;\weight,\negweight)
        &=\sum_{\world\in\fomodels{\sentence}{\domain}}
          W(\world;\weight,\negweight),
\end{align*}
where $\mathcal{P}_{\sentence}$ is the predicate vocabulary of $\sentence$.
First-order model counting (\fomc{}) is the special case with all weights equal to $1$, denoted $\symfomc(\sentence,\domain)=|\fomodels{\sentence}{\domain}|$.
\end{definition}

For a polynomial $f\in\real[\mathbf{u}]$, $[\mathbf{u}^{\mathbf{k}}]f$ denotes the coefficient of the monomial $\mathbf{u}^{\mathbf{k}}$ with exponent vector $\mathbf{k}$.
Given a condition $A$ on exponent vectors $\mathbf{k}\in\mathbb{N}^{s}$ (such as arithmetic comparisons or ordering conditions on the exponents), we write $\mathbf{k}\models A$ when $\mathbf{k}$ satisfies $A$ and define $[A]f=\sum_{\mathbf{k}\in\mathbb{N}^{s}:\,\mathbf{k}\models A}[\mathbf{u}^{\mathbf{k}}]f$, where only the finitely many monomials appearing in $f$ contribute.
Crucially, $A$ filters monomial exponents rather than imposing numerical values on the indeterminates $\mathbf{u}$.

The cardinality $|R|$ of a predicate $R$ is interpreted as $|\world_R|$.
\emph{Cardinality constraints} are arithmetic equalities and inequalities over predicate cardinalities, which can be enforced using symbolic weights followed by coefficient selection~\citep{kuzelka_weighted_2021-1}.
\begin{example}\label{ex:fomc}\label{ex:cardinality}
Let $\sentence=\forall x:\bigl((Red(x)\lor Black(x))\land\neg(Red(x)\land Black(x))\bigr)$ over a domain of three distinguishable balls.
Setting $\weight(Red)=u$ and all other weights to $1$ yields the weighted model count $(1+u)^3$.
Because the coefficient of $u^j$ counts assignments with exactly $j$ red balls, the number of assignments with at most two red balls is $[u\leq2](1+u)^3=7$.
Setting all weights to $1$ recovers the unrestricted count $2^3=8$.
\end{example}

We also use \emph{unary evidence}, consisting of ground literals that fix unary predicate values on named domain elements.
For a unary predicate $P$ and a subset $S\subseteq\domain$, the \emph{closed-world evidence} is
\begin{equation}
    \label{eq:closed_world_evidence}
    \closedevidence{P}{\domain}{S}
    =\{P(c):c\in S\}\cup\{\neg P(c):c\in\domain\setminus S\},
\end{equation}
which we also identify with the conjunction of its literals.
Restricting evidence to unary predicates is essential for domain-liftability, since arbitrary binary evidence does not preserve the same tractability guarantee~\citep{van_den_broeck_conditioning_2012-1}.

\subsection{Order Axioms and Lifted Inference}
\label{sub:wfomc_order_axioms}

The linear-order axiom $Linear(L,Pred)$ requires $L$ to be a \emph{reflexive total order} and $Pred$ to be its \emph{strict immediate-successor relation}, defined by
\begin{equation*}
    Pred(x,y)\ \leftrightarrow\
    \left(x\neq y\land L(x,y)\land
    \neg\exists z\,(z\neq x\land z\neq y\land L(x,z)\land L(z,y))\right).
\end{equation*}
When the successor relation is not needed, we write simply $Linear(L)$.
Rather than axiomatizing transitivity and immediate succession directly inside $\ctwo{}$, lifted solvers handle this axiom through a specialized counting procedure~\citep{toth_lifted_2022-1,zou_faster_2025}, supplying the order structure required by relative-position constraints.

\begin{theorem}[Summary of results from \cite{beame_symmetric_2015-1,kuzelka_weighted_2021-1,wang_lifted_2024,toth_lifted_2022-1}]\label{thm:complexity}
    Let $\sentence$ be a \emph{fixed} \ctwo{} sentence, and let $(\weight, \negweight)$ be a pair of rational weighting functions.
    For any domain $\domain$, any set $\cardconst$ of cardinality constraints, any unary evidence $\evidence$, and any linear order axiom $\axiom$, the \wfomc{} problem $\symwfomc(\sentence\land \cardconst\land \evidence\land \axiom, \domain; \weight, \negweight)$ can be computed in time polynomial in the domain size $|\domain|$, the number of bits required to represent the weights, the size (and parameters) of the cardinality constraints, and the number of ground literals in the unary evidence.
\end{theorem}

The fixed-sentence assumption is essential, because the problem is already \#\P-complete for two-variable first-order logic $\mathrm{FO}^2$ (and hence $\ctwo{}$) when the sentence is part of the input~\citep{beame_symmetric_2015-1}.
Since our compilation employs polynomial weights and may generate sentences whose size grows with the program's entities, we analyse the resulting complexity implications in \cref{sub:liftability}.
In our implementation, compiled queries are evaluated with \textsc{FastWFOMC}~\citep{timothy_van_bremen_faster_2021} and \textsc{IncrementalWFOMC2}~\citep{zou_faster_2025}, where the latter provides support for linear-order axioms.
The backend solver is publicly available at \url{https://github.com/yuanhong-wang/WFOMC}.

\section{The \ourlang{} Language and Counting Semantics}
\label{sec:cofola}

A \ourlang{} program specifies a space of combinatorial structures and asks how many satisfy its constraints.
Modelling a problem consists of declaring the objects to be chosen or arranged, specifying their dependencies, and imposing constraints on admissible outcomes.
The examples below introduce these declarations and constraints informally before \cref{sub:language_semantics} formalizes the counting semantics; \cref{fig:abstract_syntax} in the appendix collects the core grammar.

\subsection{Modelling Dependent Choices}

An object declaration \code{I = expr} names a combinatorial object and specifies its admissible values.
Initializer declarations fix an input collection, whereas choice declarations specify alternatives that may depend on earlier objects.
In particular, \code{choose} does not perform a random draw; rather, it declares all permitted subcollections as alternatives to be counted.

\begin{example}\label{ex:water_polo_cofola}
The water-polo problem in \cref{prob:water_polo} is expressed by
\begin{cofolacode}{}
players = set(player0...15)
starting_team = choose(players, 7)
goalie = choose(starting_team, 1)
\end{cofolacode}
\noindent The first line fixes the available players; the half-open range \code{player0...15} denotes the 15 entities \code{player0} through \code{player14}.
The second line permits any seven-player subset, and the third permits any one-player subset of that realized starting team rather than of the full player set.
For example, when \code{player0} through \code{player6} are selected, choosing \code{player0} as goalie is valid, whereas choosing \code{player7} is not.
\end{example}

A solution assigns a value to \emph{every} declared object consistently with its dependencies.
Keeping the same starting team while changing its goalie therefore yields a distinct solution, so the program counts team--goalie pairs and returns $\binom{15}{7}\cdot7=45{,}045$.
By contrast, adding \code{bench = players - starting_team} does not change the count, because the bench is uniquely determined once the starting team is fixed.
This distinction between a nondeterministic choice and a deterministic value carries over to all object types introduced below.

Constraints filter these joint assignments.
For example, \code{player0 in starting_team} requires a specific player to be selected.
Constraints combine under the Boolean connectives \code{and}, \code{or}, and \code{not}; for instance, \code{(player0 in goalie) or (player0 not in starting_team)} requires that \code{player0} either be the goalie or not be selected.

Each declaration introduces a fresh name with a fixed object type, and any referenced source must be declared earlier, keeping the dependency graph acyclic.
The language performs no implicit type casting.
Sets and bags can be initialized by listing entities, whereas the remaining four object types are constructed from an existing set or bag.
\cref{tab:language_objects} summarizes the constructors and constraints for all six types, and the subsections below explain the counting conventions that distinguish them.

\begin{table}[tbp]
\centering
\begin{threeparttable}
\caption{The six object types of \ourlang{} and representative syntax.}
\label{tab:language_objects}
\small
\setlength{\tabcolsep}{5pt}
\renewcommand{\arraystretch}{1.18}
\begin{tabularx}{\linewidth}{@{}>{\raggedright\arraybackslash\bfseries}p{.13\linewidth}>{\raggedright\arraybackslash}p{.17\linewidth}>{\raggedright\arraybackslash}p{.28\linewidth}>{\raggedright\arraybackslash}X@{}}
\toprule
Type & \textbf{Values} & \textbf{Constructors} & \textbf{Operations and constraints} \\
\midrule
Set & Unordered\newline distinct entities
    & \code{set(e1, ...)}\newline \code{choose(I, k)}
    & \code{+}, \code{\&}, \code{-}, \code{|I|}\newline
      \code{in}, \code{subset}, \code{disjoint}, \code{==} \\
\addlinespace[3pt]
Bag & Unordered\newline with multiplicities
    & \code{bag(e: k, ...)}\newline \code{choose(I, k)}\newline \code{choose_replace(S, k)}
    & Set operations and constraints,\newline
      plus \code{++}, \code{supp(I)}, \code{I.count(e)} \\
\midrule
Tuple & Ordered list\newline repetitions allowed
    & \code{tuple(I)}\newline \code{choose_tuple(I, k)}
    & \code{T[i] == e}, \code{T[i] in A}, \code{e in T}\newline
      \code{T.count(A)}, \code{T.dedup_count(A)} \\
\addlinespace[3pt]
Sequence & Ordered list\newline repetitions allowed
    & \code{sequence(I)}\newline \code{choose_sequence(I, k)}
    & \code{together(S) in Q}, \code{(A < B) in Q}\newline
      \code{p in Q}, \code{p not in Q}, \code{Q.count(p)} \\
\midrule
Partition & Unordered\newline multiset of $k$ parts
    & \code{partition(I, k)}
    & \code{cst for part in P} \\
\addlinespace[3pt]
Composition & Ordered list\newline of $k$ parts
    & \code{compose(I, k)}
    & \code{C[i]}, \code{cst for part in C} \\
\bottomrule
\end{tabularx}
\end{threeparttable}
\end{table}

\subsection{Sets and Bags}
\label{sec:basic_objects}

Sets and bags represent unordered collections.
The declaration \code{S = set(a, b)} fixes a set of two distinct entities, whereas \code{B = bag(a: 2, b: 1)} fixes a bag containing two copies of \code{a} and one copy of \code{b}.
Its cardinality \code{|B|} is $3$, and the multiplicity \code{B.count(a)} is $2$.
The \emph{support} of a bag is the set of distinct entities present regardless of their multiplicities; thus \code{supp(B)} evaluates to $\{a,b\}$ and \code{|supp(B)|} is $2$.

The constructor \code{choose(I, k)} declares a subcollection of size $k$ without replacement and preserves the source type.
For the bag \code{B} above, \code{choose(B, 2)} permits only $\multiset{a:2}$ and $\multiset{a:1,b:1}$, because selecting either copy of $a$ together with $b$ yields the same bag.
Omitting $k$ permits all subcollections, including the empty one.
With replacement, \code{choose_replace(S, k)} permits any size-$k$ bag over the entities of a set \code{S}.
When \code{R = choose_replace(S)} has a nonempty source and no explicit size argument, a size bound such as \code{|R| <= 3} is needed to keep the space of alternatives finite.

Once collections are declared, deterministic operations build derived objects without introducing additional choices.
For operands of the same type, \code{+}, \code{\&}, and \code{-} denote union, intersection, and difference.
On bags, union takes the larger multiplicity, intersection takes the smaller multiplicity, and difference subtracts the second multiplicity from the first, truncating at zero.
The bag-only operator \code{++} instead sums multiplicities; combining two bags that each contain one copy of $a$ yields one copy under \code{+} but two under \code{++}.
Derived objects can themselves serve as sources for subsequent choices, as in \code{choose(B1 & B2, k)}.

Constraints restrict the admissible values of these objects.
Membership is written \code{e in I} or \code{e not in I} and tests only whether the entity is present.
For two objects of the same type, \code{I1 subset I2} requires containment, \code{I1 disjoint I2} forbids shared entities, and \code{I1 == I2} or \code{I1 != I2} tests equality or inequality.
Bag containment and equality compare multiplicities, whereas bag disjointness depends only on supports.
All of these constraints can be combined using Boolean connectives.

Size constraints compare linear combinations of cardinalities and multiplicities against an integer constant using \code{==}, \code{!=}, \code{<}, \code{<=}, \code{>}, or \code{>=}.
For example, \code{|I1| - 2|I2| == 0} requires the first object to have twice the size of the second, while \code{B.count(a) >= 2} requires at least two copies of $a$.
The multiplicity query \code{B.count(e)} applies only to a bag and a single entity, not to a collection.
Because the surface syntax requires a constant right-hand side, a comparison between two object sizes is written as \code{|I1| - |I2| < 0}.

\begin{example}\label{ex:mathcounts}
The MATHCOUNTS problem in \cref{prob:mathcounts} illustrates these bag operations.
Both input collections are declared as bags so that intersections with the chosen bag retain the selected multiplicities.
\begin{cofolacode}{}
vowels = bag(A: 1, O: 1, U: 1)
consonants = bag(M: 1, T: 2, H: 1, C: 1, N: 1, S: 1)
magnets = vowels ++ consonants
chosen = choose(magnets, 4)
chosen_vowels = chosen & vowels
chosen_consonants = chosen & consonants
|chosen_vowels| - |chosen_consonants| < 0
\end{cofolacode}
\noindent The inputs comprise nine distinct letters and ten magnets in total, counting T twice.
Intersecting \code{chosen} with \code{vowels} and \code{consonants} determines the selected vowels and consonants without adding choices, and their cardinalities count copies rather than distinct letters.
For instance, $\multiset{T:2,M:1,A:1}$ satisfies the constraint with one vowel and three consonants, whereas $\multiset{T:2,A:1,O:1}$ does not.
\end{example}

\subsection{Ordered Objects: Tuples and Sequences}
\label{sec:ordered_objects}

Both \code{tuple(I)} and \code{sequence(I)} arrange all entities of a source set or bag into an ordered list while retaining its multiplicities.
They differ in the constraints they support, with tuples offering \emph{absolute}-position constraints and sequences offering \emph{relative}-order and adjacency patterns.
Thus ``the first letter is C'' calls for a tuple, whereas ``the mathematics books are together'' calls for a sequence; both types still count distinct lists.
This separation enables the compiler to represent tuples via a position-to-entity function and sequences via a linear-order axiom (\cref{sub:enc_tuples,sub:enc_sequences}).
The current language does not combine both constraint families on a single ordered object.

The constructors \code{choose_tuple(I, k)} and \code{choose_sequence(I, k)} abbreviate selecting $k$ entities and arranging them into a list.
Their \code{choose_replace_*} variants select with replacement before arranging; the compiler desugars these abbreviations into dependent declarations.
As with basic choices, the size argument $k$ may be omitted.

Tuples support zero-indexed absolute-position constraints.
Specifically, \code{T[i] == e} fixes the entity at position $i$, and \code{T[i] in S} restricts it to a source collection.
Tuples also support membership queries \code{e in T}, requiring \code{e} to occur at least once in \code{T}, as well as counting queries \code{T.count(S)} and \code{T.dedup_count(S)}, which count total occurrences and distinct entities from a source collection within the tuple, respectively.

\begin{example}\label{ex:fourletter}
The following problem and \ourlang{} program illustrate tuple constraints.
\begin{problem}\label{prob:fourletter}
How many different four-letter arrangements can be formed using the six letters $A,B,C,D,E,F$, if the first letter must be $C$, another letter must be $B$, and no letter is repeated?
\end{problem}
\begin{cofolacode}{}
letters = set(A, B, C, D, E, F)
arr = choose_tuple(letters, 4)
arr[0] == C
arr.count(set(B)) > 0
\end{cofolacode}
\noindent The list $[C,A,B,D]$ is a valid solution, whereas $[C,A,D,E]$ is excluded because it omits $B$.
After fixing $C$ at position $0$, we choose two of the four letters other than $B$ and $C$ and arrange them with $B$ in the remaining three positions, yielding $\binom{4}{2}3!=36$ solutions.
\end{example}

Sequences support two global patterns.
Specifically, \code{together(S) in Q} requires all occurrences of entities in a set \code{S} to form at most one contiguous block, and \code{(A < B) in Q} requires \emph{every} occurrence matching \code{A} to strictly precede \emph{every} occurrence matching \code{B}, where \code{A} and \code{B} may be sets or individual entities.

Sequences also support two local patterns expressing adjacency.
The ordered pattern \code{(A, B)} matches an occurrence from \code{A} immediately followed by one from \code{B}, whereas \code{next_to(A, B)} permits either order.
For a local pattern \code{p}, \code{p in Q} requires at least one match and \code{p not in Q} forbids all matches, while \code{Q.count(p)} counts matching adjacent position pairs.
Each adjacent position pair is counted at most once even when the pattern arguments overlap.
For example, $[a,a,b]$ has one \code{next_to(a, b)} match and one \code{next_to(a, a)} match.
\cref{app:ordered_constraint_semantics} gives the formal position-based definitions of all ordered constraints.

\begin{example}\label{ex:books}
Arranging seven books so that the two mathematics books stay together and precede the physics book is modelled by
\begin{cofolacode}{}
math_books = set(math1, math2)
physics_books = set(physics)
books = set(math1, math2, physics, book4, book5, book6, book7)
arr = sequence(books)
together(math_books) in arr
(math_books < physics_books) in arr
\end{cofolacode}
\noindent The two mathematics books form a contiguous block whose internal order may vary, and the physics book must appear later but need not be immediately adjacent to the block.
Treating the mathematics block as one unit yields $6!$ orders of six items, half placing the block before physics; multiplying by the two internal orders of the mathematics books gives $2\cdot6!/2=720$ solutions.
\end{example}

\subsection{Grouped Objects: Partitions and Compositions}
\label{sec:grouped_objects}

The constructors \code{partition(I, k)} and \code{compose(I, k)} divide a source set or bag into exactly $k$ parts.
For a set source, the parts are pairwise disjoint subsets whose union equals the source; for a bag source, the part multiplicities sum to the source multiplicity for every entity.
A composition orders its parts, whereas a partition is an unordered multiset of parts.
In particular, identical bag-valued parts are permitted, and permuting the parts of a partition does not produce a new solution.
Unlike some combinatorial conventions, both partitions and compositions permit empty parts; non-empty groupings can be enforced by explicit size constraints as shown below.

An individual part of a composition can be indexed via \code{C[i]} for $i\in[k]_0$, yielding an object of the same type as the source collection, whereas a partition cannot be indexed.
An indexed composition part may itself serve as the source of a further partition or composition.
For either grouped type, a part-wise constraint \code{cst for part in P} requires the atomic condition \code{cst} to hold for every part occurrence.
Here \code{part} is a local binder ranging over existing parts, and \code{cst} may combine deterministic operations and queries over that part and previously declared objects.
For example, \code{|part| > 0 for part in P} excludes empty parts.
Each part-wise statement introduces a fresh scoped binder even when the same variable name is reused.

\ourlang{} does \emph{not} permit nondeterministic choices, arrangements, or groupings inside a part-wise condition.
Specifically, the body of a part-wise statement cannot contain constructors such as \code{choose}, \code{choose_replace}, \code{tuple}, \code{sequence}, \code{partition}, or \code{compose}, even when their arguments would force a unique outcome.
This restriction guarantees that relabelling the parts of a partition together with their deterministic subexpressions preserves every part-wise test while leaving all declared objects unchanged (\cref{prop:part_relabelling}).

\begin{example}\label{ex:basketball}
Dividing ten players into two unordered teams of five is expressed by
\begin{cofolacode}{}
team = set(player0...10)
groups = partition(team, 2)
|part| == 5 for part in groups
\end{cofolacode}
\noindent Selecting five players determines the complementary five, and swapping the two teams leaves the partition unchanged, giving $\binom{10}{5}/2=126$ solutions.
Replacing \code{partition} with \code{compose} distinguishes the first and second teams and yields $252$ solutions.
\end{example}

\begin{example}\label{ex:partwise_shared_choice}
A globally chosen set can constrain every part without introducing a separate choice per part.
\begin{cofolacode}{}
S = set(a, b, c, d)
R = choose(S, 2)
P = partition(S, 2)
|p| == 2 for p in P
|p & R| == 1 for p in P
\end{cofolacode}
\noindent Each of the three partitions into pairs admits four values of \code{R}, giving $3\cdot4=12$ structures.
The intersection \code{p & R} is determined by the current part and the shared value of \code{R}; placing \code{choose} inline inside either part-wise body would be invalid.
\end{example}

\subsection{Formal Counting Semantics}
\label{sub:language_semantics}

We summarize the formal counting semantics of \ourlang{} here and defer the full specification to \cref{app:full_semantics}.
The semantics is defined denotationally, mapping each valid program to a combinatorial counting problem.

Let $D$ be a finite entity domain and let $\mathcal{U}$ be the universe of combinatorial instances over $D$, encompassing sets, finite bags, ordered lists, and collections thereof.
A \emph{combinatorial object} is a triple $(v,(d_1,\ldots,d_k),f)$, where $v$ is the object variable, $d_1,\ldots,d_k$ are its dependency variables, and $f:\mathcal{U}^k\to 2^{\mathcal{U}}$ maps each dependency assignment to the set of admissible values for $v$.
A deterministic operation returns a singleton set, whereas a choice operation may return multiple alternatives.

A \emph{combinatorial structure} $\structure$ assigns a value in $\mathcal{U}$ to every object variable while satisfying all declarations, so that
\begin{equation*}
    \structure(v)\in f(\structure(d_1),\ldots,\structure(d_k))
    \qquad\text{for each }(v,(d_1,\ldots,d_k),f)\in O.
\end{equation*}
The set of all such structures forms the \emph{combinatorial base} $\mathcal{B}$.
A constraint $R(v_1,\ldots,v_r)$ specifies a relation $R\subseteq\mathcal{U}^r$ and holds under $\structure$ when $(\structure(v_1),\ldots,\structure(v_r))\in R$, with Boolean connectives interpreted in the standard way.

\begin{definition}[Combinatorial counting problem and program semantics]
A combinatorial counting problem is a triple $\langle D,O,C\rangle$ of entities, objects, and constraints.
Its answer is
\begin{equation*}
    \comcount{D,O,C}
    =\bigl|\{\structure\in\mathcal{B}:\structure\models c
                 \text{ for every }c\in C\}\bigr|.
\end{equation*}
A valid \ourlang{} program $P$ denotes such a problem $\sem{P}=\langle D,O,C\rangle$, and its answer is $\comcount{D,O,C}$.
\end{definition}

The semantics is compositional, mapping each statement to its contribution to $\langle D,O,C\rangle$.
Object declarations extend $O$, and initializer declarations also extend $D$, inserting each entity once regardless of its bag multiplicity.
Aliases refer to existing objects, constraint statements extend $C$, and ill-typed statements or undefined references render the program invalid.
Crucially, the semantics counts \emph{joint assignments} over all declared objects rather than projecting onto a single designated output.
Wherever nondeterministic constructors are permitted, two separate occurrences denote distinct choices even when their source and parameters coincide.
In the water-polo program of \cref{ex:water_polo_cofola}, let $t$ denote the chosen team and $g$ its goalie.
The goalie declaration contributes $(g,(t),f)$ with $f(T)=\{\{e\}: e\in T\}$.
Two different teams containing the same goalie produce different assignments to $t$, so the intermediate team choice contributes to the count even without an explicit constraint on the team.

Counting joint assignments keeps the semantics compositional and matches multi-stage combinatorial specifications without requiring an explicit projection step.
As a consequence, adding a deterministic derived object never alters the count, whereas declaring an unconstrained nondeterministic auxiliary object multiplies the count by its number of admissible values.
Introducing an explicit projection or answer-variable declaration to marginalize out auxiliary nondeterministic choices is a natural direction for future language extensions.

\section{The Compilation Pipeline}
\label{sec:cofola_solution}

\begin{figure}[tbp]
\centering
\hypersetup{linkcolor=figurenavy}
\begin{tikzpicture}[
    >=Latex, font=\footnotesize,
    st/.style={draw=black!55, rounded corners=2pt, fill=gray!4, align=center, inner sep=3pt, minimum height=0.9cm},
    flow/.style={->, thick, draw=figurenavy},
    note/.style={font=\scriptsize, text=black!70, align=center}
]
\node[st] (prog) {\ourlang{}\\program};
\node[st, right=0.45cm of prog] (pre) {preprocess\\(bounds, rewrites)};
\node[st, right=0.45cm of pre] (exp) {expand\\(disjoint branches)};
\node[st, right=0.45cm of exp] (spl) {split\\(independent components)};
\node[st, right=0.45cm of spl] (enc) {encode};
\node[st, below=0.9cm of enc] (wf) {symbolic \wfomc{}};
\node[st, left=1.6cm of wf] (dec) {decode\\(select, correct)};
\node[st, left=1.2cm of dec] (comb) {combine:\\$\sum$ branches, $\prod$ components};
\draw[flow] (prog) -- (pre);
\draw[flow] (pre) -- (exp);
\draw[flow] (exp) -- (spl);
\draw[flow] (spl) -- (enc);
\draw[flow] (enc) -- node[right, note]{$(\sentence,\domain,\weight,\negweight)$} (wf);
\draw[flow] (wf) -- node[above, note]{polynomial $f$} (dec);
\draw[flow] (dec) -- node[above, note]{count} (comb);
\node[note, below=0.05cm of exp] {\cref{sub:decomposition}};
\node[note, below=0.05cm of spl] {\cref{sub:decomposition}};
\node[note, below=0.05cm of pre] {\cref{sub:preprocessing}};
\node[note, above=0.05cm of enc] {\cref{sec:encoding_rules}};
\node[note, below=0.05cm of dec] {\cref{sub:encoding}};
\end{tikzpicture}
\caption{Stages of \cref{alg:compile}. Expansion and splitting produce independent components with atomic constraints and fixed ordered lengths; each is encoded, counted, and decoded separately, and the decoded counts are summed over branches and multiplied over components.}
\label{fig:pipeline}
\end{figure}

\begin{algorithm}[tbp]
    \DontPrintSemicolon
    \SetNoFillComment
    \caption{Solve \ourlang{} programs with \wfomc{}}
    \label{alg:compile}
    \KwIn{A supported \ourlang{} program $P_0$ with semantics $\sem{P_0}=\langle D,O,C\rangle$}
    \KwOut{The answer $\comcount{D, O, C}$}
    \SetKwFunction{expand}{expand}
    \SetKwFunction{split}{split}
    \SetKwFunction{preprocess}{preprocess}
    $answer \gets 0$ \;
    $(P_{simp}, rec) \gets \preprocess(P_0)$ \label{line:global_preprocess}\;
    \lIf{$rec = \bot$}{\Return $0$ \tcp*[f]{problem unsatisfiable}}
    \ForEach{$P \gets \expand(P_{simp}, rec)$\label{line:expand}}{
        $sub\_count\gets 1$\;
        \ForEach{$P' \gets \split(P)$\label{line:decompose}}{
            $(P'_{simp}, rec') \gets \preprocess(P')$ \label{line:local_preprocess}\;
            \lIf{$rec' = \bot$}{$sub\_count \gets 0$; \textbf{break} \tcp*[f]{branch has no solutions}}
            $(\sentence, \domain, \weight, \negweight, A, \overcount, \Sigma) \gets \encode(P'_{simp}, rec')$ \label{line:encode}\;
            $f \gets \symwfomc(\sentence, \domain; \weight, \negweight)$\;
            $sub\_count \gets sub\_count \times \decode(f, A, \overcount, \Sigma)$ \label{line:solve}\;
        }
        $answer \gets answer + sub\_count$ \;
    }
    \Return $answer$\;
\end{algorithm}

Given a supported \ourlang{} program $P_0$ with semantics $\sem{P_0}=\langle D,O,C\rangle$, \cref{alg:compile} computes $\comcount{D,O,C}$ through \wfomc{}, following the pipeline in \cref{fig:pipeline}.

The encoder at the centre of the pipeline accepts only a \emph{conjunction of atomic constraints} over objects with \emph{fixed} ordered lengths, and requires bounds on sizes and multiplicities to delimit polynomial exponents.
The surrounding stages establish these preconditions.
\emph{Preprocessing} derives the required bounds and simplifies the program, while \emph{case expansion and decomposition} reduce it to independent components whose counts combine by the sum and product rules.
This section presents these stages and the encoding--decoding interface.
\cref{sec:encoding_rules} details the encoding rules, and \cref{sec:count_preservation} proves that the pipeline preserves counts.

\subsection{Preprocessing}
\label{sub:preprocessing}

Preprocessing extracts the structural bounds required by the encoder and simplifies the program without altering its count.
It returns a simplified program together with an \emph{analysis record}, or $\bot$ when a contradiction establishes unsatisfiability.
Preprocessing runs once globally (\cref{line:global_preprocess}) and again on each split component (\cref{line:local_preprocess}), where branch-specific conditions often permit tighter bounds and further simplification.

For each object $v\in O$, including any referenced parts of grouped objects, the analysis record stores four structural bounds.
\begin{itemize}
    \item $\pent(v)\subseteq D$ is the \emph{potential entity set}, containing every entity that may occur in $v$.
    \item $\exactsize(v)\in\mathbb{N}\cup\{\bot\}$ is its \emph{exact} cardinality when known, with $\bot$ denoting an unknown size.
    \item $\maxsize(v)\in\mathbb{N}\cup\{\infty\}$ is an \emph{upper bound} on its cardinality.
    \item For a bag object $v$, $\mult_v(e)$ is an \emph{upper bound} on the multiplicity of each entity $e$.
\end{itemize}
These bounds may be conservative, provided they never exclude a feasible value.
They are inferred from object dependencies and compatible size constraints, and are refined whenever rewrites or branch conditions tighten the problem.
The inference rules appear in \cref{sec:encoding_rules}, \cref{app:entity_analysis}, and \cref{app:max_size_inference} show how the encoder uses each bound.

\begin{table}[tbp]
\centering
\caption{Structural analysis bounds for the MATHCOUNTS problem in \cref{ex:mathcounts} before branch-specific refinement, showing potential entities $\pent(v)$, size upper bound $\maxsize(v)$, exact size $\exactsize(v)$, and entity multiplicity upper bound $\mult_v(e)$.}
\label{tab:analysis_record}
\small
\begin{tabular}{lllll}
\toprule
Object & $\pent(v)$ & $\maxsize(v)$ & $\exactsize(v)$ & $\mult_v(e)$ \\
\midrule
$v_{vowels}$ & $\{A, O, U\}$ & 3 & 3 & 1 for all $e$ \\
$v_{consonants}$ & $\{M, T, H, C, N, S\}$ & 7 & 7 & 2 for $T$, 1 otherwise \\
$v_{magnets}$ & $\{A, O, \dots, S\}$ & 10 & 10 & 2 for $T$, 1 otherwise \\
$v_{chosen}$ & $\{A, O, \dots, S\}$ & 4 & 4 & 2 for $T$, 1 otherwise \\
$v_{chosen\_vowels}$ & $\{A, O, U\}$ & 3 & $\bot$ & 1 for all $e$ \\
$v_{chosen\_consonants}$ & $\{M, T, H, C, N, S\}$ & 4 & $\bot$ & 2 for $T$, 1 otherwise \\
\bottomrule
\end{tabular}
\end{table}

\begin{example}\label{ex:analysis_record}
\cref{tab:analysis_record} illustrates the analysis record for the MATHCOUNTS problem.
The chosen bag has exact cardinality four, and $T$ has multiplicity bound two.
The upper bound of three on $v_{chosen\_vowels}$ follows from its source bag and does not assert that three vowels can occur in a solution satisfying the final comparison.
\end{example}

Using these bounds, the compiler simplifies the program and detects contradictions early.
Constant folding, for instance, evaluates \code{magnets = vowels ++ consonants} to an initialized bag.
A choice of $k$ elements without replacement from a set or bag known to have size $k$ must equal its source and can therefore be replaced by a reference to that source, provided the equality $|u|=k$ remains enforced on the surviving source $u$ (either by its constructor or by an explicit retained size constraint).
Known cardinalities likewise simplify arithmetic constraints only when the definitions or constraints enforcing those cardinalities remain intact.
Identical deterministic definitions may be shared, whereas independent choices are never merged merely because their constructors and arguments coincide.
\cref{app:optimizations} states the count-preservation conditions for these rewrites, and \cref{app:sanity_checks} details contradiction detection and supported-input checks.

\subsection{Case Expansion and Decomposition}
\label{sub:decomposition}

The next stage separates two structural reductions with complementary counting semantics.
\textbf{Expand} (\cref{line:expand}) partitions the solution space into disjoint branches whose counts are summed, whereas \textbf{split} (\cref{line:decompose}) factors each branch into independent components whose counts are multiplied.

Because the encoding rules accept only conjunctions of atomic constraints (including their supported negations), compound Boolean constraints are first resolved through \emph{Shannon expansion}.
Counting disjuncts separately would double-count solutions satisfying multiple disjuncts.
Shannon expansion instead treats the conjunction of current constraints as a propositional formula with one Boolean variable per atomic constraint, and invokes a SAT solver to enumerate all complete satisfying truth assignments.
Each assignment yields a conjunction of positive and negative atomic constraints, guaranteeing that distinct branches share no solutions, although a propositionally satisfiable branch may still be semantically unsatisfiable and contribute zero.
For example, consider
\begin{cofolacode}{}
members = set(Alex, Bob) + set(member0...18)
officers = choose(members, 3)
(Bob not in officers) or (Alex not in officers)
\end{cofolacode}
\noindent whose constraint is $\alpha_1 \vee \alpha_2$, where $\alpha_1 \equiv (Bob \notin v_{officers})$ and $\alpha_2 \equiv (Alex \notin v_{officers})$.
Its three satisfying assignments yield the disjoint branches $\alpha_1 \wedge \alpha_2$, $\alpha_1 \wedge \neg\alpha_2$, and $\neg\alpha_1 \wedge \alpha_2$, so committees containing neither Alex nor Bob are counted only in the first branch.
Enumerating complete truth assignments can generate exponentially many branches in the number of atoms in a compound constraint; a disjoint decomposition that leaves redundant atoms unassigned, such as $\alpha_1\vee(\neg\alpha_1\wedge\alpha_2)$, would yield fewer branches and remains a natural refinement.

Within each Boolean branch, \emph{size-range expansion} fixes the lengths of tuples and sequences whose exact lengths are unknown, since their encodings require fixed lengths to determine position domains and length-dependent decoding factors.
For each such ordered object $v$ with a finite upper bound $\maxsize(v)$, the compiler branches on $|v|=k$ for every $k\in\{0,1,\ldots,\maxsize(v)\}$, creating mutually exclusive and exhaustive cases.
When $v$ is a full arrangement (\code{tuple(u)} or \code{sequence(u)}) of a dynamic or derived collection $u$ whose constructor does not fix its size, such as \code{supp(B)} or a set union, the length-$k$ branch explicitly adds the size constraint \code{|u| == k} so that the branch guard is enforced during encoding.
Bounds are derived from the constraints of the current branch and expanded one object at a time.
Subsequent local preprocessing (\cref{line:local_preprocess}) prunes infeasible combinations and propagates the fixed sizes.
Since the number of generated branches can reach the product of the enumerated ranges, tighter upper bounds directly reduce the number of \wfomc{} calls.

Within each resulting branch, \textbf{split} decomposes the problem into \emph{independent sub-problems} by forming an undirected graph that links each object to its dependencies and links all objects appearing in the same constraint.
Distinct connected components share neither object choices nor constraints, so their solution spaces form a Cartesian product, justifying multiplying the \emph{decoded counts} in \cref{line:solve}.
For example, the program
\begin{cofolacode}{}
interested_girls = set(girl0...6)
interested_boys  = set(boy0...11)
trip_girls = choose(interested_girls, 3)
trip_boys  = choose(interested_boys, 5)
\end{cofolacode}
\noindent splits into two independent components with total count $\binom{6}{3}\binom{11}{5}$.

Combining expansion and decomposition expresses the program's count as a sum over disjoint branches of products over independent components.
Let $B_1,\ldots,B_m$ be the disjoint branches covering all solutions, and let $Q_{i1},\ldots,Q_{ik_i}$ be the independent components of branch $B_i$.
Writing $\mathsf{Sol}(Q)$ for the set of satisfying combinatorial structures of $Q$, we have
\begin{equation}
    \comcount{D,O,C}
    =\sum_{i=1}^{m}\prod_{j=1}^{k_i}\#\mathsf{Sol}(Q_{ij}).
    \label{eq:compiler_sum_product}
\end{equation}

\subsection{Encoding, Counting, and Decoding}
\label{sub:encoding}

For each preprocessed component, $\encode$ traverses its declarations and constraints in dependency order and applies the rules of \cref{sec:encoding_rules} to produce $(\sentence,\domain,\weight,\negweight,A,\overcount,\Sigma)$.
Here $(\sentence,\domain,\weight,\negweight)$ specifies the symbolic \wfomc{} instance, whose evaluation returns a polynomial $f$.
The decoder then uses the exponent constraints in $A$ to select valid monomials and applies the redundancy corrections specified by $\overcount$ and $\Sigma$.

We illustrate decoding on \cref{ex:cardinality}, where the model counter returns $f(u)=(1+u)^3=1+3u+3u^2+u^3$.
The exponent of $u$ tracks the number of red balls, and its coefficient counts colourings with that many red balls.
Requiring at most two red balls selects the terms of degree $0$, $1$, and $2$ and sums their coefficients to obtain $1+3+3=7$.
In that example each colouring has a unique logical representation.
When multiple logical representations correspond to a single combinatorial structure, the encoder additionally supplies $\overcount$ and $\Sigma$ so that the decoder corrects this overcounting, as detailed in \cref{sub:enc_sequences,sub:enc_partitions}.
\cref{sec:count_preservation} proves that this encoding--decoding pipeline preserves the exact combinatorial count of every supported component.

\section{Encoding Rules}
\label{sec:encoding_rules}

\begin{table}[tbp]
\centering
\small
\caption{Encodings of set operations and constraints. Predicate $S_v$ represents set $v$, and $B_u$ represents the support of bag $u$. Free variable $x$ is universally quantified.}
\label{tab:set_encoding_rules}
\begin{tabular}{@{}ll@{}}
\toprule
Cofola declaration or constraint & Logical encoding \\
\midrule
\code{v = u1 + u2} & $S_v(x)\leftrightarrow(S_{u_1}(x)\lor S_{u_2}(x))$ \\
\code{v = u1 \& u2} & $S_v(x)\leftrightarrow(S_{u_1}(x)\land S_{u_2}(x))$ \\
\code{v = u1 - u2} & $S_v(x)\leftrightarrow(S_{u_1}(x)\land\neg S_{u_2}(x))$ \\
\code{v = supp(u)} & $S_v(x)\leftrightarrow B_u(x)$ \\
\midrule
\code{e in v} & $S_v(e)$ \\
\code{u1 subset u2} & $S_{u_1}(x)\to S_{u_2}(x)$ \\
\code{u1 disjoint u2} & $\neg(S_{u_1}(x)\land S_{u_2}(x))$ \\
\code{u1 == u2} & $S_{u_1}(x)\leftrightarrow S_{u_2}(x)$ \\
\bottomrule
\end{tabular}
\end{table}

This section presents the encoding rules by object type for components produced by preprocessing, expansion, and decomposition.
Constraints specific to a single object type, such as sequence patterns or bag multiplicity comparisons, are encoded alongside the objects they restrict.
Size constraints, whose atoms may span multiple object types, are encoded at the end of the encoding phase together with the decoding formula (\cref{sub:enc_size_constraints}).

For readability, we state each rule using \ourlang{} syntax, writing \code{v} and \code{e} in program syntax alongside $v$ and $e$ in mathematical notation.
For any entity $e$, $U_e/1$ denotes its identifying predicate fixed by the closed-world evidence $\closedevidence{U_e}{\domain}{\{e\}}$ (\cref{eq:closed_world_evidence}).

\subsection{Sets and Bags}
\label{sub:enc_sets}
\label{sub:enc_bags}
\label{sub:enc_bags_detail}

Sets are represented by unary predicates.
For an initialized set \code{v = set(e1, ..., en)}, the encoder introduces $S_v/1$ and adds the closed-world evidence $\closedevidence{S_v}{\domain}{\{e_1,\ldots,e_n\}}$, completely fixing its interpretation.
For \code{v = choose(u)}, it introduces $S_v/1$, adds $\forall x:S_v(x)\to S_u(x)$ to $\sentence$, and adds the cardinality constraint $|S_v|=k$ when $\exactsize(v)=k$ is known.
Set union, intersection, difference, support extraction, and set constraints are translated into first-order formulas relating the corresponding predicates, as summarized in \cref{tab:set_encoding_rules}.
Size constraints are deferred to \cref{sub:enc_size_constraints} because a single linear comparison may mix atoms from different object types.

Bags reuse the unary-predicate encoding for their supports and additionally track entity multiplicities through \emph{symbolic weights}.
For a bag $v$ with potential entity set $\pent(v)$ and inferred upper bound $\mult_v(e)$ on each entity's multiplicity, we introduce a unary support predicate $B_v/1$ together with a predicate $B_{v,e}/1$ and a fresh symbolic indeterminate $x_{v,e}$ for each $e\in\pent(v)$, setting
\[
    \forall x:B_{v,e}(x)\leftrightarrow(B_v(x)\land U_e(x)),
    \qquad
    \weight(B_{v,e})=\sum_{r=1}^{\mult_v(e)}x_{v,e}^{r},
    \qquad
    \negweight(B_{v,e})=1.
\]
The resulting \wfomc{} is a polynomial in the indeterminates $x_{v,e}$ whose exponent on $x_{v,e}$ records the multiplicity of $e$ in $v$.
Multiplicity conditions are therefore enforced by reading off and constraining these exponents in $A$; \cref{lem:coefficients} proves that this representation captures bag multiplicities faithfully.

An initialized bag \code{v = bag(e1: k1, ..., en: kn)} introduces $B_v/1$ with support fixed by $\closedevidence{B_v}{\domain}{\{e_i\mid k_i>0\}}$.
Because each multiplicity $k_i$ is fixed by the initializer, no symbolic multiplicity variables are created.
Throughout this section, $\mu_{v,e}$ denotes the multiplicity expression for $e$ in $v$, namely the constant $k_i$ for an initialized bag or the symbolic variable $x_{v,e}$ for a derived bag.

For \code{v = choose(u)}, we introduce $B_v/1$ with $\forall x:B_v(x)\to B_u(x)$ in $\sentence$, attach the weighted predicates $B_{v,e}$ above, and add $\mu_{v,e}\leq\mu_{u,e}$ to $A$ for every $e\in\pent(v)$ so that the selected multiplicity of $e$ never exceeds its source multiplicity.
When $\exactsize(v)=k$ is known, we also add $\sum_{e\in\pent(v)}\mu_{v,e}=k$ to $A$.
For \code{v = choose_replace(u, k)}, the source $u$ is a set represented by $S_u$.
We use the same weighted predicates for $v$, add $\forall x:B_v(x)\to S_u(x)$ to $\sentence$, allow each entity in $\pent(v)$ up to $k$ copies, and enforce $\sum_{e\in\pent(v)}\mu_{v,e}=k$ in $A$.
The unsized form \code{v = choose_replace(u)} relies on the inferred finite multiplicity bounds and adds a total-size equation when an exact size is known.

Deterministic bag operations combine support formulas with element-wise exponent equations, summarized in \cref{tab:bag_encoding_rules}.

\begin{table}[tbp]
\small
\centering
\caption{Encodings of bag operations. Predicate $B_v$ represents the support of bag $v$. Free $x$ is universally quantified. Each multiplicity equation is an exponent condition for every $e\in\pent(v)$, with local abbreviations $\mu=\mu_{v,e}$ and $\mu_i=\mu_{u_i,e}$.}
\label{tab:bag_encoding_rules}
\begin{tabular}{@{}lll@{}}
\toprule
Cofola declaration & Support condition & Multiplicity condition \\
\midrule
\code{v = u1 + u2} & $B_v(x)\leftrightarrow(B_{u_1}(x)\lor B_{u_2}(x))$ & $\mu=\max(\mu_1,\mu_2)$ \\
\code{v = u1 \& u2} & $B_v(x)\leftrightarrow(B_{u_1}(x)\land B_{u_2}(x))$ & $\mu=\min(\mu_1,\mu_2)$ \\
\code{v = u1 - u2} & $B_v(x)\to B_{u_1}(x)$ & $\mu=\max(\mu_1-\mu_2,0)$ \\
\code{v = u1 ++ u2} & $B_v(x)\leftrightarrow(B_{u_1}(x)\lor B_{u_2}(x))$ & $\mu=\mu_1+\mu_2$ \\
\bottomrule
\end{tabular}
\end{table}

\begin{example}\label{ex:bag_encoding}
Consider \code{B = bag(a: 2, b: 2)} followed by \code{C = choose(B, 2)}.
Abbreviating the multiplicity indeterminates by $x_a=x_{C,a}$ and $x_b=x_{C,b}$, the encoding yields the \wfomc{} polynomial $f=(1+x_a+x_a^2)(1+x_b+x_b^2)$.
The size condition $A=\{x_a + x_b = 2\}$ retains only the terms $x_a^2+x_ax_b+x_b^2$, corresponding to $\multiset{a,a}$, $\multiset{a,b}$, and $\multiset{b,b}$, respectively.
Each accepted term has coefficient $1$, so summing their coefficients returns $3$.
\end{example}

Bag constraints likewise combine support formulas with exponent conditions where multiplicities matter:
\begin{itemize}
    \item \code{e in v} $\cofolato B_v(e)$.
    \item \code{u1 subset u2} $\cofolato \forall x:B_{u_1}(x)\to B_{u_2}(x)$ and $\mu_{u_1,e}\leq\mu_{u_2,e}$ for every $e\in\pent(u_1)\cap\pent(u_2)$.
    \item \code{u1 disjoint u2} $\cofolato \forall x:\neg(B_{u_1}(x)\land B_{u_2}(x))$.
    \item \code{u1 == u2} $\cofolato \forall x:B_{u_1}(x)\leftrightarrow B_{u_2}(x)$ and $\mu_{u_1,e}=\mu_{u_2,e}$ for every $e\in\pent(u_1)\cap\pent(u_2)$.
\end{itemize}
Negated set constraints (such as \code{e not in v} or \code{u1 != u2}) negate the corresponding first-order formulas in \cref{tab:set_encoding_rules}.
For bags, \code{e not in v} and negated disjointness similarly negate their support formulas, whereas negated containment or equality requires at least one entity $e\in\pent(u_1)\cup\pent(u_2)$ to violate the corresponding multiplicity relation in $A$.

\subsection{Tuples}
\label{sub:enc_tuples}

Tuple encodings rely on functionality axioms of the form $Func(F, Dom, CoDom)$, where $F$ is a binary predicate and $Dom$ and $CoDom$ are unary predicates, asserting that $F$ defines a total function from $Dom$ to $CoDom$.
Although the standard first-order definition of functionality requires three variables, the following proposition gives an equivalent formulation in $\ctwo{}$ using a cardinality constraint (proved in \cref{app:proof_func_axiom}).
\begin{proposition}\label{prop:func_axiom}
    Let $Func(F, Dom, CoDom)$ be a functionality axiom, and define
    \begin{equation}
        \Psi = \left(\forall x\exists y: Dom(x) \to F(x,y)\right) \land \left(\forall x\forall y: F(x,y) \to Dom(x)\land CoDom(y)\right) \land \left(|F| - |Dom| = 0\right).
    \end{equation}
    Then for any \ctwo{} sentence $\Phi$ (possibly with additional axioms and evidence), an interpretation satisfies $\Phi \land Func(F, Dom, CoDom)$ if and only if it satisfies $\Phi \land \Psi$.
\end{proposition}
We write $F^{-1}/2$ for the inverse relation of $F$ and $Img_F/1$ for its image predicate.
We further write $Func_{inj}$, $Func_{surj}$, and $Func_{bij}$ for functionality axioms strengthened with injectivity, surjectivity, and bijectivity, respectively; all three admit $\ctwo{}$ encodings with cardinality constraints, detailed in \cref{app:function_relations}.
Using these axioms, a tuple is encoded as a function from a fixed set of position constants to its source collection, with set and bag sources requiring different functionality constraints (\cref{tab:tuple_func_encoding}).

\begin{table}[tbp]
    \centering
    \caption{Encoding tuple $T$ as a function from its position set to a set source or a bag's support. The first column gives the value of source $u$; the conditions use its predicate $S_u$ or $B_u$.}
    \label{tab:tuple_func_encoding}
    \small
    \setlength{\tabcolsep}{10pt}
    \renewcommand{\arraystretch}{1.3}
        \begin{tabular}{@{}lll@{}}
            \toprule
            Source value & Example function $F_T$ & Encoding condition \\
            \midrule
            \addlinespace[2pt]
            $u=\{a,b,c\}$ &
            \begin{tikzpicture}[baseline=-0.55ex, >=Latex, font=\tiny, setbox/.style={draw=black!55, rounded corners=2pt, fill=gray!3}, point/.style={circle, fill=black, inner sep=1.05pt}, map/.style={->, thick, draw=figurenavy}]
                \node[setbox, minimum width=0.95cm, minimum height=1.15cm] at (1.65,0) {};
                \node[font=\tiny\bfseries] at (-1.1,0.42) {$I_T$};
                \node[font=\tiny\bfseries] at (1.65,0.42) {$S_u$};
                \node[point,label=left:$idx_{T,1}$] (ti1) at (-1.1,0.25) {};
                \node[point,label=left:$idx_{T,2}$] (ti2) at (-1.1,0) {};
                \node[point,label=left:$idx_{T,3}$] (ti3) at (-1.1,-0.25) {};
                \node[point,label=right:$a$] (ta) at (1.65,0.25) {};
                \node[point,label=right:$b$] (tb) at (1.65,0) {};
                \node[point,label=right:$c$] (tc) at (1.65,-0.25) {};
                \draw[map] (ti1) -- (tb);
                \draw[map] (ti2) -- (tc);
                \draw[map] (ti3) -- (ta);
            \end{tikzpicture} &
            \makecell[l]{$Func_{bij}(F_T,I_T,S_u)$} \\
            \midrule
            \addlinespace[2pt]
            $u=\multiset{a{:}2,b{:}1}$ &
            \begin{tikzpicture}[baseline=-0.55ex, >=Latex, font=\tiny, setbox/.style={draw=black!55, rounded corners=2pt, fill=gray!3}, point/.style={circle, fill=black, inner sep=1.05pt}, map/.style={->, thick, draw=figurenavy}]
                \node[setbox, minimum width=0.95cm, minimum height=1.15cm] at (1.65,0) {};
                \node[font=\tiny\bfseries] at (-1.12,0.42) {$I_T$};
                \node[font=\tiny\bfseries] at (1.7,0.42) {$B_u$};
                \node[point,label=left:$idx_{T,1}$] (bi1) at (-1.15,0.25) {};
                \node[point,label=left:$idx_{T,2}$] (bi2) at (-1.15,0) {};
                \node[point,label=left:$idx_{T,3}$] (bi3) at (-1.15,-0.25) {};
                \node[point,label=right:$a$] (ba) at (1.75,0.2) {};
                \node[point,label=right:$b$] (bb) at (1.75,-0.2) {};
                \draw[map] (bi1) -- (ba);
                \draw[map] (bi2) -- (ba);
                \draw[map] (bi3) -- (bb);
            \end{tikzpicture} &
            \makecell[l]{$Func(F_T,I_T,B_u)$\\ $|(F_T^{-1})_a|=2,\ |(F_T^{-1})_b|=1$} \\
            \bottomrule
        \end{tabular}
\end{table}

Consider \code{T = tuple(u)} with a set source $u$, where the length $k=\exactsize(T)$ is fixed after size-range expansion (\cref{sub:decomposition}).
We allocate $k$ fresh position constants $idx_{T,1},\ldots,idx_{T,k}$ in $\domain$ and fix the position predicate $I_T/1$ via closed-world evidence $\closedevidence{I_T}{\domain}{\{idx_{T,1},\ldots,idx_{T,k}\}}$.\footnote{All position constants for a component are allocated before encoding its objects, so the unary evidence for sets and bags is interpreted over the augmented domain and excludes these fresh constants from their supports.}
The axiom $Func_{bij}(F_T,I_T,S_u)$ then represents the tuple as a bijection from positions to the source set.
Analogously, a direct selection \code{T = choose_tuple(u, k)} from a set can be encoded with $Func_{inj}(F_T,I_T,S_u)$ without introducing an intermediate chosen set.

When the source $u$ is a bag, we use the same position set $I_T$ together with $Func(F_T,I_T,B_u)$.
To match the multiplicities of the source bag, we introduce a unary fibre predicate $(F_T^{-1})_e/1$ for each $e\in\pent(u)$ via
\[
    \forall x:(F_T^{-1})_e(x)\leftrightarrow
        \exists y\,(F_T(x,y)\land U_e(y)),
\]
assign $\weight((F_T^{-1})_e)=x^{\mathrm{occ}}_{T,e}$, and add $x^{\mathrm{occ}}_{T,e}=\mu_{u,e}$ to $A$.
This ensures that the number of tuple positions mapped to $e$ equals its multiplicity in $u$.

Membership \code{e in T} is encoded directly via the image predicate as $Img_{F_T}(e)$ (and negated as $\neg Img_{F_T}(e)$).
Because \ourlang{} indexes tuple positions from zero, position $i\in[k]_0$ corresponds to $idx_{T,i+1}$.
Although \code{T[i] == e} corresponds semantically to the binary ground literal $F_T(idx_{T,i+1},e)$, our lifted solvers restrict evidence to unary predicates (\cref{sub:wfomc}).
We therefore introduce $Idx_{T,i+1}/1$ fixed by $\closedevidence{Idx_{T,i+1}}{\domain}{\{idx_{T,i+1}\}}$ and add
\[
    \forall x\forall y:(Idx_{T,i+1}(x)\land F_T(x,y))\to U_e(y).
\]
For \code{T[i] in A}, we replace $U_e(y)$ with the support predicate $S_A(y)$, and negated positional constraints negate the corresponding formula.
Within a fixed-length-$k$ branch, any positive positional assertion with $i\geq k$ evaluates directly to false (pruning the branch during preprocessing, \cref{app:sanity_checks}), whereas its negation (\code{T[i] != e} or \code{T[i] not in A}) is vacuously satisfied and simplifies to true without adding a formula.

\begin{example}\label{ex:tuple_encoding}
Consider the \ourlang{} program in \cref{ex:fourletter}.
Omitting the occurrence-count constraint for now, the encoded sentence is
\begin{align*}
    \sentence={}&Func_{inj}(F_\code{arr},I_\code{arr},S_\code{letters})\\
    &{}\land\bigl(\forall x\forall y:
         (Idx_{\code{arr},1}(x)\land F_\code{arr}(x,y))\to U_C(y)\bigr)\\
    &{}\land\closedevidence{S_\code{letters}}{\domain}{\{A,B,C,D,E,F\}}\\
    &{}\land\closedevidence{I_\code{arr}}{\domain}{\{idx_{\code{arr},1},\ldots,idx_{\code{arr},4}\}}\\
    &{}\land\closedevidence{Idx_{\code{arr},1}}{\domain}{\{idx_{\code{arr},1}\}}
      \land\closedevidence{U_C}{\domain}{\{C\}},
\end{align*}
over $\domain=\{A,B,C,D,E,F,idx_{\code{arr},1},\ldots,idx_{\code{arr},4}\}$.
\end{example}

\Needspace{6\baselineskip}
\subsection{Sequences}
\label{sub:enc_sequences}

Sequences differ from tuples as they support relative-order and adjacency constraints, which we encode using the linear-order axiom $Linear(L,Pred)$ of \cref{sub:wfomc_order_axioms}.

For \code{Q = sequence(u)} over a set source $S_u/1$ of known size $k$, we add $Linear(L,Pred)$ to $\sentence$, where $L/2$ is a reflexive total order over the entire domain $\domain$ and $Pred/2$ is its immediate-successor relation.
Because $L$ orders all domain elements rather than only the source set, we add
\begin{equation*}
    \forall x\forall y:\; (S_u(x)\land\neg S_u(y))\to L(x,y)
\end{equation*}
so that elements of the source precede all elements outside it.
Each concrete sequence of the $k$ selected elements then admits $(|\domain|-k)!$ logical models corresponding to orderings of the remaining domain elements, so we multiply $\overcount$ by $(|\domain|-k)!$.
When $u$ is a dynamic or derived set whose definition does not itself fix cardinality (such as \code{supp(B)} or a set union), the length-$k$ expansion branch retains the explicit constraint \code{|u| == k}, which translates into $x_u=k$ in $A$ (\cref{sub:enc_size_constraints}) so that every accepted model has exactly $k$ elements in $S_u$ and $\overcount$ remains exact.

For a bag source $u$, a linear order on the entity domain cannot represent repeated occurrences directly.
Instead, we allocate $k$ fresh position constants $idx_{Q,1},\ldots,idx_{Q,k}$ fixed by $I_Q/1$, add $Linear(L,Pred)$, and enforce the prefix condition above with $I_Q$ in place of $S_u$ so that the positions precede all other domain elements.
To assign each position an entity from the bag, we introduce $Pos_{Q,e}/1$ for each $e\in\pent(u)$ and add
\begin{equation*}
    \left(\forall x:I_Q(x)\leftrightarrow\bigvee_{e\in\pent(u)}Pos_{Q,e}(x)\right)
    \land\bigwedge_{e\in\pent(u)}
       \left(\forall x:Pos_{Q,e}(x)\to
          \bigwedge_{e'\in\pent(u)\setminus\{e\}}\neg Pos_{Q,e'}(x)\right).
\end{equation*}
Here $Pos_{Q,e}$ labels positions, whereas $U_e$ identifies the entity in $\domain$.
We then set $\weight(Pos_{Q,e})=x^{\mathrm{occ}}_{Q,e}$ and add $x^{\mathrm{occ}}_{Q,e}=\mu_{u,e}$ to $A$ so that each entity labels exactly as many positions as its multiplicity in $u$.
Finally, we multiply $\overcount$ by $k!\cdot(|\domain|-k)!$ to account for permutations of the named position constants and orderings of the remaining domain elements, respectively.
\cref{fig:seq_from_bag_encoding} illustrates this position-based encoding for a four-element bag.

\begin{figure}[tbp]
\centering
\begin{tikzpicture}[
    >=Latex,
    font=\small,
    idxnode/.style={
        circle,
        draw=black!65,
        fill=gray!5,
        minimum size=8mm,
        inner sep=0pt
    },
    box/.style={
        draw=black!60,
        rounded corners=3pt,
        fill=gray!3,
        inner sep=4pt
    },
    lab/.style={
        draw=black!45,
        rounded corners=2pt,
        fill=white,
        inner sep=2pt
    },
    order/.style={->, thick, draw=figurenavy},
    assign/.style={->, thick, draw=black!70},
    brace/.style={decorate, decoration={brace, amplitude=5pt}}
]

\node[box] (bag) at (-3.7,-1.35)
    {$u=\multiset{a{:}2,\ b{:}1,\ c{:}1}$};

\node[lab] (ea) at (-4.6,-0.1) {$a$};
\node[lab] (eb) at (-3.7,-0.1) {$b$};
\node[lab] (ec) at (-2.8,-0.1) {$c$};

\node[idxnode] (i1) at (0,0) {$idx_{Q,1}$};
\node[idxnode] (i2) at (1.7,0) {$idx_{Q,2}$};
\node[idxnode] (i3) at (3.4,0) {$idx_{Q,3}$};
\node[idxnode] (i4) at (5.1,0) {$idx_{Q,4}$};

\draw[order] (i1) -- node[above, font=\scriptsize] {$Pred$} (i2);
\draw[order] (i2) -- node[above, font=\scriptsize] {$Pred$} (i3);
\draw[order] (i3) -- node[above, font=\scriptsize] {$Pred$} (i4);

\draw[brace] (-0.35,0.55) -- (5.45,0.55)
    node[midway, above=5pt, font=\footnotesize]
    {linear order $Linear(L,Pred)$ on positions};

\node[lab] (l0) at (0,-1.35) {$Pos_{Q,a}$};
\node[lab] (l1) at (1.7,-1.35) {$Pos_{Q,b}$};
\node[lab] (l2) at (3.4,-1.35) {$Pos_{Q,a}$};
\node[lab] (l3) at (5.1,-1.35) {$Pos_{Q,c}$};

\draw[assign] (i1) -- (l0);
\draw[assign] (i2) -- (l1);
\draw[assign] (i3) -- (l2);
\draw[assign] (i4) -- (l3);

\draw[assign, dashed] (ea) to[out=-20,in=-160] (i1);
\draw[assign, dashed] (ea) to[out=-25,in=-160] (i3);
\draw[assign, dashed] (eb) to[out=-20,in=-160] (i2);
\draw[assign, dashed] (ec) to[out=-25,in=-160] (i4);

\node[font=\footnotesize, align=center] at (-1.7,0.3)
    {encoded sequence\\$[a,b,a,c]$};

\end{tikzpicture}
\caption{Encoding sequence $Q$ from bag $u$ by ordering fresh positions and assigning an entity label to each position. Each predicate label is true at the position above it.}
\label{fig:seq_from_bag_encoding}
\end{figure}

Sequence constraints encompass global and local patterns, both of which are encoded uniformly once matching positions are identified.
For any pattern argument $A$, let $R_A=S_A$ when $A$ is a set and $R_A=U_e$ when $A$ is an entity $e$.
For a set-sourced sequence with source predicate $S_u/1$, we introduce $M_{Q,A}/1$ via $\forall x:M_{Q,A}(x)\leftrightarrow(S_u(x)\land R_A(x))$ to mark matching occurrences of $A$ in $Q$, defining $M_{Q,B}$ analogously.

To encode \code{together(A) in Q}, we introduce $First_{Q,A}/1$ to mark the starting element of every contiguous block of $M_{Q,A}$,
\begin{equation*}
    \forall x:First_{Q,A}(x)\leftrightarrow
       \left(M_{Q,A}(x)\land\neg\exists y\,(M_{Q,A}(y)\land Pred(y,x))\right),
\end{equation*}
set $\weight(First_{Q,A})=x^{\mathrm{blk}}_{Q,A}$, and add $x^{\mathrm{blk}}_{Q,A}\leq 1$ to $A$ to allow at most one block; its negation \code{together(A) not in Q} reuses $First_{Q,A}$ and instead adds $x^{\mathrm{blk}}_{Q,A}>1$ to $A$, requiring at least two separated blocks.
To encode \code{(A < B) in Q}, we require every occurrence of $A$ to strictly precede every occurrence of $B$,
\begin{equation*}
    \forall x\forall y:(M_{Q,A}(x)\land M_{Q,B}(y))\to(L(x,y)\land x\ne y),
\end{equation*}
and \code{(A < B) not in Q} negates this formula.

For a local pattern \code{p}, let $Adj(x,y)$ abbreviate $Pred(x,y)$ for the directed pattern \code{(A,B)} and $Pred(x,y)\lor Pred(y,x)$ for the undirected pattern \code{next_to(A,B)}.
Then \code{p in Q} is encoded as
\begin{equation*}
    \exists x\exists y:M_{Q,A}(x)\land M_{Q,B}(y)\land Adj(x,y),
\end{equation*}
and \code{p not in Q} negates this formula.
Counting matches of a local pattern reuses these matching predicates and is described in \cref{sub:enc_size_constraints}.

When the sequence has a bag source $u$, matching is defined on position labels rather than directly on domain entities,
\[
    \forall x:M_{Q,A}(x)\leftrightarrow
       \left(I_Q(x)\land\bigvee_{e\in\pent(u)}
          \left(Pos_{Q,e}(x)\land\exists y\,(U_e(y)\land R_A(y))\right)\right).
\]
The same block, precedence, and adjacency formulas then apply unchanged over the ordered positions.

\begin{example}\label{ex:sequence_encoding}
Consider the \ourlang{} program in \cref{ex:books}, writing $Q=\code{arr}$, $u=\code{books}$, $m=\code{math\_books}$, and $h=\code{physics\_books}$.
The compiled sentence is
\begin{align*}
    \sentence={}&Linear(L,Pred)\land\evidence\\
    &{}\land\bigl(\forall x\forall y:(S_u(x)\land\neg S_u(y))\to L(x,y)\bigr)\\
    &{}\land\bigl(\forall x:M_{Q,m}(x)\leftrightarrow(S_u(x)\land S_m(x))\bigr)\\
    &{}\land\bigl(\forall x:M_{Q,h}(x)\leftrightarrow(S_u(x)\land S_h(x))\bigr)\\
    &{}\land\bigl(\forall x:First_{Q,m}(x)\leftrightarrow
         (M_{Q,m}(x)\land\neg\exists y\,(M_{Q,m}(y)\land Pred(y,x)))\bigr)\\
    &{}\land\bigl(\forall x\forall y:(M_{Q,m}(x)\land M_{Q,h}(y))
         \to(L(x,y)\land x\ne y)\bigr),
\end{align*}
over $\domain=\{\mathrm{math1},\mathrm{math2},\mathrm{physics},\mathrm{book4},\mathrm{book5},\mathrm{book6},\mathrm{book7}\}$, with
\[
    \evidence=\closedevidence{S_u}{\domain}{\domain}
       \land\closedevidence{S_m}{\domain}{\{\mathrm{math1},\mathrm{math2}\}}
       \land\closedevidence{S_h}{\domain}{\{\mathrm{physics}\}}.
\]
Setting $\weight(First_{Q,m})=x^{\mathrm{blk}}_{Q,m}$ and $A=\{x^{\mathrm{blk}}_{Q,m}\leq 1\}$ enforces contiguity; since $|S_u|=|\domain|=7$, the overcounting factor is $\overcount=(7-7)!=1$.
\end{example}

\subsection{Partitions and Compositions}
\label{sub:enc_partitions}

Partitions and compositions introduce predicates for their $k$ parts and enforce coverage and disjointness (or multiplicity conservation).
Because the parts of a partition are unordered, partition encodings additionally break label symmetry and account for any remaining overcounting.

For a composition \code{G = compose(u, k)}, let $g_1,\ldots,g_k$ denote its labelled part objects, with support predicates $R_{g_i}=S_{g_i}$ for a set source and $R_{g_i}=B_{g_i}$ for a bag source.
Using the corresponding source predicate $R_u$, we enforce coverage by adding
\begin{equation*}
    \forall x:R_u(x)\leftrightarrow\bigvee_{i=1}^{k}R_{g_i}(x).
\end{equation*}
For a set source, we also require pairwise disjointness via
\begin{equation*}
    \bigwedge_{1\leq i<j\leq k}\forall x:\neg(S_{g_i}(x)\land S_{g_j}(x)).
\end{equation*}
For a bag source, entities may be split across parts with multiplicities, so coverage alone does not suffice.
We therefore attach the bag encoding $\forall x:B_{g_i,e}(x)\leftrightarrow(B_{g_i}(x)\land U_e(x))$ with $\weight(B_{g_i,e})=\sum_{r=1}^{\mult_{g_i}(e)}x_{g_i,e}^{r}$ to each part $g_i$ and $e\in\pent(g_i)$, and add $\sum_{i=1}^{k}\mu_{g_i,e}=\mu_{u,e}$ to $A$ for every $e\in\pent(u)$ to conserve multiplicities.

\begin{table}[tbp]
\centering
\small
\setlength{\tabcolsep}{3pt}
\renewcommand{\arraystretch}{1.12}
\begin{minipage}[t]{0.48\linewidth}
\centering
\caption{Encoding of the partitions of $\{a, b, c, d\}$ into $2$ parts. Nondecreasing part sizes break some symmetries, while the two labellings of each equally sized positive-part partition in the $2|2$ row are corrected by dividing by $2!$ during decoding.}
\label{tab:set_partition_symmetry_encoding}
\begin{tabular}{@{}p{0.2\linewidth}p{0.5\linewidth}@{}}
\toprule
\textbf{part sizes} & \textbf{ordered candidates} \\
\midrule
$0|4$ & $\emptyset\,|\,abcd$ \\
\midrule
$1|3$ & \makecell[l]{$a\,|\,bcd,\quad b\,|\,acd$\\$c\,|\,abd,\quad d\,|\,abc$} \\
\midrule
$2|2$ & \makecell[l]{$ab\,|\,cd \Leftrightarrow cd\,|\,ab$\\$ac\,|\,bd \Leftrightarrow bd\,|\,ac$\\$ad\,|\,bc \Leftrightarrow bc\,|\,ad$} \\
\bottomrule
\end{tabular}
\end{minipage}
\hfill
\begin{minipage}[t]{0.5\linewidth}
\centering
\caption{Encoding of the partitions of the bag $\multiset{a: 2, b: 2}$ into $2$ parts.
Each part signature lists the multiplicities of $(a,b)$ in that part; for instance, $(0,0)|(2,2)$ assigns no elements to the first part and all four to the second.
Non-strict lexicographic ordering on part signatures breaks all label symmetries.}
\label{tab:bag_partition_symmetry_encoding}
\begin{tabular}{@{}p{0.48\linewidth}p{0.45\linewidth}@{}}
\addlinespace[0.5em]
\toprule
\textbf{part signatures} & \textbf{ordered candidates} \\
\midrule
$(0,0)|(2,2)$ & $\emptyset\,|\,aabb$ \\
$(0,1)|(2,1)$ & $b\,|\,aab$ \\
$(0,2)|(2,0)$ & $bb\,|\,aa$ \\
$(1,0)|(1,2)$ & $a\,|\,abb$ \\
$(1,1)|(1,1)$ & $ab\,|\,ab$ \\
\bottomrule
\end{tabular}
\end{minipage}
\end{table}

For a set partition \code{G = partition(u, k)}, we introduce labelled part objects $g_1,\ldots,g_k$ with predicates $S_{g_1},\ldots,S_{g_k}$ and the same coverage and disjointness conditions as for a composition.
We assign each part a size indeterminate $x_{g_i}$ with $\weight(S_{g_i})=x_{g_i}$ and add $x_{g_1}\leq\cdots\leq x_{g_k}$ to $A$ to enforce nondecreasing part sizes.

Nondecreasing part sizes do not break symmetry among distinct nonempty parts of equal size.
In \cref{tab:set_partition_symmetry_encoding}, each partition in the $1|3$ row has a unique ordered labelling, whereas in the $2|2$ row both $\{a,b\}\mid\{c,d\}$ and $\{c,d\}\mid\{a,b\}$ satisfy the size ordering.
The decoder therefore keeps the $1|3$ count unchanged and divides the $2|2$ count by $2$.
In general, we record each set partition's variable tuple $(x_{g_1},\ldots,x_{g_k})$ in $\Sigma$.
During decoding, for each monomial with exponent vector $\mathbf{d}$, we inspect the recorded part sizes $(d_{g_1},\ldots,d_{g_k})$ and divide by $r!$ for every block of $r$ parts sharing the same \emph{positive} size (empty parts are identical sets and introduce no duplicate labellings).
Writing $q_\Sigma(\mathbf d)$ for the product of these factorial factors across all set partitions in $\Sigma$, we obtain $q_\Sigma(\mathbf d)=1$ for $1|3$ and $q_\Sigma(\mathbf d)=2$ for $2|2$.

For a bag partition \code{G = partition(u, k)}, the coverage and multiplicity-conservation rules match those of a bag composition.
To eliminate label symmetry completely, we define the \emph{signature} of each bag part $g_i$ as its multiplicity vector $\boldsymbol\mu_{g_i}=(\mu_{g_i,e})_{e\in\pent(u)}$ under a fixed entity order, and add the non-strict lexicographic chain
\[
    \boldsymbol\mu_{g_1}\leq_{\mathrm{lex}}\boldsymbol\mu_{g_2}
       \leq_{\mathrm{lex}}\cdots\leq_{\mathrm{lex}}\boldsymbol\mu_{g_k}
\]
to $A$, where $\mathbf a\leq_{\mathrm{lex}}\mathbf b$ holds if $\mathbf a=\mathbf b$ or the first coordinate at which they differ is smaller in $\mathbf a$.
Unlike the set case, identical part signatures correspond to identical bags, so the non-strict lexicographic chain selects a unique canonical ordered representative for every unordered bag partition and requires no quotient factor in the decoder (\cref{tab:bag_partition_symmetry_encoding}).

Finally, indexing a composition part via \code{v = G[i]} links support predicates by $\forall x:R_v(x)\leftrightarrow R_{g_{i+1}}(x)$ and, for bag-valued parts, reuses the part's multiplicity expressions without introducing a new choice.

\subsection{Size Constraints and Decoding}
\label{sub:enc_size_constraints}

Because part-wise size queries on partitions and compositions are reduced to set or bag cardinalities during preprocessing, the remaining size constraints involve only sets, bags, tuples, and sequences.
Consider a linear size constraint
\begin{equation*}
    \alpha_1a_1+\alpha_2a_2+\cdots+\alpha_ma_m\bowtie K,
\end{equation*}
where $a_1,\ldots,a_m$ are size atoms, $\alpha_1,\ldots,\alpha_m, K \in \mathbb{Z}$, and $\bowtie\in\{<,\leq,=,\ne,\geq,>\}$.
Each atom is translated into a linear expression over symbolic weight variables (summarized in \cref{tab:size_atom_encoding}), and the resulting comparison is added to $A$, with identical atoms within a component reusing the same variable.

For sets and bags, two kinds of size atoms occur.
\begin{itemize}
    \item \textbf{Cardinality \code{|v|}.} For a set, we set $\weight(S_v)=x_v$ (reusing $x_v$ if already assigned) and translate \code{|v|} to $x_v$. For a bag, we translate \code{|v|} to $\sum_{e\in\pent(v)}\mu_{v,e}$ using the multiplicity expressions from \cref{sub:enc_bags_detail}.
    \item \textbf{Bag multiplicity \code{v.count(e)}.} This atom translates directly to $\mu_{v,e}$.
\end{itemize}

For ordered objects, three kinds of size atoms occur, two for tuples (represented via $F_T$ and its image $Img_{F_T}$) and one for sequences.
\begin{itemize}
    \item \textbf{Position count \code{T.count(A)}.} Counts positions in tuple $T$ occupied by entities in $A$. We introduce $C^{\mathrm{cnt}}_{T,A}/1$ with $\weight(C^{\mathrm{cnt}}_{T,A})=x^{\mathrm{cnt}}_{T,A}$, defined by
    \[
        \forall x:C^{\mathrm{cnt}}_{T,A}(x)\leftrightarrow
           \exists y\,(F_T(x,y)\land R_A(y)),
    \]
    and translate the atom to $x^{\mathrm{cnt}}_{T,A}$.
    \item \textbf{Distinct-value count \code{T.dedup_count(A)}.} Counts distinct entities from $A$ occurring in tuple $T$. We introduce $C^{\mathrm{dist}}_{T,A}/1$ with $\weight(C^{\mathrm{dist}}_{T,A})=x^{\mathrm{dist}}_{T,A}$, defined by
    \[
        \forall x:C^{\mathrm{dist}}_{T,A}(x)\leftrightarrow
           (Img_{F_T}(x)\land R_A(x)),
    \]
    and translate the atom to $x^{\mathrm{dist}}_{T,A}$.
    \item \textbf{Pattern count \code{Q.count(p)}.} Counts adjacent position pairs matching local pattern $p$ in sequence $Q$. Using the matching predicates from \cref{sub:enc_sequences}, we introduce $C^{\mathrm{pat}}_{Q,p}/2$ with $\weight(C^{\mathrm{pat}}_{Q,p})=x^{\mathrm{pat}}_{Q,p}$ and translate the atom to $x^{\mathrm{pat}}_{Q,p}$. For the directed pattern \code{(A,B)}, we add
    \[
        \forall x\forall y:C^{\mathrm{pat}}_{Q,p}(x,y)\leftrightarrow
           (Pred(x,y)\land M_{Q,A}(x)\land M_{Q,B}(y)),
    \]
    and for \code{next_to(A,B)}, we add
    \[
        \forall x\forall y:C^{\mathrm{pat}}_{Q,p}(x,y)\leftrightarrow
          \left(Pred(x,y)\land
             \bigl((M_{Q,A}(x)\land M_{Q,B}(y))
                    \lor(M_{Q,B}(x)\land M_{Q,A}(y))\bigr)\right).
    \]
    In both cases, $Pred(x,y)$ fixes the pair orientation so that each matching adjacent pair is counted once even when $A$ and $B$ overlap.
\end{itemize}
All three auxiliary definitions use at most two logical variables, staying within the $\ctwo{}$ fragment required for domain-lifted inference.

Encoding all size constraints completes the compiled component $(f,A,\overcount,\Sigma)$, where $f=\sum_{\mathbf d}c_{\mathbf d}\mathbf{x}^{\mathbf d}$ is the \wfomc{} polynomial and $A$ collects all exponent conditions emitted during encoding, including linear comparisons, element-wise $\min$/$\max$ equations for bag operations, finite disjunctions for negated bag comparisons, and lexicographic chains for bag partitions.
Each exponent vector $\mathbf d$ assigns every symbolic variable $x$ its exponent $\mathbf d[x]$, and $\mathbf d\models A$ indicates that $\mathbf d$ satisfies every condition in $A$.
Combining the exponent filter $\mathbf d\models A$, the set-partition quotient $q_\Sigma(\mathbf d)$ from \cref{sub:enc_partitions}, and the sequence overcounting factor $\overcount$ from \cref{sub:enc_sequences} yields the component count
\begin{equation}
    \decode(f,A,\overcount,\Sigma)
    =\frac{1}{\overcount}
       \sum_{\mathbf d\models A}\frac{c_{\mathbf d}}{q_\Sigma(\mathbf d)}.
    \label{eq:decoder_coefficients}
\end{equation}
When $\Sigma$ is empty, $q_\Sigma(\mathbf d)=1$ for every term and \eqref{eq:decoder_coefficients} reduces to $[A]f/\overcount$.
Because \cref{alg:decode} (\cref{app:decoder_algorithm}) evaluates $A$ and accumulates exact rational quotients once per nonzero monomial in $f$, decoding preserves exact integer counts whenever $f$ is exact, and its runtime scales linearly with the number of nonzero monomials in $f$.

\begin{example}
\label{ex:mathcounts_size_encoding}
In \cref{ex:mathcounts}, four magnets are chosen from the letters of MATHCOUNTS with fewer vowels than consonants.
Writing $\mu_e$ for the multiplicity of letter $e$ in \code{chosen}, the two bag intersections preserve these multiplicities, giving $\mu_{\mathrm{vow}}=\mu_A+\mu_O+\mu_U$ and $\mu_{\mathrm{con}}=\mu_M+\mu_T+\mu_H+\mu_C+\mu_N+\mu_S$.
Thus \code{choose(magnets, 4)} adds $\mu_{\mathrm{vow}}+\mu_{\mathrm{con}}=4$ to $A$, and \code{|chosen_vowels| - |chosen_consonants| < 0} adds $\mu_{\mathrm{vow}}-\mu_{\mathrm{con}}<0$.
Because no sequences or set partitions appear ($\overcount=1$, $\Sigma=\emptyset$), \eqref{eq:decoder_coefficients} returns $\decode(f,A,1,\emptyset)=[A]f$.
\end{example}

\section{Count Preservation and Liftability}
\label{sec:count_preservation}

In this section, we establish two theoretical properties of the \ourlang{}-to-\wfomc{} compiler.
\cref{sub:representation_labelled} proves that the compiler preserves exact combinatorial counts on its supported fragment, and \cref{sub:liftability} characterizes when compiled families retain a fixed sentence so that domain-liftability applies, motivating the bag factorization developed in \cref{sec:bag_factorization}.

\subsection{Count Preservation}
\label{sub:representation_labelled}

Count preservation requires finite inferred bounds on the lengths and multiplicities ranged over by weight polynomials, together with at most one linear order per component.
The first condition excludes unbounded instances as well as cases where static analysis cannot certify a finite bound.
The second is forced by the encoding, because the \wfomc{} backend supports at most one linear order per component.\footnote{\citet{kuang2025weighted} proved that \wfomc{} with two linear orders is $\#\P_1$-complete (complete for counting problems with unary-encoded inputs), so this restriction is necessary for tractability.}

\begin{definition}[Supported fragment]
\label{def:supported_fragment}
A combinatorial counting problem is \emph{supported} if it is the semantics of a valid, well-typed \ourlang{} program with an encoding rule for every atomic object and constraint, and if it satisfies the following conditions.
\begin{itemize}
    \item Every tuple and sequence has a \emph{finite} inferred upper bound on its length, and every bag multiplicity represented by a polynomial weight has a \emph{finite} inferred upper bound.
    \item Every component produced by \textbf{split} contains \emph{at most one} sequence.
\end{itemize}
\end{definition}

Fix a preprocessed component $Q$ obtained from a supported program after case expansion and decomposition, with atomic constraints and fixed tuple and sequence lengths.
Let $\sem{Q}=\langle D,O,C\rangle$ be its semantics and let $rec$ be its analysis record, with sound bounds on labelled assignments before partition-ordering conditions are imposed.
Encoding $Q$ produces
\begin{equation*}
    \encode(Q,rec)=(\sentence,\domain,\weight,\negweight,A,\overcount,\Sigma).
\end{equation*}
Write $\mathsf{Sol}(Q)$ for its set of satisfying structures, so that $\comcount{D,O,C}=|\mathsf{Sol}(Q)|$.
As outlined in \cref{fig:proof_structure}, the proof factors the correspondence between polynomial weight choices and $\mathsf{Sol}(Q)$ through two maps, $\rho$ and $\lambda$, whose fibre sizes match the two decoder divisors $\overcount$ and $q_\Sigma(\mathbf d)$.

\begin{figure}[tbp]
\centering
\hypersetup{linkcolor=figurenavy}
\begin{tikzpicture}[
    >=Latex, font=\footnotesize,
    st/.style={draw=black!55, rounded corners=3pt, fill=gray!4, align=center, inner sep=4pt, minimum height=1.1cm, text width=3.3cm},
    flow/.style={->, thick, draw=figurenavy},
    note/.style={font=\scriptsize, align=center, text=black!75}
]
\node[st] (rep) {accepted weighted\\representations\\$\mathsf{Rep}_A(Q)$};
\node[st, right=2.4cm of rep] (lab) {labelled\\structures\\$\mathsf{Lab}(Q)$};
\node[st, right=2.4cm of lab] (sol) {satisfying\\structures\\$\mathsf{Sol}(Q)$};
\draw[flow] (rep) -- node[above]{readout $\rho$} node[below=7mm, note]{$\overcount$ representations\\per labelled structure\\(\cref{lem:encoding_exactness})} (lab);
\draw[flow] (lab) -- node[above]{forget labels $\lambda$} node[below=7mm, note]{$q_\Sigma(\mathbf d)$ labellings\\per satisfying structure\\(\cref{lem:label_forgetting})} (sol);
\end{tikzpicture}
\caption{Structure of the correctness proof. Dividing by $\overcount\cdot q_\Sigma(\mathbf d)$ in the decoder ensures that every satisfying structure contributes exactly one (\cref{prop:component_encoding_correctness}).}
\label{fig:proof_structure}
\end{figure}

Recall that the weight of a ground literal $\ell$ is $\weight(R)$ if $\ell=R(\mathbf a)$ and $\negweight(R)$ if $\ell=\neg R(\mathbf a)$.
In our encoding, each literal weight is a finite sum of monomials with coefficient $1$, written $\sum_{j=1}^{n_\ell}m_{\ell,j}$.

\begin{definition}[Weighted representation]
    \label{def:weighted_representation}
    Let $\world$ be a model of $\sentence$ over $\domain$, including its evidence, and let $L(\world)$ be the set of ground literals true in $\world$.
    A \emph{term selection} for $\world$ is a function
    \[
        \tau:L(\world)\to\mathbb{N}_{>0}
        \qquad\text{such that}\qquad
        \tau(\ell)\leq n_\ell\quad\text{for every }\ell\in L(\world).
    \]
    A \emph{weighted representation} is a pair $r=(\world,\tau)$.
    Its exponent vector $\mathbf d(r)$ is the sum of the exponent vectors of the selected terms.
    The representation is \emph{accepted} if $\mathbf d(r)\models A$, and $\mathsf{Rep}_A(Q)$ denotes the set of accepted representations.
\end{definition}

A single model $\world$ may admit multiple weighted representations, with acceptance governed by the selected terms.
Expanding the \wfomc{} sum without collecting equal monomials yields the following lemma, proved in \cref{app:proof_compiler_correctness}.

\begin{lemma}[Coefficients count representations]
    \label{lem:coefficients}
    Let $f=\symwfomc(\sentence,\domain;\weight,\negweight)=\sum_{\mathbf d}c_{\mathbf d}\mathbf x^{\mathbf d}$.
    Then $c_{\mathbf d}=|\{r: \mathbf d(r)=\mathbf d\}|$ for every exponent vector $\mathbf d$.
\end{lemma}

To relate weighted representations to $\mathsf{Sol}(Q)$ while retaining partition slot labels, we define a \emph{labelled structure} of $Q$ just like a satisfying structure, except that each partition is represented as a slot-ordered sequence $(q_1,\ldots,q_k)$ satisfying the partition-ordering conditions of \cref{sub:enc_partitions}.
All part references and deterministic local expressions are evaluated at those slots, and all analysis bounds hold.
Write $\mathsf{Lab}(Q)$ for the set of labelled structures.
For an accepted representation $r=(\world,\tau)$, its \emph{readout} $\rho(r)$ recovers set memberships, tuple entries, and sequence order from $\world$, bag multiplicities from $\tau$ (or the declared constants for initialized bags), and partition or composition parts in slot order (\cref{def:readout}).

\begin{lemma}[Readout and representation multiplicity]
    \label{lem:encoding_exactness}
    Every accepted representation reads out to a labelled structure, and every labelled structure has exactly $\overcount$ accepted representations under $\rho$.
\end{lemma}
\begin{proof}[Proof sketch]
    Set predicates, bag multiplicity terms, tuple graphs, labelled group parts, and auxiliary constraint predicates are uniquely determined by the labelled structure.
    Only a sequence of length $k$ introduces extra choices, namely $(|\domain|-k)!$ linear orders outside its prefix together with $k!$ position-name assignments for labelled positions, yielding $\overcount$ (or $1$ when no sequence occurs).
    Induction along the dependency order establishes the claim (see \cref{app:proof_compiler_correctness}).
\end{proof}

Next, we define $\lambda:\mathsf{Lab}(Q)\to\mathsf{Sol}(Q)$ by forgetting each partition's slot order and reindexing the normalized auxiliary part variables and deterministic local expressions according to a fixed canonical enumeration determined solely by the partition value (\cref{sec:grouped_objects,app:partwise_semantics}).
Because part-wise constraints are uniform conjunctions of deterministic atomic tests, reindexing merely permutes the conjuncts while fixing external objects (\cref{prop:part_relabelling}), ensuring $\lambda(\hat s)\in\mathsf{Sol}(Q)$.
We call $\hat s$ an \emph{admissible labelling} of $s=\lambda(\hat s)$ and set $\pi=\lambda\circ\rho:\mathsf{Rep}_A(Q)\to\mathsf{Sol}(Q)$.

\begin{lemma}[Number of admissible labellings]
    \label{lem:label_forgetting}
    Every satisfying structure of $Q$ has an admissible labelling.
    For every accepted representation $r$, the satisfying structure $\pi(r)$ has exactly $q_\Sigma(\mathbf d(r))$ admissible labellings.
\end{lemma}
\begin{proof}[Proof sketch]
    In each set partition, nonempty parts are pairwise disjoint and hence distinct, so parts of each positive size can permute freely within their slot block, whereas empty parts are identical and yield no additional slot sequences; the product of these factorial counts equals $q_\Sigma(\mathbf d(r))$.
    Bag partitions admit a unique signature-sorted slot sequence even when parts coincide.
    Relabelling invariance (\cref{prop:part_relabelling}) and sound analysis bounds ensure that precisely these labellings belong to $\mathsf{Lab}(Q)$ (see \cref{app:proof_compiler_correctness}).
\end{proof}

\cref{tab:representation_accounting} summarizes the representation multiplicity and decoder divisor for each ordinary object encoding.
Combining \cref{lem:coefficients,lem:encoding_exactness,lem:label_forgetting} establishes count preservation for a single component.

\begin{table}[tbp]
\centering
\begin{threeparttable}
\caption{Representation multiplicities and decoder divisors for the ordinary object encodings.}
\label{tab:representation_accounting}
\small
\setlength{\tabcolsep}{5pt}
\renewcommand{\arraystretch}{1.15}
\begin{tabularx}{\linewidth}{@{}>{\raggedright\arraybackslash}p{.24\linewidth}>{\raggedright\arraybackslash}X>{\centering\arraybackslash}p{.16\linewidth}>{\centering\arraybackslash}p{.10\linewidth}@{}}
\toprule
\textbf{Object} & \textbf{Source of representation multiplicity} & \textbf{Multiplicity} & \textbf{Divisor} \\
\midrule
\textbf{Set, Bag, Tuple} & Predicates, multiplicity terms, and graphs uniquely fixed & $1$ & --- \\
\textbf{Composition} & Ordered part slots uniquely fixed & $1$ & --- \\
\textbf{Partition (bag)} & Lexicographic signature order fixes one slot sequence & $1$ & --- \\
\midrule
\textbf{Sequence (set prefix)} & Compatible linear orders outside the $k$-element prefix ($\rho$) & $(|\domain|-k)!$ & $\overcount$ \\
\textbf{Sequence (labelled pos.)} & Prefix position-name assignments and orders outside prefix ($\rho$) & $k!\,(|\domain|-k)!$ & $\overcount$ \\
\textbf{Partition (set)} & Permutations within blocks of equal positive part size ($\lambda$) & $\prod_{m>0}r_m!$ & $q_\Sigma$ \\
\bottomrule
\end{tabularx}
\begin{tablenotes}[flushleft]
\footnotesize
\item[] A dash denotes a divisor of $1$. Here $k$ is the fixed sequence length, $\domain$ is the component domain (including allocated position constants), and $r_m$ counts nonempty set-partition parts of size $m$.
\end{tablenotes}
\end{threeparttable}
\end{table}

\begin{proposition}[Correctness of component encoding]
    \label{prop:component_encoding_correctness}
    Let $Q$ be a preprocessed component obtained from a supported \ourlang{} program by case expansion and decomposition, with atomic constraints and fixed tuple and sequence lengths. Let $\sem{Q}=\langle D,O,C\rangle$ and let $rec$ be its analysis record, with sound bounds on labelled assignments before canonical selectors are imposed.
    If $f$ is the exact \wfomc{} polynomial for $\encode(Q,rec)$, then
    \begin{equation*}
        \decode(f,A,\overcount,\Sigma)=\comcount{D,O,C}.
    \end{equation*}
\end{proposition}
\begin{proof}
    By \cref{lem:label_forgetting,lem:encoding_exactness}, $\pi=\lambda\circ\rho:\mathsf{Rep}_A(Q)\to\mathsf{Sol}(Q)$ is surjective, and for every accepted representation $r$, the number of accepted representations mapping to $\pi(r)$ is $\overcount\cdot q_\Sigma(\mathbf d(r))>0$.
    Applying \eqref{eq:decoder_coefficients} and \cref{lem:coefficients} therefore yields
    \begin{align*}
        \decode(f,A,\overcount,\Sigma)
        &=\frac{1}{\overcount}\sum_{\mathbf d\models A}\frac{c_{\mathbf d}}{q_\Sigma(\mathbf d)}
         =\frac{1}{\overcount}\sum_{r\in\mathsf{Rep}_A(Q)}\frac{1}{q_\Sigma(\mathbf d(r))}\\
        &=\sum_{s\in\mathsf{Sol}(Q)}\ \sum_{r:\,\pi(r)=s}\frac{1}{\overcount\cdot q_\Sigma(\mathbf d(r))}
         =\sum_{s\in\mathsf{Sol}(Q)}1
         =|\mathsf{Sol}(Q)|=\comcount{D,O,C}.\qedhere
    \end{align*}
\end{proof}

\begin{theorem}[Correctness of the \ourlang{}-to-\wfomc{} compiler]
    \label{thm:compiler_correctness}
    Let $P$ be a \ourlang{} program whose semantics $\sem{P}=\langle D,O,C\rangle$ is supported in the sense of \cref{def:supported_fragment}.
    Suppose that analysis is sound on labelled assignments before canonical selectors are imposed and that preprocessing preserves assignments up to uniquely recoverable deterministic values.
    If each symbolic counting call returns the exact polynomial, then \cref{alg:compile} returns $\comcount{D,O,C}$.
\end{theorem}
\noindent Requiring analysis soundness \emph{before} canonical partition selectors are imposed ensures that every admissible slot permutation of a satisfying partition remains within the recorded bounds, so the factorial quotient $q_\Sigma$ in \cref{lem:label_forgetting} matches the exact number of accepted labellings.
\begin{proof}[Proof sketch]
    Preprocessing either soundly detects unsatisfiability or preserves solutions bijectively up to uniquely determined eliminated objects.
    Size-range and Shannon expansion partition the solution space into disjoint branches, each decomposing as the Cartesian product of its components, so \eqref{eq:compiler_sum_product} holds.
    By \cref{prop:component_encoding_correctness}, \cref{line:solve} returns each component's exact count, and \cref{alg:compile} multiplies counts within branches and sums them across branches (see \cref{app:proof_pipeline_correctness}).
\end{proof}

\subsection{Sentence Size and Liftability}
\label{sub:liftability}

Count preservation alone does not guarantee tractable solving, because \cref{thm:complexity} yields a polynomial bound in $|\domain|$ for rational weights only when the compiled \ctwo{} sentence $\sentence$ is fixed.
For a family of \ourlang{} programs, we must therefore determine when the predicate vocabulary and sentence template remain fixed after expansion and how the numbers of symbols and monomials scale.

\begin{proposition}[Fixed-sentence and polynomial-representation regimes]
    \label{prop:liftability_regimes}
    Consider a family of supported \ourlang{} programs whose numbers of object rules, group slots, constraints, individually named entity observations, and constrained positions remain fixed after normalization and case expansion.
    \begin{enumerate}
        \item[(i)] \emph{Set-only growth.} If all growing collections use only the set rules, including tuple function graphs over sets, sequences ordering a selected set, and set partitions and compositions, then the predicate vocabulary, the \ctwo{} sentence $\sentence$, and the number of weight symbols remain fixed across the family; scaling changes only the domain $\domain$, unary evidence $\evidence$, and cardinality parameters.
        \item[(ii)] \emph{Fixed-support bag multiplicity growth.} If the set of distinct entities is fixed and only bag multiplicities grow (without growing ordered-object lengths), the sentence template and symbol set remain fixed, while multiplicity bounds and exponent degrees grow.
        \item[(iii)] \emph{Polynomial monomial and branch counts.} Whenever the number of weight symbols is fixed and their degrees are bounded polynomially in $|\domain|$, the number of possible monomials is polynomial in $|\domain|$. 
        If in addition ordered-object length ranges are polynomial in $|\domain|$ for a fixed number of ordered objects, the number of size-expansion branches is also polynomial in $|\domain|$.
    \end{enumerate}
\end{proposition}
\begin{proof}[Proof sketch]
    In (i), each set-based rule contributes a fixed number of predicates and formula templates independent of $|\domain|$, and size constraints introduce at most one symbol per tracked set or count predicate.
    In (ii), the ordinary per-entity predicates and symbols are indexed by the fixed set of entities, so increasing multiplicities only extends the polynomial sums $\sum_j x_{v,e}^j$.
    For (iii), with $m$ symbols of degree at most $p(|\domain|)$, there are at most $(p(|\domain|)+1)^m$ monomials, and a fixed number of ordered objects with polynomial length ranges yields polynomially many length combinations.
\end{proof}

These bounds characterize representation size rather than end-to-end runtime, since compilation relies on symbolic polynomial weights.
\cref{prop:liftability_regimes} does not by itself bound the cost of symbolic products, coefficient arithmetic, and decoding, but it identifies precisely when the fixed-sentence premise of \cref{thm:complexity} holds.

By contrast, the ordinary per-entity bag encoding introduces predicates $B_{v,e}$, $(F_T^{-1})_e$, $Pos_{Q,e}$, and $B_{g_i,e}$ together with their multiplicity symbols for every coordinate retained outside the singleton support encoding of \cref{app:singleton_encoding}.
As the number of such coordinates grows, both the sentence size and the symbol set expand, and the number of monomials can grow exponentially.
For example, in \code{S = choose(U, 2); B = choose_replace(S, 2)} over an initialized set $U$, even though each chosen $S$ has size two, every entity in $U$ may appear twice in $B$, introducing one multiplicity predicate and symbol per entity.
Similarly, a bag-sourced sequence $Q$ introduces a position-label predicate $Pos_{Q,e}$ for each potential entity, and growing its length further enlarges $\domain$, its unary evidence, and the order factor $\overcount$.
\cref{sec:bag_factorization} addresses this multiset scaling bottleneck by sharing multiplicity symbols across exchangeable entities.

\section{Exchangeable Bag Factorization}
\label{sec:bag_factorization}

\cref{sub:liftability} showed that the ordinary bag encoding can introduce one multiplicity predicate and symbol per entity, causing both the sentence and the symbol set to grow with the number of individually tracked coordinates.
Exchangeable bag factorization addresses this bottleneck by assigning all members of an exchangeable entity class a single shared local weight polynomial $f_G$ raised to the class size, reducing the symbol count from one per entity and object to one per tracked object and class.

We first illustrate the idea on two examples.
In \cref{ex:bag_encoding}, entities \code{a} and \code{b} in \code{B} each have multiplicity $2$ and appear symmetrically in both the \wfomc{} polynomial $f = (1+x_a+x_a^2)(1+x_b+x_b^2)$ and the size constraint $A = \{x_a + x_b = 2\}$, so a single shared indeterminate $x$ captures their combined contribution as $(1+x+x^2)^2$.
The next example shows how the same factorization extends to dependent bag operations.

\begin{example}\label{ex:bag_factorization}
Consider the following declarations subject to \code{|X| == 3}, \code{|Y| == 3}, and \code{|Z| == 2}.
\begin{cofolacode}{}
B = bag(a: 2, b: 2, c: 2, d: 2)
X = choose(B)
Y = choose(B)
Z = X & Y
\end{cofolacode}
\noindent For each of the four entities, choosing multiplicities $i$ and $j$ in $X$ and $Y$ determines multiplicity $\min(i,j)$ in $Z$.
All four entities therefore contribute the same factor $F = \sum_{i=0}^{2}\sum_{j=0}^{2}x^i y^j z^{\min(i,j)}$ to the \wfomc{}, where $x,y,z$ track contributions to $|X|,|Y|,|Z|$.
Multiplying one factor per entity gives $F^4$, whose coefficient of $x^3y^3z^2$ is $96$, the exact number of satisfying structures.
Factorization thus reduces the number of multiplicity symbols from $12$ (three per entity) to $3$ (one per tracked object).
\end{example}

The same construction applies to any bag operation whose output multiplicity for an entity depends solely on that entity's multiplicities in the operands (for instance, \code{Z = X - Y} replaces $\min(i,j)$ with $\max(i-j,0)$).
Below, \cref{sub:bag_applicability} states sufficient conditions for identifying exchangeable classes, \cref{sub:bag_local_factor} constructs their shared local factors and \wfomc{} encoding, and \cref{sub:bag_correctness} proves count preservation and analyses the impact on symbol growth.

\subsection{Exchangeability and Applicability Conditions}
\label{sub:bag_applicability}

Throughout this section, $Q=\langle D,O,C\rangle$ denotes a single preprocessed component of a supported program (\cref{sec:count_preservation}), equipped with the analysis record of \cref{sub:preprocessing}, whose objects are \emph{only} sets and bags.
We view a set as a bag whose entity multiplicities are at most one, and refer to both as bags.
Write $\mu_{v,e}$ for the multiplicity of entity $e$ in object $v$ in a satisfying structure, and $\mult_v(e)$ for its inferred finite upper bound (at most one for sets).
Because $Q$ contains neither sequences nor set partitions, $\overcount=1$ and $\Sigma=\emptyset$, so decoding reduces to summing accepted coefficients (\cref{prop:component_encoding_correctness}).

Sharing a local factor is sound only for entities that no operation or constraint can distinguish.
\begin{definition}[Exchangeable entities]
\label{def:set_bag_exchangeability}
\normalfont
Two entities $e,e'\in D$ are \emph{exchangeable in $Q$} if, in every satisfying combinatorial structure of $Q$, simultaneously exchanging their multiplicities in all objects of $O$ while leaving every other entity's multiplicities unchanged yields another satisfying structure of $Q$.
A nonempty set $G\subseteq D$ is an \emph{exchangeable class in $Q$} if every pair of its members is exchangeable in $Q$.
\end{definition}

Let $I\subseteq O$ be the nonempty set of all initialized objects with fixed multiplicities, and let $O_+=O\setminus I$ be the \emph{dynamic} objects.
Because any entity absent from all initialized objects has multiplicity zero in every dynamic object, we restrict attention to $D_I=\bigcup_{R\in I}\support{R}$.
Grouping entities in $D_I$ by their \emph{initialized multiplicity profiles} $\mathbf m_e=(\mu_{R,e})_{R\in I}$ and inferred bound vectors $(\mult_v(e))_{v\in O_+}$ yields the following syntactically checkable sufficient conditions.

\Needspace{12\baselineskip}
\begin{definition}[Applicability of bag factorization]
\label{def:bag_factorization_applicability}
\normalfont
A nonempty group $G\subseteq D_I$ is an exchangeable class eligible for factorization if it satisfies the following conditions.
\begin{enumerate}
    \item Its members share a common initialized multiplicity profile $\mathbf m_e$ and, in each dynamic object $v\in O_+$, a common recorded finite upper bound $\mult_v(e)$ for every $e\in G$.
    \item Every constraint in $C$ is either a positive containment or equality constraint, or a size constraint using only object cardinalities, multiplicities of entities outside $G$, and numerical constants.
\end{enumerate}
\end{definition}

Because all operations and positive comparisons act entitywise, swapping the multiplicities of two members of $G$ across all objects preserves initialized values, operation relations, and every term occurring in size constraints, establishing exchangeability under \cref{def:set_bag_exchangeability}.
Entities referenced by \code{count} multiplicity terms are excluded from $G$, and negated comparisons are excluded because they require a witnessing entity somewhere in the domain.

\subsection{Local States and Factor Encoding}
\label{sub:bag_local_factor}

\begin{definition}[Local states]
\label{def:bag_local_exchangeability}
\normalfont
For an entity $e\in D_I$, a \emph{local state} is a vector $\mathbf q=(q_v)_{v\in O}\in\mathbb Z_{\geq0}^{O}$, where $q_v$ represents the multiplicity of $e$ in $v$, such that
\[
    \begin{aligned}
        q_R&=\mu_{R,e} &&\text{for every }R\in I,\\
        0&\leq q_v\leq\mult_v(e) &&\text{for every }v\in O_+,
    \end{aligned}
\]
and its coordinates satisfy the local relation in \cref{tab:bag_factor_local_rules} for each dynamic object's defining operation and each positive containment or equality constraint in $C$.
The set of all local states for entity $e$ is denoted by $\Omega_e$.
\end{definition}

\begin{table}[tbp]
\centering\small
\caption{Local relations for a single entity in a factorized bag component. Fixed-size variants of the choice operations use the same local relations together with an aggregate size condition.}
\label{tab:bag_factor_local_rules}
\begin{tabular}{@{}ll@{}}
\toprule
Operation or constraint & Local relation \\ \midrule
\code{v = choose(u)} & $q_v\leq q_u$ \\
\code{v = choose_replace(u)} & $q_v\leq\mult_v(e)q_u$ \\
\code{v = supp(u)} & $q_v=\indicator[q_u>0]$ \\
\code{v = u1 \& u2} & $q_v=\min(q_{u_1},q_{u_2})$ \\
\code{v = u1 + u2} & $q_v=\max(q_{u_1},q_{u_2})$ \\
\code{v = u1 ++ u2} & $q_v=q_{u_1}+q_{u_2}$ \\
\code{v = u1 - u2} & $q_v=\max(q_{u_1}-q_{u_2},0)$ \\
\code{u1 subset u2} & $q_{u_1}\leq q_{u_2}$ \\
\code{u1 == u2} & $q_{u_1}=q_{u_2}$ \\
\bottomrule
\end{tabular}
\end{table}

For any class $G$ satisfying \cref{def:bag_factorization_applicability}, all members share the same initialized multiplicities, inferred bounds, operation relations, and constraints in $C$, so $\Omega_e=\Omega_{e'}$ for all $e,e'\in G$.
Write $\Omega_G$ for this common state set, which may be empty.
In \cref{ex:bag_factorization}, for instance, with coordinates ordered as $(B,X,Y,Z)$, the vector $(2,1,2,1)$ is a local state of each entity, and all four entities share the state set $\{(2,i,j,\min(i,j))\mid i,j\in\{0,1,2\}\}$.

To minimize the number of symbols, we track only the object sizes needed by aggregate constraints.
Let $O_c\subseteq O_+$ consist of the dynamic objects whose cardinalities appear in an arithmetic size constraint or whose choice operation has an explicit or inferred exact size, with or without replacement.
Introducing one symbol $x_{v,G}$ for each $v\in O_c$, we define the shared local weight polynomial
\begin{equation}
    f_G(\mathbf{x})=\sum_{\mathbf q\in\Omega_G}
        \prod_{v\in O_c}x_{v,G}^{q_v}.
    \label{eq:lifted_bag_factor}
\end{equation}
The sum ranges over complete local states, including coordinates for objects outside $O_c$, so when several states share the same monomial, its coefficient counts all of those intermediate choices.
In the terminology of \cref{def:weighted_representation}, the weight $f_G$ therefore has one term per local state, and distinct states remain distinct terms even when their monomials coincide.
For a factorized class, the readout of \cref{sub:representation_labelled} recovers set memberships and bag multiplicities from the complete local-state label of the selected term, including coordinates without a size symbol (\cref{app:proof_factorization}).
Multiplying one factor per member of $G$ yields the class contribution $f_G(\mathbf{x})^{|G|}$.

\begin{example}
\label{ex:mixed_bag_factorization}
Consider the \ourlang{} program
\begin{cofolacode}{}
U = set(e1, e2, e3, e4)
S = choose_replace(U)
B = choose(S)
T = supp(S)
|T| == 2 and |S| == 4
\end{cofolacode}
\noindent The four entities form the exchangeable class $G=\{e_1,e_2,e_3,e_4\}$.
Writing $s,b,t$ for an entity's coordinates in $S,B,T$, its local states satisfy $0\leq b\leq s\leq 4$ and $t=\indicator[s>0]$.
Only $S$ and $T$ belong to $O_c$; assigning indeterminates $x$ and $z$ to $|S|$ and $|T|$, for each $s>0$ the $s+1$ choices of $b$ produce the same monomial $x^sz$, yielding
\[
    f_G(x,z)=1+z\sum_{s=1}^{4}(s+1)x^s
    =1+z(2x+3x^2+4x^3+5x^4).
\]
The required count is $[x^4z^2]f_G(x,z)^4=150$.
\end{example}

To incorporate factorized classes into the \wfomc{} instance, we introduce a unary predicate $L_G/1$ for each factorized class $G$, fix its interpretation via $\closedevidence{L_G}{\domain}{G}$, and set $\weight(L_G)=f_G$ and $\negweight(L_G)=1$.
The $|G|$ true groundings contribute $f_G^{|G|}$, replacing the ordinary multiplicity encodings for all entities in $G$, while entities outside every factorized class retain their ordinary encodings.
For each $v\in O_c$, its size expression in $A$ sums the exponents of $x_{v,G}$ across all factorized classes together with the contributions of ordinarily encoded entities.
Because entities referenced in \code{count} multiplicity constraints are excluded from every factorized class, $A$ continues to reference their individual multiplicity symbols directly.

\subsection{Count Preservation and Complexity Impact}
\label{sub:bag_correctness}

\begin{proposition}[Correctness with factorization]
\label{prop:factorization_correctness}
Let $Q$ be a preprocessed set/bag component as in \cref{prop:component_encoding_correctness} with sound analysis bounds, and let $G_1,\ldots,G_m\subseteq D_I$ be pairwise disjoint exchangeable classes satisfying \cref{def:bag_factorization_applicability}, each encoded with its exact local factor $f_{G_j}$ from \eqref{eq:lifted_bag_factor}.
Write $\encode'(Q,rec)=(\sentence',\domain,\weight',\negweight',A',\overcount,\Sigma)$ for the resulting factorized encoding and $f'=\symwfomc(\sentence',\domain;\weight',\negweight')$.
Then the following hold.
\begin{enumerate}
    \item[(i)] When each factor term of $f_{G_j}$ is labelled by its complete local state $\mathbf q\in\Omega_{G_j}$, there is a readout-preserving bijection $\Phi$ between the accepted weighted representations of $\encode(Q,rec)$ and of $\encode'(Q,rec)$.
    \item[(ii)] If $f'$ is computed exactly, then $\decode(f',A',\overcount,\Sigma)=\comcount{D,O,C}$.
\end{enumerate}
Consequently, \cref{thm:compiler_correctness} remains valid when factorization is enabled.
\end{proposition}
\begin{proof}
Weighted representations and readout are defined in \cref{def:weighted_representation,sub:representation_labelled}.
For (i), fix any entity $e\in G_j$.
Because all dynamic operations and positive comparisons in $Q$ act entitywise, the admissible ordinary support and multiplicity-term choices for $e$ across $O$ correspond bijectively to the local states $\mathbf q\in\Omega_{G_j}$, with matching contributions to every size expression in $O_c$ (see \cref{lem:exact_factorization} in \cref{app:proof_factorization}).
Replacing those per-entity choices by the factor term of $L_{G_j}(e)$ labelled with the same complete local state $\mathbf q$ (which remains uniquely recoverable even when distinct states yield the same monomial) and acting as the identity on all entities outside $\bigcup_j G_j$ defines a bijection $\Phi$ on weighted representations.
Because the classes $G_1,\ldots,G_m$ are pairwise disjoint and each replacement preserves every object's size contribution, $\Phi$ preserves both acceptance ($A$ versus $A'$) and readout ($\rho'\circ\Phi=\rho$).

For (ii), since $Q$ contains no sequence or set partition, $\overcount=1$ and $\Sigma=\emptyset$ ($q_\Sigma\equiv 1$) in both encodings.
By \cref{lem:encoding_exactness,lem:label_forgetting}, every satisfying structure of $Q$ has a unique ordinary accepted representation, and hence by (i) a unique factorized accepted representation under $\pi'=\lambda\circ\rho'$.
Applying \eqref{eq:decoder_coefficients} and \cref{lem:coefficients} (as in the proof of \cref{prop:component_encoding_correctness}) gives $\decode(f',A',\overcount,\Sigma)=|\mathsf{Sol}(Q)|=\comcount{D,O,C}$.
Finally, components that are not eligible set/bag components or contain no factorized classes retain the ordinary encoding, and the remaining steps of \cref{thm:compiler_correctness} are independent of the component encoding.
\end{proof}

Each factorized class uses $|O_c|$ symbols regardless of its cardinality $|G|$, whereas the ordinary encoding introduces up to $|G|\cdot|O_+|$ per-entity symbols.
However, the number of monomials in $f_G$ and $f_G^{|G|}$ can still grow with multiplicity bounds and class size, and entities outside factorized classes retain their ordinary coordinates; sharing symbols therefore does not by itself guarantee a fixed sentence or polynomial total runtime in the sense of \cref{sub:liftability}.
In practice, the implementation enforces a compilation budget and falls back to the ordinary encoding whenever a candidate class fails applicability or budget checks (\cref{app:lifted_bag_factorization}).
For entities remaining in the ordinary encoding, eligible $0/1$-bounded \emph{singleton} entities likewise omit per-entity multiplicity symbols and are tracked directly through their support predicates (\cref{app:singleton_encoding}).
\cref{sub:experiments_bag_lifting} evaluates how factorization affects coverage and runtime as the number of distinct entity types grows.

\section{Experiments}
\label{sec:experiments}

In this section, we empirically evaluate the modelling expressiveness of \ourlang{} and the practical performance of our \wfomc{}-based pipeline.
We compare it with four alternative counting pipelines across real-world, synthetic, and growing-domain benchmarks, and perform controlled studies to isolate the effects of lifted inference and exchangeable bag factorization.
Our evaluation aims to answer the following questions:
\begin{description}
    \item[Q1.] What fraction of naturally occurring counting problems can be modelled directly in \ourlang{} using native primitives, what prevents the rest, and how does direct modelling compare with CoLa on representative constructions?
    \item[Q2.] How do coverage and runtime compare across real models and controlled synthetic programs (including unsatisfiable ones)?
    \item[Q3.] How do the pipelines compare as problem size grows, and to what extent does the scaling advantage come from lifted rather than grounded inference?
    \item[Q4.] How does exchangeable bag factorization affect coverage and end-to-end runtime at original sizes and as the number of entity types grows?
\end{description}

\subsection{Setup}
\label{sub:experiments_setup}

Our implementation, \ourlang{}-\wfomc{}\footnote{\url{https://github.com/lucienwang1009/cofola}}, compiles a program as described in \cref{sec:cofola_solution,sec:encoding_rules,sec:bag_factorization} and counts the resulting instances with the lifted solvers of \cref{sub:wfomc}.
In the ablation study (\cref{sub:experiments_bag_lifting}) we run the same programs with exchangeable bag factorization disabled and enabled.
We compare \ourlang{}-\wfomc{} with four alternative pipelines.
\begin{itemize}
    \item \textbf{CNF--Ganak} grounds the ordinary symbolic \wfomc{} encoding to propositional CNF and counts it with the exact model counter Ganak~\citep{sharma_ganak_2019}, whose evaluated version accepts symbolic weights.\footnote{\url{https://github.com/meelgroup/ganak/tree/devel}}
    \item \textbf{ASP--Clingo}~\citep{gebser2022answer} enumerates the answer sets of a generated ASP encoding.\footnote{\url{https://potassco.org/clingo/python-api/5.8/}}
    \item \textbf{ESSENCE--Conjure}~\citep{frisch_essence_2008,akgun_conjure_2022,gent_minion_2006} refines a generated \textsc{Essence} model with Conjure and Savile Row and enumerates its solutions with Minion.\footnote{\url{https://github.com/conjure-cp/conjure/releases/tag/v2.6.0}}
    \item \textbf{CoLa--CoSo}~\citep{totis_lifted_2023} translates the program to CoLa and counts it with the official CoSo implementation.\footnote{\url{https://github.com/PietroTotis/CoSo}}
\end{itemize}
All pipelines start from the same \ourlang{} programs, which are translated automatically into each target formalism (\cref{app:asp_essence_encodings}).
The comparison therefore measures complete pipelines from a common specification, not the best hand-written model in each formalism.
\cref{sub:experiments_language} separately examines direct modelling in \ourlang{} and CoLa using representative MATH problems.

A run is \emph{successful} if it returns the reference count within a time limit of $100$ seconds, including process startup, and a memory limit of $32$\,GiB.
Each program was run three times per pipeline in separate processes, and a program is \emph{solved} if all three runs are successful.
Reference counts are the official answers for the real models, and independent combinatorial formulas or integer dynamic programs.
Every unsuccessful run is classified as a timeout, a count mismatch, an unsupported input, or another error (including runs that exhaust the $32$\,GiB memory limit).
Following the PAR-2 convention of SAT solver evaluations~\citep{froleyks_sat_2020}, each successful run contributes its runtime and each unsuccessful run contributes $200$ seconds; PAR-2 is the mean over all runs, so lower is better and timeouts or errors are penalized consistently across all pipelines.

All experiments were run on a server with Intel Xeon Gold 5218 CPUs at 2.30\,GHz and 503\,GiB of RAM, with each run pinned to one physical core.
The main evaluation spans four benchmark populations, namely $247$ real models from MATH (\cref{sub:experiments_language,sub:experiments_real_world}), $144$ satisfiable and $144$ paired unsatisfiable synthetic programs (\cref{sub:experiments_synthetic}), and $160$ growing-domain instances across eight parameterized families (\cref{sub:experiments_growing}).

\subsection{Modelling Coverage}
\label{sub:experiments_language}

\begin{table}[tbp]
\centering\small
\caption{Qualitative comparison of direct modelling in \ourlang{} and CoLa on five representative MATH problems, with reference counts. Programs for the first two problems appear in \cref{ex:water_polo_cofola,ex:books}.}
\label{tab:math_modelling_examples}
\setlength{\tabcolsep}{5pt}
\renewcommand{\arraystretch}{1.15}
\begin{tabularx}{\linewidth}{@{}>{\raggedright\arraybackslash}p{0.29\linewidth}>{\raggedright\arraybackslash}p{0.42\linewidth}>{\raggedright\arraybackslash}X@{}}
\toprule
\textbf{MATH problem} & \textbf{\ourlang{}} & \textbf{CoLa} \\
\midrule
\textbf{Dependent choices.} Select $7$ starters from $15$ players and designate one as goalie (count $45{,}045$; \cref{prob:water_polo}).
    & Chained \code{choose} over a derived set; counts joint choices directly (\cref{ex:water_polo_cofola}).
    & Single configuration over a fixed base set; requires manual reformulation into disjoint groups. \\
\addlinespace[4pt]
\textbf{Relative order.} Arrange $7$ books with the two math books together and before the physics book (count $720$).
    & Sequence constraints \code{together} and \code{<} (\cref{ex:books}).
    & Position-indexed permutation constraints only; no native block or relative-order primitives. \\
\addlinespace[4pt]
\textbf{Pattern counts.} Count $15$-toss sequences with $2$ HH, $3$ HT, $4$ TH, and $5$ TT overlapping adjacent pairs (count $560$).
    & \code{coin = set(H, T)}\newline
      \code{q = choose_replace_sequence(coin, 15)}\newline
      \code{q.count((H, H)) == 2}\newline
      \code{q.count((H, T)) == 3}\newline
      \code{q.count((T, H)) == 4}\newline
      \code{q.count((T, T)) == 5}
    & No native primitive for counting overlapping adjacent patterns. \\
\addlinespace[4pt]
\textbf{Repeated entities.} Order $4$ indistinguishable crocodiles, $1$ catfish, and $2$ indistinguishable squid without consecutive crocodiles (count $3$).
    & \code{c = bag(croc: 4, catfish, squid: 2)}\newline
      \code{order = sequence(c)}\newline
      \code{not ((croc, croc) in order)}
    & Supports multiset permutations, but lacks a native sequence-level forbidden-adjacency primitive without enumerating adjacent index pairs. \\
\addlinespace[4pt]
\textbf{Joint objects.} Three distinguishable planes take off sequentially, each on one of two runways; count joint order--runway schedules (count $48$).
    & \code{planes = set(blue, red, white)}\newline
      \code{take_off = tuple(planes)}\newline
      \code{runways = compose(planes, 2)}
    & Organizes a model around one target configuration rather than jointly declaring multiple objects over the same base set. \\
\bottomrule
\end{tabularx}
\end{table}

We examine a manually selected collection of $392$ candidate counting questions from the \emph{Counting \& Probability} category of the MATH dataset~\citep{hendrycksmath2021}.
The collection was assembled in four steps.
\begin{enumerate}
    \item Select the \emph{Counting \& Probability} category, which contains $1{,}245$ problems.
    \item Retain problems containing the phrase ``how many'', leaving $505$ problems.
    \item Retain questions whose information is fully available in the textual or mathematical statement, leaving $408$ problems.
    \item Remove $16$ Venn-diagram questions that ask for the size of a region given other sizes, rather than for a number of structures, leaving $392$ candidates.
\end{enumerate}
We wrote direct \ourlang{} models for $247$ of them, achieving an author-implemented direct modelling coverage of $63.0\%$.
Here, direct modelling means encoding the problem with \ourlang{}'s native objects and constraints without introducing auxiliary witness objects to simulate missing primitives or manually splitting the problem into pre-counted cases.
The $247$ models use all six object types, dependencies between choices, and cardinality, membership, multiplicity, position, relative-order, and part-size constraints, and form the real benchmark below.

The remaining $145$ questions fall into eight categories according to the primary missing native primitive, analysed with examples in \cref{app:unsupported}.
Most of them ($89$) involve arithmetic on numerical attributes of entities or ask for the number of distinct numerical outcomes, and $33$ identify arrangements under rotations or reflections.
The others involve grids, equality between positions, windows and runs, comparison with a reference arrangement, joint attribute conditions, or a history-dependent process.
Since we restrict the benchmark to direct models, $63.0\%$ is a lower bound on what can be encoded in \ourlang{} through indirect count-preserving reformulations.
For instance, position-equality in sequences admit indirect encodings using uniquely determined auxiliary singleton sets (see \cref{prob:unmodelled_palindromes} for an example), although doing so complicates the formulation and reduces lifted-solving coverage.
Overall, numerical attributes and general symmetry groups remain the extensions that would most increase direct modelling coverage.

Since automatic translation results measure the translation pipeline rather than the intrinsic expressiveness of CoLa, we compare direct modelling in \ourlang{} and CoLa qualitatively at the language level using five representative MATH problems (\cref{tab:math_modelling_examples}).
These examples highlight recurring constructions, namely dependent choices, relative-order constraints, overlapping adjacency-pattern counts, forbidden adjacency on repeated entities, and jointly counted objects, that \ourlang{} expresses directly as declared objects and constraints, whereas CoLa lacks native primitives for them.

Thus, while both languages cover basic selections and groupings over fixed base collections, \ourlang{} lets users express multi-step dependent choices, jointly counted objects, and sequence-level structural constraints directly from the problem statement, without first reformulating them into an equivalent single-configuration problem.
In answer to \textbf{Q1}, direct \ourlang{} models cover $63.0\%$ ($247/392$) of the candidate MATH problems without auxiliary witness reformulations, with numerical attributes and general symmetry groups accounting for the main gaps.
Moreover, \ourlang{} directly supports dependent choices, joint objects, relative-order constraints, and adjacency patterns that lack native primitives in CoLa.

\subsection{Real Benchmark}
\label{sub:experiments_real_world}

\begin{table}[tbp]
\centering
\begin{threeparttable}
\caption{Solved programs and PAR-2 on the four benchmark populations. Bold marks the best value in each row.}
\label{tab:experiment_summary}
\small
\setlength{\tabcolsep}{5pt}
\renewcommand{\arraystretch}{1.15}
\begin{tabular*}{\linewidth}{@{\extracolsep{\fill}}llrrrrr@{}}
\toprule
\textbf{Population} & \textbf{Measure}
    & \makecell{\textbf{Cofola}\\\textbf{WFOMC}}
    & \makecell{\textbf{CNF}\\\textbf{Ganak}}
    & \makecell{\textbf{ASP}\\\textbf{Clingo}}
    & \makecell{\textbf{ESSENCE}\\\textbf{Conjure}}
    & \makecell{\textbf{CoLa}\\\textbf{CoSo}} \\
\midrule
\multirow{2}{*}{Real ($247$)} & Solved & \textbf{246} & 239 & 243 & 239 & 108 \\
    & PAR-2 (s) & \textbf{2.20} & 8.42 & 4.49 & 14.08 & 112.99 \\
\midrule
\multirow{2}{*}{SAT ($144$)} & Solved & \textbf{134} & 97 & 83 & 81 & 38 \\
    & PAR-2 (s) & \textbf{16.41} & 67.68 & 88.00 & 96.56 & 148.43 \\
\midrule
\multirow{2}{*}{UNSAT ($144$)} & Solved & \textbf{135} & 106 & 101 & 83 & 45 \\
    & PAR-2 (s) & \textbf{16.16} & 55.13 & 62.19 & 92.89 & 138.22 \\
\midrule
\multirow{2}{*}{Growing ($160$)} & Solved & \textbf{160} & 114 & 102 & 99 & 77 \\
    & PAR-2 (s) & \textbf{1.23} & 59.47 & 79.80 & 95.01 & 104.36 \\
\bottomrule
\end{tabular*}
\begin{tablenotes}[flushleft]
\footnotesize
\item[] A program is solved if all three runs return the reference count within $100$ seconds. PAR-2 averages over all runs and counts every unsuccessful run as $200$ seconds.
\end{tablenotes}
\end{threeparttable}
\end{table}

\begin{figure}[tbp]
\centering
\begin{subfigure}[t]{0.495\textwidth}
    \centering
    \includegraphics[width=\linewidth]{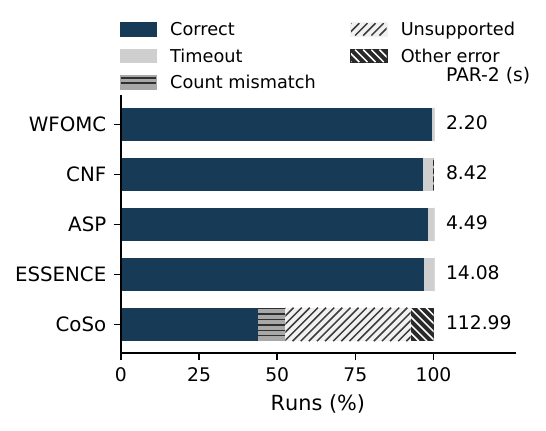}
    \caption{Outcome composition and PAR-2.}
    \label{fig:real_overview}
\end{subfigure}\hfill
\begin{subfigure}[t]{0.42\textwidth}
    \centering
    \includegraphics[width=\linewidth]{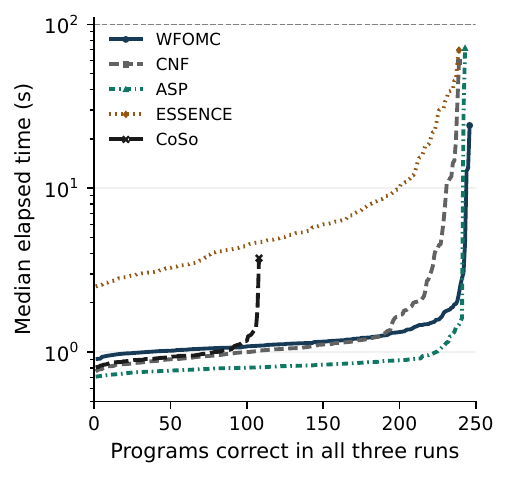}
    \caption{Runtime on solved programs.}
    \label{fig:real_cactus}
\end{subfigure}
\caption{Results on the $247$ real programs. (a) Outcomes over all $741$ runs per pipeline and PAR-2 in seconds; a count mismatch is a returned count that differs from the reference. (b) Number of programs solved within a given median runtime, over the programs solved by each pipeline.}
\label{fig:real_results}
\end{figure}

Before examining each benchmark in detail, \cref{tab:experiment_summary} summarizes solved counts and PAR-2 across all four populations.
\cref{fig:real_results} illustrates the outcomes and runtimes of the five pipelines on the $247$ real MATH programs.
\ourlang{}-\wfomc{} solves $246$ of the $247$ real models and has the lowest PAR-2, followed by ASP--Clingo with $243$.
The sole real program missed by \ourlang{}-\wfomc{} is a six-character licence-plate problem that forms a bag-sourced sequence over $10$ digits and $26$ letters with an adjacency-block constraint, where the sequence precludes bag factorization and none of the five pipelines solves the instance (all three runs of \ourlang{}-\wfomc{}, ASP--Clingo, and ESSENCE--Conjure time out, CNF--Ganak incurs two timeouts and one error, and CoLa--CoSo rejects the input as unsupported).
Although ASP--Clingo rises earlier in the cactus plot on small solved programs, the PAR-2 difference comes from the hardest instances.
CNF--Ganak and ESSENCE--Conjure each solve $239$ programs, with ESSENCE--Conjure substantially slower on the programs it solves.

The real instances are small, which gives little opportunity to amortize the costs of encoding and symbolic manipulation, and the fixed costs of our predominantly Python implementation dominate runs of about one second.
The measurements do not separate this effect from differences between the inference methods.
The real benchmark therefore provides stronger evidence of robustness than of an intrinsic runtime advantage of either approach.

In particular, CoLa--CoSo solves $108$ of the $247$ real programs, with failures falling into three categories.
First, consistent with the modelling differences in \cref{sub:experiments_language}, most failures occur because our automatic translator rejects programs lacking a direct single-configuration CoLa counterpart, such as chained dependent choices, sequences, and groupings permitting empty parts (since CoLa requires nonempty parts).
Second, even though we run the unmodified official CoSo implementation, it raises execution errors on several translated programs (as well as on all Workers instances in \cref{sub:experiments_growing}).
Third, some translated programs return counts differing from the reference.
Together, these outcomes reflect the difficulty of translating general \ourlang{} models automatically into CoLa's single-configuration formalism, motivating the verified single-object comparison in \cref{sub:experiments_synthetic} and \cref{sub:experiments_growing}.

\subsection{Synthetic Programs}
\label{sub:experiments_synthetic}

\begin{figure}[tbp]
\centering
\includegraphics[width=.8\linewidth]{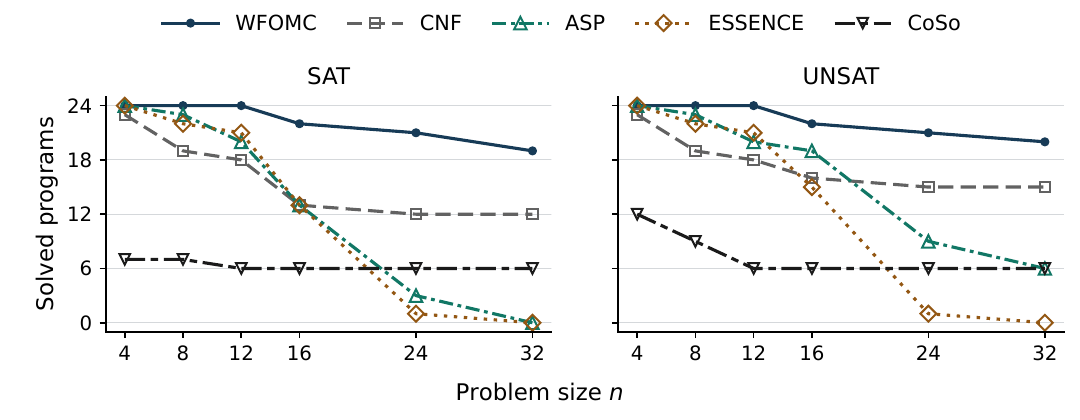}
\caption{Synthetic programs solved at each base size $n$, out of $24$ SAT programs (left) and their $24$ UNSAT companions (right). Each point counts programs for which all three runs returned the reference count.}
\label{fig:synthetic_paired_strata}
\end{figure}

Since the real MATH benchmark has fixed sizes and no unsatisfiable instances, we construct a paired synthetic benchmark that scales base size systematically and evaluates both nonzero counting and zero-count detection.
Each satisfiable program (SAT) is paired with an unsatisfiable companion (UNSAT) that alters a single statement.
For instance, when two disjoint committees exactly fill a team, increasing one committee's required size by one renders the companion infeasible.
The benchmark spans eight structural families, four constructing a single object from a fixed collection and four coupling multiple choices or deriving an object from an earlier choice (see \cref{tab:synthetic_parameters}).
Evaluating three parameter configurations per family across six base sizes $n\in\{4,8,12,16,24,32\}$ (where $n$ counts elements for sets and distinct entity types for bags) yields $144$ SAT/UNSAT pairs.

As shown in \cref{fig:synthetic_paired_strata}, \ourlang{}-\wfomc{} solves $134$ of the $144$ SAT programs and $135$ of their UNSAT companions, achieving the highest coverage and lowest PAR-2 on both populations.
CNF--Ganak ranks second in coverage with $97$ SAT and $106$ UNSAT programs solved, whereas ASP--Clingo and ESSENCE--Conjure lose coverage steeply as $n$ grows.
Although both enumerating pipelines solve every program at $n=4$, by $n=32$ ASP--Clingo solves only six UNSAT programs and ESSENCE--Conjure solves none.

\ourlang{}-\wfomc{} solves every program through $n=12$.
Of the $19$ instances it misses at larger sizes, $16$ belong to the joint bag-selection family (constraining the intersection size of two chosen bags), where coverage drops from all six programs at $n=12$ to two at $n=16$ and zero at $n=24$ and $32$.
Staged diagnostics across all $108$ runs of this family (\cref{tab:joint_bag_diagnostics} in \cref{app:factor_applicability_inventory}) show that bag factorization succeeds cleanly at the local level, building both exchangeable class factors in about $0.05$ seconds in every run and keeping the outer symbol count fixed at six (versus $3n$ without factorization, which already runs out of memory at $n=8$).
The bottleneck is instead polynomial growth in those six outer symbols, whose expanded outer polynomial reaches up to $10^7$ nonzero monomials at $n=16$ and $10^8$ at $n=24$.
Across the $48$ failed runs, $29$ time out during decoding, $19$ exhausts memory during decoding, illustrating the trade-off noted in \cref{sub:bag_correctness} that sharing symbols removes exponential symbol growth without bounding multivariate monomial expansion when multiple aggregate sizes are tracked simultaneously.
The remaining three failures occur on ordered selections (tuples with colour quotas) at $n=32$.

For CoLa--CoSo, our automatic translator rejects all $72$ SAT/UNSAT pairs ($144$ programs) in the four families involving joint or dependent choices.
Among the $72$ single-object pairs ($144$ programs), preprocessing settles two UNSAT cases directly, and for all remaining $142$ models we verified that the generated CoLa formulation counts the same structures as the source program.
Even on these verified single-object models, CoLa--CoSo solves only $38$ SAT and $45$ UNSAT programs within the resource limits, with the remaining runs failing through timeouts or solver errors.

In answer to \textbf{Q2}, \ourlang{}-\wfomc{} achieves the highest coverage and lowest PAR-2 across the real, synthetic SAT, and synthetic UNSAT populations, whereas ASP--Clingo and ESSENCE--Conjure lose coverage rapidly as $n$ increases and CoLa--CoSo is restricted to single-object specifications.

\subsection{Growing Domains}
\label{sub:experiments_growing}

\begin{figure}[tbp]
\centering
\includegraphics[width=.94\textwidth]{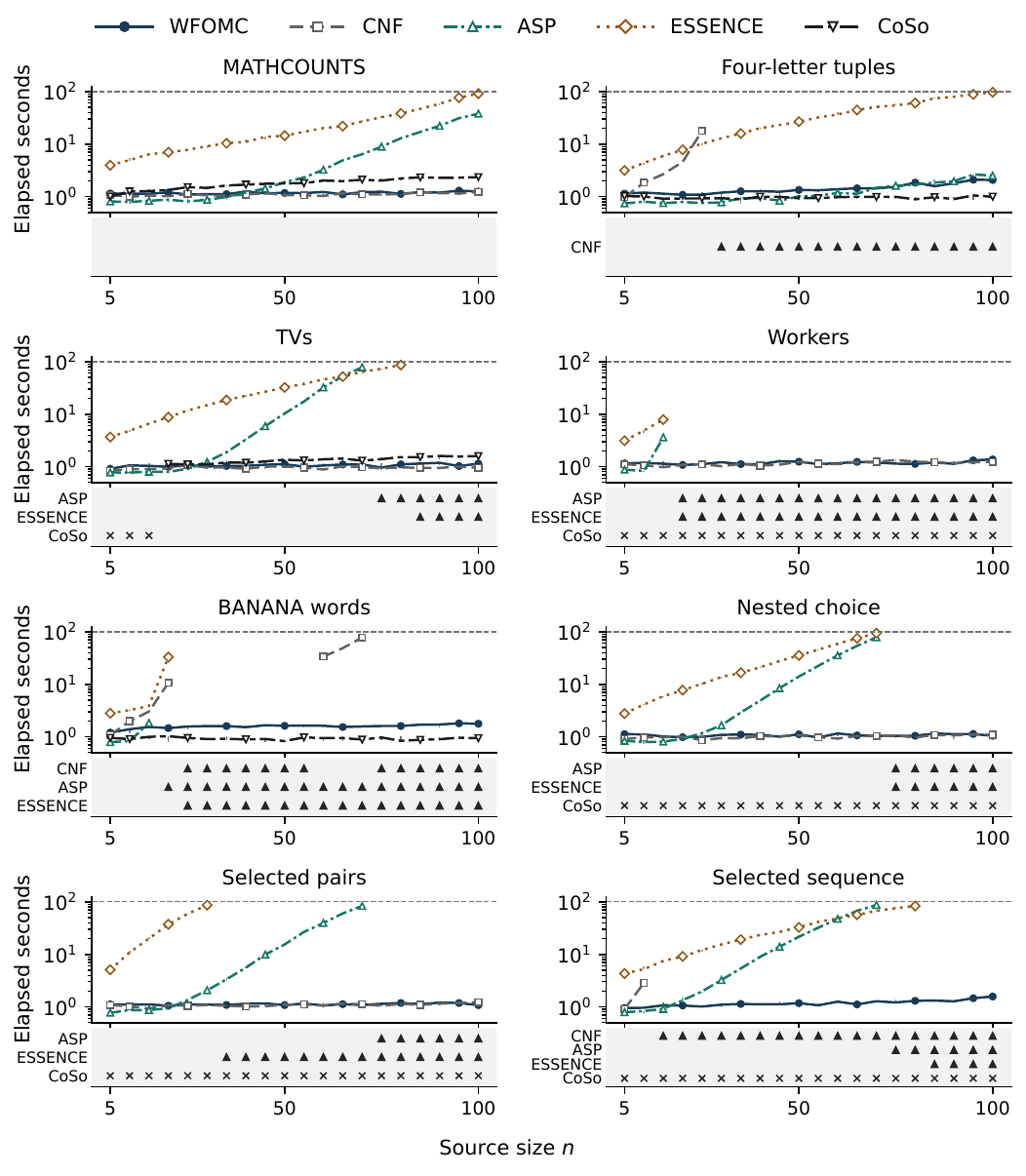}
\caption{Runtime on the eight growing families as a function of the base size $n$. Triangles indicate timeouts and crosses indicate unsupported inputs or errors.}
\label{fig:growing_family_runtime}
\end{figure}

To probe scaling, we construct $160$ instances in $8$ parameterized families, with $n$ ranging from $5$ to $100$ in steps of $5$.
Here $n$ is the cardinality of the base set or the total size of the base bag, including repeated entities.

Three families reproduce problems from \citet{totis_lifted_2023} to compare against the closest prior benchmark, namely \emph{TVs} (select $5$ of $n$ TVs with at least $2$ defective out of a fixed fraction), \emph{Workers} (compose $n$ workers into three labelled groups of prescribed sizes summing to $n$), and \emph{BANANA} (permute a bag of letters B, A, and N whose multiplicities scale to total size $n$).
Two families scale up \cref{prob:mathcounts} (\emph{MATHCOUNTS}) and \cref{prob:fourletter} (\emph{Four-letter}).
The remaining three isolate \emph{dependent choices}, in which an object is constructed from the outcome of an earlier choice, namely \emph{Nested choice} (choose a $4$-person team from $n$ people and then one captain from that team, count $4\binom{n}{4}$), \emph{Selected pairs} (choose $4$ of $n$ people and partition them into two unordered pairs, count $3\binom{n}{4}$), and \emph{Selected sequence} (choose $5$ of $n$ items and arrange them with two distinguished items adjacent, count $2\cdot4!\binom{n-2}{3}$).
Exact parameterizations are listed in \cref{tab:growing_parameters}.

\begin{figure}[tbp]
\centering
\includegraphics[width=.9\linewidth]{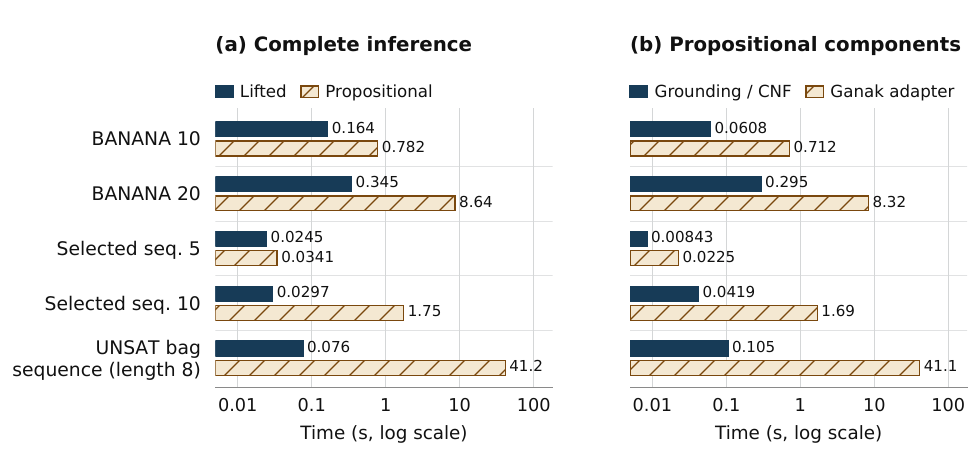}
\caption{Inference time on five instances. (a) Total inference time for the lifted and propositional methods. (b) Time spent on grounding and CNF construction and on the Ganak call, including input serialization, execution, and output parsing.}
\label{fig:matched_inference_costs}
\end{figure}

\cref{fig:growing_family_runtime} shows that \ourlang{}-\wfomc{} solves every instance in about one second, with no visible growth up to $n=100$.
CNF--Ganak solves all instances in five families but fails on larger instances of BANANA, Four-letter, and Selected sequence, all of which involve tuples or sequences.
The runtimes of ASP--Clingo and ESSENCE--Conjure grow with the number of solutions, and both reach the time limit in most families.
CoLa--CoSo is faster than \ourlang{}-\wfomc{} on BANANA, but its runtime grows on TVs, CoSo raises errors on all Workers instances, and our translator rejects the three dependent families.


To isolate the cost of inference from encoding differences, we compared lifted and propositional inference on identical compiled inputs in two controlled settings (with bag factorization disabled).
First, \cref{fig:matched_inference_costs} examines five sequence instances where CNF--Ganak struggles, namely four instances from BANANA and Selected sequence together with a length-eight bag sequence in which a single-copy entity must be adjacent to three distinct entities (making it unsatisfiable).
On this unsatisfiable sequence, lifted inference takes $0.076$ seconds whereas propositional inference takes $41.2$ seconds, of which the Ganak call accounts for $41.1$ seconds.
Second, to test non-sequence structures spanning sets, bags, and set partitions, we ran the same matched-encoding comparison across all $20$ scales ($n=5,\ldots,100$) of Nested choice, MATHCOUNTS, and Selected pairs (\cref{tab:matched_nonseq_inference} in \cref{app:benchmark_diagnostics}).
Both methods solve all $60$ instances exactly, and as grounded CNFs on these families remain compact and linear in $n$ (at most $1{,}783$ variables and $3{,}962$ clauses at $n=100$), Ganak achieves faster inference on $46$ of the $60$ instances, with propositional-to-lifted inference time ratios of $0.60$, $0.49$, and $0.79$ across the three families.
However, while propositional inference time grows linearly with $n$, lifted inference stays nearly flat at $0.12$--$0.13$ seconds and overtakes propositional inference at $n=100$ on all three families ($0.124$ versus $0.170$ seconds on Nested choice, $0.131$ versus $0.153$ seconds on MATHCOUNTS, and $0.124$ versus $0.233$ seconds on Selected pairs).

In answer to \textbf{Q3}, \ourlang{}-\wfomc{} solves all $160$ growing instances in about one second each while grounded and enumerating baselines lose coverage as size grows.
On identical encodings, lifted inference yields decisive speedups on the five sequence instances where propositional grounding or counting explodes, whereas on the three non-sequence families with compact linear CNFs, propositional inference is faster at small and moderate $n$ before flat lifted inference overtakes it by $n=100$.

\subsection{Effect of Bag Factorization}
\label{sub:experiments_bag_lifting}

\begin{figure}[tbp]
\centering
\includegraphics[width=.75\textwidth]{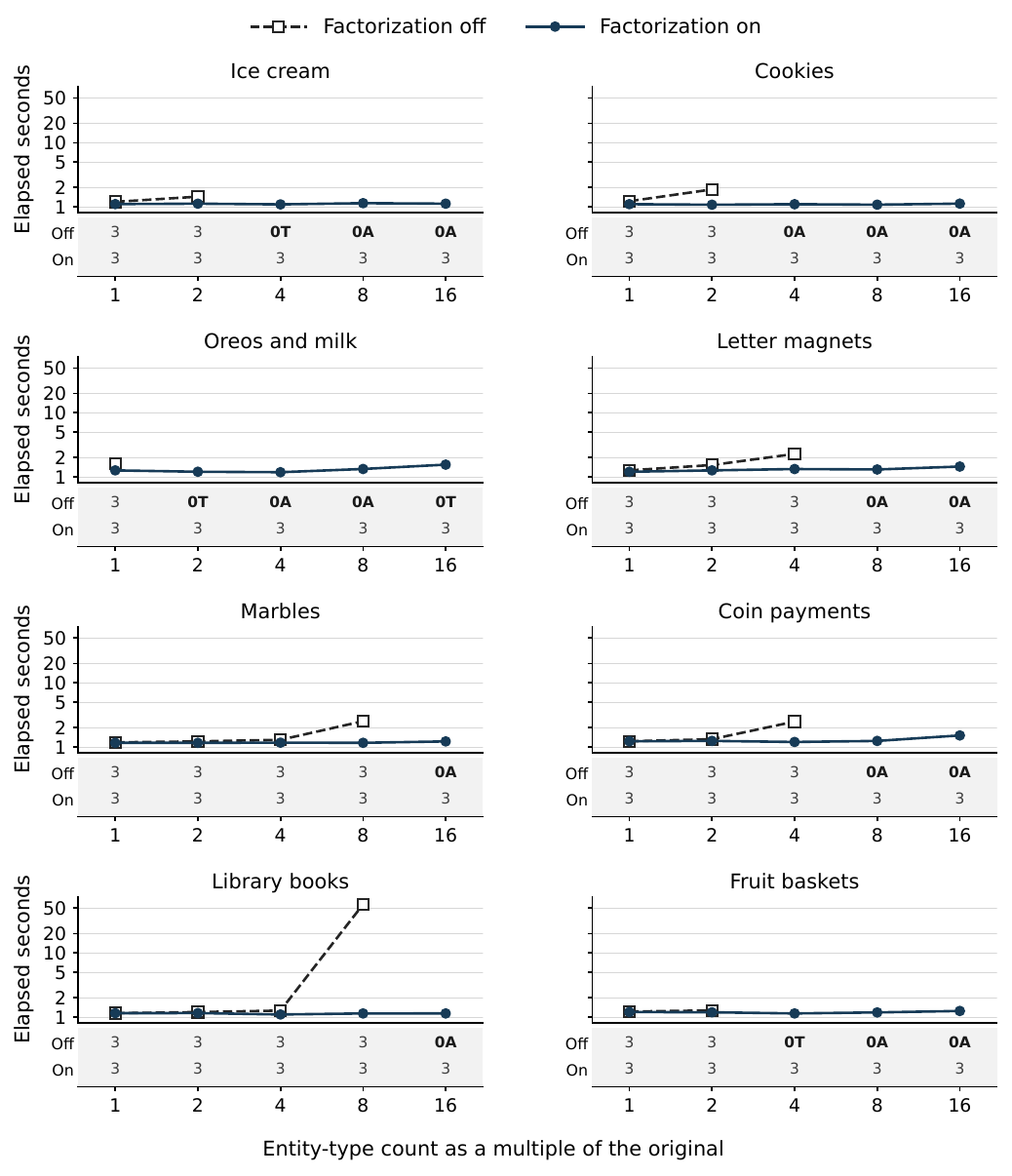}
\caption{End-to-end runtime as the number of entity types increases in eight MATH-derived bag programs. The rows give the number of correct runs out of three for each setting. \texttt{0T} denotes three timeouts and \texttt{0A} denotes three memory allocation failures.}
\label{fig:bag_lifting_ablation}
\end{figure}

At original MATH sizes, bag factorization adds negligible overhead.
Across the $60$ of $247$ programs where factorization applies, both settings solve all $60$ programs, with factorization increasing geometric mean end-to-end runtime by $1.4\%$ (\cref{app:factor_applicability_inventory}).

To measure scaling behaviour as an exchangeable class grows, we inspected all $52$ bag-bearing MATH programs and evaluated all eight sources that yield an exchangeable bag factor either at original scale (five programs, a subset of the $60$ above) or once entity types are doubled (three single-type sources, namely the coin, library, and fruit-basket problems; selection details and independent reference counts appear in \cref{app:factor_applicability_inventory} and \cref{tab:bag_scaling_references}).
For each program, we scaled the number of entity types by factors of $2$, $4$, $8$, and $16$, keeping the selection rules and per-type stock unchanged.
For example, the original library problem counts the ways to have some of six indistinguishable copies of a book checked out and the rest in the library, with at least one copy in each, modelled as
\begin{cofolacode}{}
books = bag(book: 6)
checked_out = choose(books)
|checked_out| > 0
|checked_out| < 6
\end{cofolacode}
\noindent At scale $k$, \code{books} stocks six copies of each of $k$ distinct titles and the upper bound on \code{|checked_out|} becomes $6k$, yielding $7^k-2$ valid allocations.
Scaling all eight sources produces eight original programs and $32$ expanded instances, evaluated with three runs per setting.

\cref{fig:bag_lifting_ablation} shows the results.
As the number of entity types grows, the ordinary encoding quickly hits an exponential wall, whereas factorization avoids it entirely.
Both settings solve all eight original programs, but at $2$, $4$, $8$, and $16$ times the original number of types, the ordinary encoding solves $7$, $4$, $2$, and $0$ of the eight instances, respectively, whereas the factorized encoding solves all eight instances at every scale.
On the $13$ of $32$ expanded instances solved in both settings, factorization achieves a geometric mean speedup of $1.77\times$.
At the largest scale ($16\times$), median runtimes with factorization remain between $1.12$ and $1.56$ seconds.

The library problem illustrates the algebraic compression achieved by the class power $f_G^{|G|}$.
With eight titles, the ordinary encoding tracks eight separate multiplicities from zero to six, generating $7^8=5{,}764{,}801$ monomials.
Factorization replaces these with a single shared symbol recording the \emph{total} number of books borrowed from zero to $48$, compressing the polynomial to $49$ monomials whose coefficients retain the exact number of borrowing patterns at each total.
Excluding totals zero and $48$ leaves the same $5{,}764{,}799$ patterns in both settings, while median end-to-end time drops from $57.20$ to $1.14$ seconds.

In answer to \textbf{Q4}, exchangeable bag factorization adds negligible overhead on the $60$ original MATH models (a $1.4\%$ increase in geometric mean runtime), while on the $32$ expanded instances it raises coverage from $13$ to all $32$ and achieves a $1.77\times$ geometric mean speedup on the $13$ instances solved in both settings.

\section{Related Work}
\label{sec:related}

\ourlang{} connects high-level combinatorial modelling with symbolic lifted model counting.
Its contribution lies in a count-preserving compilation for jointly defined objects, rather than a new general-purpose model-counting algorithm.
We position this contribution relative to modelling languages, symbolic combinatorics, and the logical and algebraic inference methods on which it builds.

\subsection{Modelling Languages for Combinatorial Problems}

Constraint-programming languages have long provided high-level vocabularies for combinatorial structures and constraints over them.
MiniZinc \citep{nethercote2007minizinc} and its predecessor Zinc \citep{marriott_design_2008} model finite-domain CSPs with sets, arrays, and global constraints, and compile to a wide range of CP/SAT/SMT backends.
\textsc{Essence} \citep{frisch_essence_2008} extends this style with high-level types such as multisets and partitions and serves as the source language for the Conjure refinement pipeline.
Answer set programming \citep{gebser2022answer} and probabilistic logic programming, such as ProbLog \citep{de_raedt_problog_2007-1}, similarly allow users to state combinatorial problems declaratively over first-order rules.
These frameworks can support counting through enumeration, propositional model counters~\citep{thurley2006sharpsat}, dedicated answer-set counters~\citep{eiter_treewidth_2021,kabir_exact_2024}, or probabilistic inference.

For counting, the representation must account for how many solver assignments denote each intended structure.
Conjure addresses this requirement through count-preserving refinement~\citep{akgun_conjure_2022}.
Its representation rules eliminate introduced symmetries, including conditional symmetries in unused components, and preservation at each refinement step yields preservation under composition.
For example, representing a two-element subset of $\{a,b,c,d\}$ by distinct entries $(t_1,t_2)$ permits $12$ assignments.
Requiring $t_1<t_2$ under a fixed order leaves six, one for each subset.
The six subsets remain distinct solutions; this correction removes duplicate representations of each one.
Thus count preservation under refinement is established prior work.
The further question for \ourlang{} is how to retain a representation in which lifted inference can aggregate the contributions of different structures.

The closest prior work is CoLa, a declarative language for combinatorial counting problems, paired with the \sharpcsp{}-based solver CoSo~\citep{totis_lifted_2023}.
CoLa specifies configurations such as subsets, permutations, and partitions over sets or multisets, and CoSo applies multiplication, exchangeability, and case analysis directly to the configuration representation.
\ourlang{} shares this objective but models several jointly chosen objects connected by acyclic dependencies, rather than organizing a problem around a single target configuration.
The MATH examples in \cref{tab:math_modelling_examples} illustrate dependent choices, relative-order constraints, and adjacency-pattern counts that are directly modelled in \ourlang{} but lack native primitives in CoLa.
Our automatic translation failures, however, do not by themselves establish a formal expressiveness separation (\cref{sub:experiments_real_world}), and where CoLa--CoSo applies, its specialized rules can be faster than our pipeline.
The inference architecture also differs.
CoSo implements configuration-specific lifted rules, whereas \ourlang{} supplies object encodings and decoding rules to reusable \wfomc{} algorithms~\citep{timothy_van_bremen_faster_2021,zou_faster_2025}, so improvements in those algorithms carry over without changes to the language.

\subsection{Symbolic Combinatorics}

The symbolic method of analytic combinatorics specifies combinatorial classes by constructions such as sequences, sets, multisets, and cycles, and translates each construction into an operation on generating functions~\citep{flajolet_analytic_2009}.
Systems based on it, such as the combinatorial-structure calculus of \citet{flajolet_calculus_1994}, count and generate the objects of a specified class.
\ourlang{} shares the use of generating functions to record sizes and multiplicities, and its bag factorization is a generating-function construction.
The two approaches differ in what they specify.
The symbolic method describes a recursively defined class of structures, typically to count it across all sizes, whereas a \ourlang{} program specifies dependent choices over a finite set of named entities with relational constraints among them and asks for a single exact count.
\ourlang{} therefore relies on first-order logic to express dependencies and constraints that are not constructions of a class.

\subsection{Lifted and Algebraic Model Counting}

Grounded model counters can also exploit exchangeability.
Component caching reuses the counts of repeated residual subformulas, and symmetric component caching additionally recognizes components that are isomorphic under variable renaming~\citep{van_bremen_symmetric_2021-1}.
The latter can reuse subproblems exposed during search even when the original formula has no corresponding global symmetry.
Lifted \wfomc{} exposes regularity in the quantified description before a full propositional instance is constructed.
For an unconstrained $k$-element subset of $n$ exchangeable entities, it can aggregate selections by their size and account for the $\binom{n}{k}$ distinct subsets through a combinatorial factor.
Symmetric caching obtains reuse from the ground components encountered; lifted inference can avoid constructing those components individually when the first-order structure supports aggregation.
The benefit depends on the instance and backend, and our CNF--Ganak comparison does not directly evaluate SYMGANAK.

Lifted probabilistic inference \citep{poole2003first,van_den_broeck_introduction_2021} counts at the level of first-order logic by exploiting symmetries among exchangeable individuals, and \wfomc{} (\cref{sub:wfomc}) is its best-studied formulation.
The two-variable fragment \fotwo{}~\citep{guy_van_den_broeck_completeness_2011} and its counting extension \ctwo{}~\citep{kuzelka_weighted_2021-1,kuusisto_weighted_2018-1} are domain-liftable.
Related algorithms support cardinality constraints~\citep{kuzelka_weighted_2021-1}, linear-order~\citep{toth_lifted_2022-1} and tree axioms~\citep{van_bremen_lifted_2021-1}, and graph conditions such as acyclicity and $k$-component axioms~\citep{kuang2025weighted,malhotra_lifted_2023}, under their respective fragment conditions.
Our encodings reuse cardinality and order support from this body of work.
Their domain-size guarantees apply to fixed sentences and do not automatically bound the complete compilation and symbolic decoding process, whose cost also depends on the generated formula and polynomial representation (\cref{sub:liftability}).
Preserving useful first-order structure also affects how \ourlang{} removes representation redundancy.
For set partitions, a full membership signature would introduce a symbol for every entity and part.
Our encoding instead orders parts by size and corrects the remaining equal-size label permutations during decoding (\cref{sub:enc_partitions}).
The part-wise syntax preserves this label symmetry when deterministic expressions refer to the parts or test them against shared declared choices (\cref{prop:part_relabelling}).
This keeps entity identities out of the symmetry-breaking signature while preserving every satisfying structure.

Algebraic model counting generalizes weighted model counting to semiring-valued labels and explicitly includes polynomial weights~\citep{kimmig_algebraic_2017-1}, and generating functions are a classical tool for representing combinatorial choices by exponents~\citep{Stanley_2011}.
These provide the algebraic foundation for our symbolic weights.
Our contribution is to connect them to the semantics of dependent combinatorial objects through object-specific logical encodings, exponent constraints, and decoding corrections (\cref{sec:encoding_rules,sec:count_preservation}), and to identify when multiplicity symbols can be shared without losing dependencies (\cref{sec:bag_factorization}).

\subsection{Combinatorial Word Problems}

Combinatorial counting problems are also studied as a subclass of math word problems.
\citet{dries_solving_2017-1} pioneered a two-step pipeline that first parses a probability-style word problem into a declarative model and then solves it, and CoLa was later positioned as a target language for the modelling step.
CombEval~\citep{zhou2026combeval} uses \ourlang{} and its solver to evaluate the counting ability of large language models.
\ourlang{} itself is a modelling and reasoning layer and is orthogonal to the parsing front end; an automated text-to-\ourlang{} translator is a natural direction for future work.

\section{Conclusion}
\label{sec:conclusion}

In this paper, we investigated how high-level specifications of dependent combinatorial choices can be counted exactly using reusable lifted inference algorithms.
\ourlang{} addresses this problem across six object types by compiling programs into symbolic \wfomc{}, using first-order relations for structure and dependencies, polynomial weights for multiplicities and sizes, and a count-preserving decoder that corrects repeated representations, complemented by exchangeable bag factorization for multiset scaling.
For the supported fragment under exact symbolic computation, the pipeline provably preserves joint solution counts.
Empirically, \ourlang{}-\wfomc{} solves $675$ of $695$ real, synthetic, and growing-domain instances, giving it the highest coverage of the five pipelines. In the bag-factorization study, it solves all $32$ expanded instances with factorization enabled.

The approach carries clear boundaries.
On the modelling and encoding side, each component permits at most one sequence, part-wise conditions cannot introduce fresh choices, and the language omits numerical entity attributes, general symmetry groups such as circular arrangements, and direct comparisons between tuple positions.
On the complexity and evaluation side, polynomial data complexity applies only when the compiled sentence remains fixed and does not bound symbolic polynomial arithmetic, while our evaluation relies on hand-written MATH models and automatically translated baselines.

These boundaries suggest concrete directions for future work.
Extending \ourlang{} with position-to-position comparisons, attribute arithmetic (via pseudo-Boolean constraints or counting modulo theories), and lifted support for multiple independent orders would broaden modelling coverage.
Controlling intermediate polynomial growth and selecting dynamically between lifted and grounded inference from structural properties of compiled instances could close the remaining runtime gaps.
Finally, \ourlang{} provides a verifiable formal target for translating natural-language counting problems, enabling large language models to delegate exact computation rather than estimating answers directly.

\section*{Acknowledgements}
Yuanhong Wang and Yuxu Zhou are supported by the National Natural Science Foundation of China (No.~62506141).
Juhua Pu is supported by the National Natural Science Foundation of China (No.~62577006).
Ond\v{r}ej Ku\v{z}elka is supported by the Czech Science Foundation project ``The Automatic Combinatorialist'' (24-11820S).

\bibliography{SRL}
\bibliographystyle{plainnat}

\appendix

\clearpage

\noindent{\LARGE\bfseries Appendices}\par\vspace{0.75\baselineskip}
\noindent The appendices provide the formal definitions, complete proofs, implementation details, and supplementary experimental material supporting the main text, organized to mirror the progression of the paper.
\cref{app:full_semantics} specifies the full abstract syntax and denotational counting semantics of \ourlang{}, supplementing \cref{sec:cofola}.
The next four appendices detail the compilation pipeline and establish its correctness.
\cref{app:preprocessing_algorithms} formalizes structural bound analysis, exact LP refinement, count-preserving simplifications, and supported-input checks for \cref{sec:cofola_solution}, while \cref{app:encoding_details} supplies the auxiliary $\ctwo{}$ function relations, size-atom summary, and decoding pseudocode for \cref{sec:encoding_rules}.
\cref{app:correctness_proofs} gives the complete proofs of functionality (\cref{prop:func_axiom}), component and pipeline count preservation (\cref{sec:count_preservation}), and factorization correctness (\cref{sec:bag_factorization}), and \cref{app:lifted_bag_factorization} describes the local CNF counting algorithm and singleton support encoding used to implement bag factorization.
Finally, the last two appendices support reproducibility and additional analysis for \cref{sec:experiments}, with \cref{app:asp_essence_encodings} documenting the automatic ASP and \textsc{Essence} baseline translations and \cref{app:supplementary_experiments} reporting the taxonomy of unmodelled MATH questions, benchmark parameterizations, matched-encoding inference results, and bag-factorization measurements.

\section{Syntax and Formal Semantics of \ourlang{}}
\label{app:full_semantics}

\subsection{Abstract Syntax}
\label{app:abstract_syntax}

\cref{fig:abstract_syntax} collects the declarations and constraints introduced in \cref{sec:cofola}.
Quoted strings are language tokens, while \code{*} and \code{?} denote repetition and optionality, respectively.
Here \code{obj_id} is an object identifier, \code{entity} is an entity name or non-negative integer, \code{INT} is a non-negative integer, and the optional \code{prefix} in an entity range is an identifier prefix ending in a letter.
Size comparisons require a constant right-hand side; integer coefficients may precede cardinalities, and count atoms combine via addition and subtraction.
The grammar omits precedence rules; typing and binding conditions are specified in \cref{app:object_semantics,app:constraint_semantics}.
In particular, part-wise bodies may contain only deterministic expressions (\cref{app:partwise_semantics}) and must iterate over a previously declared grouped object.

\begin{figure}[tbp]
\centering
\begin{cofolacodeblock}{}
program ::= statement*
statement ::= obj_id "=" expr | constraint
expr ::= obj_id | "(" expr ")"
    | "set" "(" set_entry ("," set_entry)* ")"
    | "bag" "(" bag_entry ("," bag_entry)* ")"
    | choose_op "(" expr ("," INT)? ")"
    | "supp" "(" expr ")"
    | expr ("+" | "&" | "-" | "++") expr
    | ("tuple" | "sequence") "(" expr ")"
    | ("partition" | "compose") "(" expr "," INT ")"
    | expr "[" INT "]"
set_entry ::= entity | prefix? INT "..." INT
bag_entry ::= set_entry | entity ":" INT
choose_op ::= "choose" | "choose_replace"
    | "choose_tuple" | "choose_replace_tuple"
    | "choose_sequence" | "choose_replace_sequence"
\end{cofolacodeblock}

\begin{cofolacodeblock}{}
constraint ::= atomic_constraint | "(" constraint ")"
    | "not" constraint | constraint ("and" | "or") constraint
    | atomic_constraint "for" obj_id "in" expr
atomic_constraint ::= entity ("in" | "not" "in") expr
    | expr ("subset" | "disjoint" | "==" | "!=") expr
    | expr "[" INT "]" ("==" | "!=") entity
    | expr "[" INT "]" ("in" | "not" "in") expr
    | pattern ("in" | "not" "in") expr
    | size_expr comp INT
pattern ::= local_pattern | global_pattern
local_pattern ::= "(" arg "," arg ")"
    | "next_to" "(" arg "," arg ")"
    | "(" local_pattern ")"
global_pattern ::= arg "<" arg | "together" "(" expr ")"
    | "(" global_pattern ")"
arg ::= expr | entity
comp ::= "==" | "!=" | "<" | ">" | "<=" | ">="
size_expr ::= size_atom | "(" size_expr ")"
    | size_expr ("+" | "-") size_atom
size_atom ::= INT? "|" expr "|"
    | expr "." "count" "(" arg ")"
    | expr "." "count" "(" local_pattern ")"
    | expr "." "dedup_count" "(" arg ")"
\end{cofolacodeblock}
\caption{Abstract syntax of \ourlang{}. The upper block gives object declarations and expressions, and the lower block gives constraints and counting expressions.}
\label{fig:abstract_syntax}
\end{figure}

\subsection{Semantic Framework}
\label{app:combinatorics_notations}
\label{app:semantic_framework}

For a finite entity domain $D$, set actions range over $\mathcal{U}_{set}=2^D$ and bag actions range over $\mathcal{U}_{mset}$, the set of finite bags with support contained in $D$ and non-negative integer multiplicities.
Under the counting semantics of \cref{sub:language_semantics}, declarations define admissible joint assignments and constraints filter which assignments are counted; \cref{app:denotational_semantics} assembles both from a source program.

An object identifier \code{I} names a typed variable $v_I$.
An environment $\eta$ records these bindings via $\eta(\code{I})=(t,v_I)$, resolving names to previously declared objects and permitting aliases that refer to the same variable.
A combinatorial structure $\structure$ assigns each variable a concrete value; throughout the semantic rules, we write $x_I=\structure(v_I)$ for the value of \code{I} under $\structure$, while an entity name \code{e} denotes $e\in D$.

For an object $(v,(d_1,\ldots,d_k),f)$, the action function $f$ maps dependency values to the set of admissible values for $v$; for instance, $\textsc{SetChoose}(x;k)$ denotes the function $x\mapsto\textsc{SetChoose}(x;k)$ with fixed parameter $k$.
A structure $\structure$ respects the declaration whenever $\structure(v)\in f(\structure(d_1),\ldots,\structure(d_k))$.
Deterministic operations return singleton sets, whereas choice operations may return multiple alternatives.
Given a fixed $\structure$, numerical expressions evaluate to integers and constraints evaluate to truth values without introducing further choices.

\subsection{Semantics of Object Expressions}
\label{app:object_semantics}

The rules below assume named operands so that dependencies are explicit.
Wherever permitted, an inline constructor is desugared into a fresh auxiliary variable recursively, while distinct occurrences of a nondeterministic constructor always introduce separate choices; by contrast, a bare identifier refers to its existing variable and \code{J = I} adds an alias without adding a choice.
Choosing-arrangement abbreviations desugar into a choice followed by an arrangement, such as \code{T = choose_tuple(B, k)} expanding to \code{U = choose(B, k)} and \code{T = tuple(U)} for a fresh \code{U} (and analogously for sequences and replacement choices).
There is no implicit conversion between sets and bags, and part-wise bodies admit only deterministic expressions (\cref{app:partwise_semantics}).

\cref{tab:semantics_act} lists the type signature, dependency tuple $d$, and action function $f$ for each object-forming expression, using type tags $\setclass$, $\multisetclass$, $\mathsf{tup}$, $\mathsf{seq}$, $\mathsf{par}$, and $\mathsf{comp}$ for the six types of \cref{tab:language_objects}, with $t\in\{\setclass,\multisetclass\}$ and $t_{\!part}$ denoting the source type inherited by a composition part.
Only initializers contribute entities to $D$; after expanding half-open ranges (such as \code{set(e1...4)} yielding $\{e_1,e_2,e_3\}$), both \code{set(e1,...,en)} and \code{bag(e1:k1,...,en:kn)} contribute $\{e_1,\ldots,e_n\}$, while all other expressions contribute $\emptyset$.
Omitted bag multiplicities default to $1$, repeated plain bag entries accumulate multiplicities, and set initializers forbid duplicate entries.

\begin{table}[tbp]
\centering
\small
\caption{Object expressions and their admissible-value functions. Type signatures list operand types followed by the result type.}
\label{tab:semantics_act}
\renewcommand{\arraystretch}{1.12}
\begin{tabular}{@{}lccl@{}}
\toprule
Expression & Type signature & Dependencies $d$ & Action $f$ on dependency values \\
\midrule
\code{set(e1,...,en)} & $()\to\setclass$ & $()$ & $\textsc{Init}(();\{e_1,\ldots,e_n\})$ \\
\code{bag(e1:k1,...,en:kn)} & $()\to\multisetclass$ & $()$ & $\textsc{Init}(();\multiset{e_1{:}k_1,\ldots,e_n{:}k_n})$ \\
\code{choose(id,n)} & $\setclass\to\setclass$ & $(v_{id})$ & $\textsc{SetChoose}(x;n)$ \\
\code{choose(id)} & $\setclass\to\setclass$ & $(v_{id})$ & $\textsc{SetChoose}(x;\cdot)$ \\
\code{choose(id,n)} & $\multisetclass\to\multisetclass$ & $(v_{id})$ & $\textsc{MsetChoose}(x;n)$ \\
\code{choose(id)} & $\multisetclass\to\multisetclass$ & $(v_{id})$ & $\textsc{MsetChoose}(x;\cdot)$ \\
\code{choose_replace(id,n)} & $\setclass\to\multisetclass$ & $(v_{id})$ & $\textsc{SetChooseR}(x;n)$ \\
\code{choose_replace(id)} & $\setclass\to\multisetclass$ & $(v_{id})$ & $\textsc{SetChooseR}(x;\cdot)$ \\
\code{supp(id)} & $\multisetclass\to\setclass$ & $(v_{id})$ & $\textsc{Supp}(x)$ \\
\code{id1 + id2} & $\setclass^2\to\setclass$ & $(v_{id1},v_{id2})$ & $\textsc{SetUnion}(x_1,x_2)$ \\
\code{id1 \& id2} & $\setclass^2\to\setclass$ & $(v_{id1},v_{id2})$ & $\textsc{SetIntersect}(x_1,x_2)$ \\
\code{id1 - id2} & $\setclass^2\to\setclass$ & $(v_{id1},v_{id2})$ & $\textsc{SetDiff}(x_1,x_2)$ \\
\code{id1 + id2} & $\multisetclass^2\to\multisetclass$ & $(v_{id1},v_{id2})$ & $\textsc{MsetUnion}(x_1,x_2)$ \\
\code{id1 \& id2} & $\multisetclass^2\to\multisetclass$ & $(v_{id1},v_{id2})$ & $\textsc{MsetIntersect}(x_1,x_2)$ \\
\code{id1 - id2} & $\multisetclass^2\to\multisetclass$ & $(v_{id1},v_{id2})$ & $\textsc{MsetDiff}(x_1,x_2)$ \\
\code{id1 ++ id2} & $\multisetclass^2\to\multisetclass$ & $(v_{id1},v_{id2})$ & $\textsc{AddUnion}(x_1,x_2)$ \\
\code{tuple(id)} & $t\to\mathsf{tup}$ & $(v_{id})$ & $\textsc{Tuple}(x)$ \\
\code{sequence(id)} & $t\to\mathsf{seq}$ & $(v_{id})$ & $\textsc{Seq}(x)$ \\
\code{partition(id,k)} & $t\to\mathsf{par}$ & $(v_{id})$ & $\textsc{Partition}(x;k)$ \\
\code{compose(id,k)} & $t\to\mathsf{comp}$ & $(v_{id})$ & $\textsc{Compose}(x;k)$ \\
\code{id[i]} & $\mathsf{comp}\to t_{\!part}$ & $(v_{id})$ & $\textsc{Part}(x;i)$ \\
\bottomrule
\end{tabular}
\end{table}

The action functions completing \cref{tab:semantics_act} operate on concrete combinatorial values, with ``$\cdot$'' marking omitted size arguments.
For initializers,
\begin{align*}
    \textsc{Init}((); \{e_1, \dots, e_n\}) &\triangleq \{\,\{e_1, \dots, e_n\}\,\}, \\
    \textsc{Init}((); \multiset{e_1{:}k_1, \dots, e_n{:}k_n}) &\triangleq \{\,\multiset{e_1{:}k_1, \dots, e_n{:}k_n}\,\}.
\end{align*}
For sets $x, x_1, x_2 \in \mathcal{U}_{set}$ and $k\in\nat$,
\begin{align*}
    \textsc{SetChoose}(x; k) &\triangleq \{\,S\in \mathcal{U}_{set}\mid S\subseteq x,\ |S| = k\,\}, &
    \textsc{SetChoose}(x; \cdot) &\triangleq 2^x, \\
    \textsc{SetUnion}(x_1, x_2) &\triangleq \{\,x_1\cup x_2\,\}, &
    \textsc{SetIntersect}(x_1, x_2) &\triangleq \{\,x_1\cap x_2\,\}, \\
    \textsc{SetDiff}(x_1, x_2) &\triangleq \{\,x_1\setminus x_2\,\}.
\end{align*}
For bags $x, x_1, x_2 \in \mathcal{U}_{mset}$, set $y\in \mathcal{U}_{set}$, and $k\in\nat$,
\begin{align*}
    \textsc{MsetChoose}(x; k) &\triangleq \{\,M\in \mathcal{U}_{mset}\mid M\subseteq_m x,\ |M| = k\,\}, \\
    \textsc{MsetChoose}(x; \cdot) &\triangleq 2^x, \\
    \textsc{SetChooseR}(y; k) &\triangleq \{\,M\in \mathcal{U}_{mset}\mid \support{M}\subseteq y,\ |M| = k\,\}, \\
    \textsc{SetChooseR}(y; \cdot) &\triangleq \{\,M\in \mathcal{U}_{mset}\mid \support{M}\subseteq y\,\}, \\
    \textsc{Supp}(x) &\triangleq \{\,\support{x}\,\}, \qquad
    \textsc{AddUnion}(x_1, x_2) \triangleq \{\,x_1\uplus x_2\,\}, \\
    \textsc{MsetUnion}(x_1, x_2) &\triangleq \{\,x_1\cup x_2\,\}, \qquad
    \textsc{MsetIntersect}(x_1, x_2) \triangleq \{\,x_1\cap x_2\,\}, \\
    \textsc{MsetDiff}(x_1, x_2) &\triangleq \{\,x_1\setminus x_2\,\}.
\end{align*}
For ordered objects over $x \in \mathcal{U}_{set}\cup \mathcal{U}_{mset}$,
\begin{equation*}
    \textsc{Tuple}(x)=\textsc{Seq}(x)
        \triangleq\bigl\{[e_0,\ldots,e_{|x|-1}]\in D^{|x|}
             \mid\multiset{e_0,\ldots,e_{|x|-1}}=x\bigr\},
\end{equation*}
where a set input is treated as having multiplicity $1$ per element; tuples and sequences share the same value domain and differ only in the constraints they support (\cref{app:ordered_constraint_semantics}).

For grouped objects over a set $x\in\mathcal{U}_{set}$ and positive integer $k$,
\begin{align*}
    \textsc{Partition}(x; k) &\triangleq \bigl\{\multiset{S_1, \dots, S_k} \,\big|\, S_i\subseteq x,\; \textstyle\bigcup_{i=1}^{k} S_i = x,\; S_i\cap S_j = \emptyset\;\forall i\ne j\bigr\}, \\
    \textsc{Compose}(x; k) &\triangleq \bigl\{[S_1, \dots, S_k] \,\big|\, S_i\subseteq x,\; \textstyle\bigcup_{i=1}^{k} S_i = x,\; S_i\cap S_j = \emptyset\;\forall i\ne j\bigr\}.
\end{align*}
For a bag $x\in\mathcal{U}_{mset}$, parts are subbags summing to $x$,
\begin{align*}
    \textsc{Partition}(x; k) &\triangleq \bigl\{\multiset{M_1, \dots, M_k} \,\big|\, M_i\subseteq_m x,\; \textstyle\biguplus_{i=1}^{k} M_i = x\bigr\}, \\
    \textsc{Compose}(x; k) &\triangleq \bigl\{[M_1, \dots, M_k] \,\big|\, M_i\subseteq_m x,\; \textstyle\biguplus_{i=1}^{k} M_i = x\bigr\}.
\end{align*}
Empty parts are allowed in all four cases.
Finally, for a composition $\pi = [S_1, \dots, S_k]\in\textsc{Compose}(x; k)$ and zero-based index $i\in[k]_0$, $\textsc{Part}(\pi; i) \triangleq \{S_{i+1}\}$, inheriting the source type $t_{\!part}$.

\subsection{Semantics of Constraints}
\label{app:constraint_semantics}

Fix a structure $\structure$ respecting all object declarations.
We distinguish the integer value $\operatorname{val}_{\structure}(a)$ of a numerical expression $a$ from the satisfaction judgement $\structure\models c$ for a constraint $c$.

\subsubsection{Basic Constraints and Numerical Expressions}

\cref{tab:semantics_atomic_cst} specifies satisfaction for basic atomic constraints, where \code{not in} and \code{!=} negate the corresponding positive tests.
Operands of \code{subset}, \code{disjoint}, and \code{==} must have the same collection type (both sets or both bags).
Entity membership applies to sets, bags, and tuples (not sequences).
Positive tuple-index assertions evaluate to false when $i\notin[|x_T|]_0$ (so their negations evaluate to true), and \code{T[i] in I} ignores multiplicities when \code{I} is a bag.

\begin{table}[tbp]
\centering
\small
\caption{Satisfaction of basic atomic constraints under $\structure$.}
\label{tab:semantics_atomic_cst}
\renewcommand{\arraystretch}{1.15}
\begin{tabular}{@{}ll@{}}
\toprule
Constraint $c$ & Condition for $\structure\models c$ \\
\midrule
\code{e in I} & $e\in x_I$ \\
\code{I1 subset I2} (sets / bags) & $x_{I1}\subseteq x_{I2}$ \ / \ $x_{I1}\subseteq_m x_{I2}$ \\
\code{I1 disjoint I2} (sets / bags) & $x_{I1}\cap x_{I2}=\emptyset$ \ / \ $\support{x_{I1}}\cap\support{x_{I2}}=\emptyset$ \\
\code{I1 == I2} & $x_{I1}=x_{I2}$ \\
\code{T[i] == e} & $i\in[|x_T|]_0\ \land\ x_T[i]=e$ \\
\code{T[i] in I} & $i\in[|x_T|]_0\ \land\ x_T[i]\in x_I$ \\
\bottomrule
\end{tabular}
\end{table}

For a tuple argument \code{A}, let $E_A$ denote its entity set under $\structure$, namely $x_A$ for a set, $\support{x_A}$ for a bag, or $\{e\}$ for an entity \code{e}.
\cref{tab:semantics_numeric} defines the integer values of numerical atoms, where $\operatorname{PatCount}_{\structure}$ is defined below.
Linear combinations evaluate by ordinary integer arithmetic, and a size comparison \code{a OP K} holds if and only if $\operatorname{val}_{\structure}(a)\bowtie K$, where $\bowtie$ is the integer relation denoted by \code{OP}.

\begin{table}[tbp]
\centering
\small
\caption{Integer values of numerical atoms under $\structure$. Here \code{I} is a set, bag, tuple, or sequence, \code{B} is a bag, \code{T} is a tuple, and \code{Q} is a sequence.}
\label{tab:semantics_numeric}
\renewcommand{\arraystretch}{1.15}
\begin{tabular}{@{}ll@{}}
\toprule
Numerical atom $a$ & Integer value $\operatorname{val}_{\structure}(a)$ \\
\midrule
\code{|I|} & $|x_I|$ \\
\code{B.count(e)} & $x_B(e)$ \\
\code{T.count(A)} & $|\{i\in[|x_T|]_0\mid x_T[i]\in E_A\}|$ \\
\code{T.dedup_count(A)} & $|\{e\in E_A\mid \exists i\in[|x_T|]_0,\ x_T[i]=e\}|$ \\
\code{Q.count(p)} & $\operatorname{PatCount}_{\structure}(x_Q,p)$ \\
\bottomrule
\end{tabular}
\end{table}

\subsubsection{Ordered Constraints}
\label{app:ordered_constraint_semantics}

Sequence constraints are evaluated over positions.
Let $L=x_Q$, let all position variables range over $[|L|]_0$, and for pattern arguments \code{A} and \code{B} (each a set or entity) write $E_A$ and $E_B$ for their entity sets.
For the two local patterns, define
\begin{align*}
    \alpha_{\code{(A,B)}}(i,j)
        &\equiv j=i+1\land L[i]\in E_A\land L[j]\in E_B,\\
    \alpha_{\code{next_to(A,B)}}(i,j)
        &\equiv j=i+1\land
        \bigl((L[i]\in E_A\land L[j]\in E_B)
              \lor(L[i]\in E_B\land L[j]\in E_A)\bigr).
\end{align*}
Then \code{Q.count(p)} evaluates to $\operatorname{PatCount}_{\structure}(L,p)=|\{(i,j)\in[|L|]_0^2\mid\alpha_p(i,j)\}|$, and \code{p in Q} holds if and only if $\operatorname{PatCount}_{\structure}(L,p)>0$ (with \code{p not in Q} requiring zero matches).
Fixing $j=i+1$ ensures each adjacent position pair is counted once even when $E_A$ and $E_B$ overlap.

For the two global patterns (with \code{S} required to be a set in \code{together(S)}),
\begin{align*}
    \structure\models\code{together(S) in Q}
       &\ \Longleftrightarrow\
       \bigl|\{i\in[|L|]_0\mid L[i]\in x_S
             \land(i=0\lor L[i-1]\notin x_S)\}\bigr|\leq1,\\
    \structure\models\code{(A < B) in Q}
       &\ \Longleftrightarrow\
       \forall i\,\forall j\,
          \bigl((L[i]\in E_A\land L[j]\in E_B)\to i<j\bigr).
\end{align*}
Both conditions hold vacuously when no matching element appears; strict precedence fails if any occurrence belongs to both $E_A$ and $E_B$.

\subsubsection{Boolean Combinations and Part-Wise Constraints}

Boolean connectives \code{and}, \code{or}, and \code{not} carry their standard truth-functional semantics over resolved constraints.

\phantomsection
\label{app:partwise_semantics}
A part-wise constraint \code{acst for part in G} binds \code{part} locally over a previously declared partition or composition \code{G}.
Its body \code{acst} must be an atomic condition built solely from deterministic expressions (recursively forbidding \code{choose}, \code{choose_replace}, \code{tuple}, \code{sequence}, \code{partition}, \code{compose}, and choosing-arrangement abbreviations), though it may reference previously declared choice objects whose values remain shared across all parts.
Enumerating the occurrences of $\structure(v_G)$ as $q_1,\ldots,q_k$ (including repeated equal parts),
\[
    \structure\models\code{acst for part in G}
        \ \Longleftrightarrow\
        \bigwedge_{i=1}^{k}
        \bigl(\structure[\code{part}\mapsto q_i]\models acst\bigr).
\]
This conjunction is independent of the enumeration order, introduces no new choices, and adds no indexing to partitions.
Each part-wise statement has its own local scope.
During normalization, the conjunction is expanded using deterministic auxiliary part variables (in composition order, or under a fixed canonical enumeration determined solely by the partition value), which are uniquely determined by $\structure$ and therefore leave the solution count unchanged.

\subsection{Denotational Semantics of Complete Programs}
\label{app:denotational_semantics}

Each statement $s$ denotes a partial transition $T_s$ on construction states $(\eta,D,O,C)\in\mathsf{State}$, undefined on invalid references, types, or bindings.
Normalization $N(P)$ verifies part-wise restrictions, expands abbreviations and inline expressions so operands precede their users, and instantiates local part bindings using fresh deterministic internal variables.
For a normalized declaration \code{I = expr} with type $t$, dependencies $d$, action $f$ (\cref{tab:semantics_act}), entity contribution $D_{\mathrm{expr}}$, and fresh variable $v$,
\begin{equation}
    T_{\code{I = expr}}(\eta,D,O,C)
       =\bigl(\eta[\code{I}\mapsto(t,v)],\ D\cup D_{\mathrm{expr}},\ O\cup\{(v,d,f)\},\ C\bigr),
    \label{eq:obj_dec_semantics}
\end{equation}
while an alias \code{J = I} with $\eta(\code{I})=(t,v)$ sets $T_{\code{J = I}}(\eta,D,O,C)=(\eta[\code{J}\mapsto(t,v)],D,O,C)$ without adding a choice, and a constraint $c$ sets $T_c(\eta,D,O,C)=(\eta,D,O,C\cup\{R_c^\eta\})$.
Denotations compose in source order via $T_\epsilon=\mathrm{id}_{\mathsf{State}}$ and $T_{s_1s_2}=T_{s_2}\circ T_{s_1}$.
Applying a valid program $P$ to the empty state yields $T_{N(P)}(\emptyset,\emptyset,\emptyset,\emptyset)=(\eta,D,O,C)$ and defines $\sem{P}=\langle D,O,C\rangle$.

\section{Preprocessing and Analysis}
\label{app:preprocessing_algorithms}

This section details the bound-inference rules, count-preserving simplifications, and contradiction checks summarized in \cref{sub:preprocessing}.

\subsection{Entity and Structural Size Analysis}
\label{app:entity_analysis}

Write $E_v=\pent(v)$, $U_v=\maxsize(v)$, $K_v=\exactsize(v)$, and, for a bag, $m_v(e)=\mult_v(e)$, with multiplicity bounds zero outside $E_v$; unknown exact sizes and unbounded upper bounds are denoted by $\bot$ and $\infty$, respectively.
Records are maintained for all collections and referenced parts, traversing objects in dependency order.
Each constructor yields candidate bounds that are then capped by capacity: for sets, $U_v=\min(U,|E_v|)$, and for bags with candidate multiplicities $r(e)$,
\begin{equation*}
    U_v=\min\left(U,\sum_{e\in E_v}r(e)\right),
    \qquad
    m_v(e)=\min(r(e),U_v),
\end{equation*}
capping $U_v$ by $k$ first whenever $K_v=k$ is known.
By induction over dependencies, these bounds are sound over-approximations; potential sets may include entities with bound zero, and upper bounds need not be tight.

An initialized set $S$ has $E_v=S$ and $U_v=K_v=|S|$; an initialized bag $M$ inherits its declared entities and multiplicities with $U_v=K_v=|M|$.
A choice without replacement inherits its source's potential entities and upper bounds; an explicit size $k$ sets $K_v=k$ and caps size and bag multiplicities by $k$, whereas an unsized choice does not inherit the source's exact size.
A choice with replacement inherits only potential entities, with an explicit size $k$ setting $K_v=k$ and bounding all multiplicities, while an unsized replacement choice starts unbounded (unless the potential source is empty, giving capacity zero).
\cref{tab:structural_set_bounds,tab:structural_bag_bounds} give the candidate bounds for derived sets and bags.
Notice that difference inherits its left operand's bounds because potential occurrence in the right operand does not guarantee actual occurrence.
Exact sizes propagate across set difference $K_v=K_{u_1}-K_{u_2}$ when $u_2$ is known to be contained in $u_1$ (e.g., chosen without replacement from $u_1$), and across additive bag union $K_v=K_{u_1}+K_{u_2}$.

\begin{table}[tbp]
\small
\centering
\caption{Structural bounds for derived sets, where $E_i$ and $U_i$ refer to operand $u_i$.}
\label{tab:structural_set_bounds}
\begin{tabular}{@{}lll@{}}
\toprule
Definition of $v$ & $E_v$ & $U_v$ \\
\midrule
$\textsc{SetUnion}(u_1,u_2)$ & $E_1\cup E_2$ & $\min(U_1+U_2,|E_1\cup E_2|)$ \\
$\textsc{SetIntersect}(u_1,u_2)$ & $E_1\cap E_2$ & $\min(U_1,U_2,|E_1\cap E_2|)$ \\
$\textsc{SetDiff}(u_1,u_2)$ & $E_1$ & $U_1$ \\
$\textsc{Supp}(u_1)$ & $E_1$ & $\min(U_1,|E_1|)$ \\
\bottomrule
\end{tabular}
\end{table}

\begin{table}[tbp]
\small
\centering
\caption{Candidate bounds for derived bags before common normalization. Absent entities have bound zero.}
\label{tab:structural_bag_bounds}
\begin{tabular}{@{}llll@{}}
\toprule
Definition of $v$ & $E_v$ & Candidate $r(e)$ & Candidate $U$ \\
\midrule
$\textsc{MsetUnion}(u_1,u_2)$ & $E_1\cup E_2$ & $\max(m_1(e),m_2(e))$ & $\sum_e r(e)$ \\
$\textsc{MsetIntersect}(u_1,u_2)$ & $E_1\cap E_2$ & $\min(m_1(e),m_2(e))$ & $\min(U_1,U_2)$ \\
$\textsc{MsetDiff}(u_1,u_2)$ & $E_1$ & $m_1(e)$ & $U_1$ \\
$\textsc{AddUnion}(u_1,u_2)$ & $E_1\cup E_2$ & $m_1(e)+m_2(e)$ & $U_1+U_2$ \\
\bottomrule
\end{tabular}
\end{table}

A full tuple or sequence inherits its source's entities and cardinality bounds, whereas partial and replacement arrangements follow the choice rules above; bag multiplicity bounds are capped by the ordered object's size bound.
A grouped part inherits its source's upper bounds without inheriting an exact size.

\subsection{Size-Constraint Inference and Refinement}
\label{app:max_size_inference}

Linear-programming (LP) refinement incorporates unconditional cardinality comparisons using $=,<,\leq,>,\geq$ (Boolean conditions enter once Shannon expansion fixes them as branch facts), while constraints involving $\neq$, individual bag multiplicities, tuple counts, or sequence patterns remain enforced during encoding.

Each referenced object receives a real variable $z_v$ with $0\leq z_v\leq U_v$.
Eligible integer comparisons are converted into exact integer inequalities $Hz\leq h$, $z\geq 0$ (representing equalities as two inequalities and shifting strict integer inequalities by $1$).
For any objective $c^Tz$, a rational dual multiplier $y$ returned by the simplex solver is verified in exact arithmetic against $y\geq 0$ and $c+H^Ty\geq 0$, certifying $c^Tz\geq -y^Th$ for every feasible $z$ even if $y$ is suboptimal.
Certifying bounds $L_v\leq z_v\leq U'_v$ yields integer endpoints $\ell_v=\max(0,\lceil L_v\rceil)$ and $u_v=\lfloor U'_v\rfloor$; if $\ell_v>u_v$ the branch is infeasible, otherwise $u_v$ tightens $U_v$ and bag multiplicity bounds, and $\ell_v=u_v$ establishes an exact size.
Real infeasibility is certified separately by minimizing $t\geq 0$ subject to $Hz-t\mathbf{1}\leq h$ and $z\geq 0$; a certified positive lower bound on $t$ proves infeasibility.
For full tuples and sequences, $|v|=|u|$ propagates bounds in both directions along the source dependency.

\begin{lemma}[Soundness of static bounds and LP refinement]
    \label{lem:analysis_soundness}
    For any valid program or branch $Q$, every satisfying labelled assignment prior to imposing canonical partition selectors satisfies $v\subseteq E_v$, $|v|\leq U_v$, $|v|=K_v$ whenever $K_v\neq\bot$, and $v(e)\leq m_v(e)$ (for bags) for every object $v$ in the analysis record produced by \cref{app:entity_analysis,app:max_size_inference}. Moreover, whenever LP refinement certifies infeasibility or $\ell_v>u_v$, $Q$ has no satisfying labelled assignment.
\end{lemma}
\begin{proof}
We proceed by induction along dependency order for the structural rules of \cref{app:entity_analysis}.
Initialized sets and bags have fixed support, cardinality, and multiplicities, establishing the base case.
For a without-replacement choice from source $u$, every element and multiplicity of $v$ is bounded by $u$ (and by $k$ when an explicit size $k$ is specified); for a replacement choice from set $u$, every element lies in $u$ and an explicit size $k$ bounds both $|v|$ and every multiplicity.
For each derived set operation in \cref{tab:structural_set_bounds} and bag operation in \cref{tab:structural_bag_bounds}, the candidate sets $E_v$, candidate multiplicities $r(e)$, and candidate size bounds $U$ are pointwise upper bounds on the corresponding set or bag semantics in \cref{app:object_semantics}, and exact-size propagation holds because $|u_1\setminus u_2|=|u_1|-|u_2|$ whenever $u_2\subseteq u_1$ and $|u_1\uplus u_2|=|u_1|+|u_2|$.
Full tuples and sequences preserve the source's entities and cardinality; partial and replacement arrangements inherit bounds identically to choices; and every part of a composition or partition is contained in its source, so inheriting the source's upper bounds holds for every slot ordering prior to canonical selectors.
Capping $U_v$ by $|E_v|$ (for sets), $\sum_{e\in E_v}r(e)$ (for bags), and $K_v$ (when known), and capping $m_v(e)$ by $U_v$, preserves all feasible values because every entity multiplicity is non-negative and sums to $|v|$.

For LP refinement (\cref{app:max_size_inference}), the vector $z$ of actual object cardinalities in any satisfying labelled assignment is a non-negative integer vector satisfying $z_v\leq U_v$ and every unconditional linear cardinality constraint, and hence $Hz\leq h$ (including strict integer inequalities shifted by $1$ and equalities $|v|=|u|$ for full ordered objects).
For any verified dual certificate $y\geq 0$ with $c+H^Ty\geq 0$ checked in exact rational arithmetic, weak duality yields $c^Tz\geq c^Tz+y^T(Hz-h)=(c+H^Ty)^Tz-y^Th\geq -y^Th$, so every feasible $z_v$ lies in $[L_v,U'_v]$.
Because $z_v\in\mathbb{Z}_{\geq 0}$, it also lies in $[\ell_v,u_v]=[\max(0,\lceil L_v\rceil),\lfloor U'_v\rfloor]$, certifying infeasibility when $\ell_v>u_v$, soundness of tightening $U_v$ and $m_v(e)$ by $u_v$, and exactness when $\ell_v=u_v$.
Similarly, a certified positive lower bound on $t\geq 0$ under $Hz-t\mathbf{1}\leq h$ rules out $t=0$ and hence proves $Hz\leq h$ has no non-negative solution.
\end{proof}

\subsection{Count-Preserving Simplifications}
\label{app:optimizations}

Each preprocessing rewrite preserves satisfying assignments bijectively up to uniquely recoverable deterministic values.
\begin{itemize}
    \item \textbf{Constant folding.} Deterministic set and bag operations over initialized operands are evaluated directly and replaced by the resulting literal collection.
    \item \textbf{Full-choice elimination.} Let $v$ be a without-replacement choice of explicit size $k$ from source $u$ with inferred exact size $K_u=k$. Before replacing $v$ by a reference to $u$ (or, for an ordered choice, removing only the selection step), the rewriter ensures that $|u|=k$ remains enforced without relying on $v$ itself. If $u$ is an initialized collection or a constructor with fixed size $k$, its definition already enforces $|u|=k$; if $u$ is an unsized choice constructor admitting a size parameter, size $k$ is embedded directly into its definition; otherwise (for derived sources such as \code{supp}, unions, intersections, or differences), an explicit deduplicated size constraint \code{|u| == k} is added to the program. Chains of full-choice replacements collapse to their final surviving source, which carries this enforcing definition or constraint into its split component, ensuring that local preprocessing after splitting also preserves $|u|=k$. Unsized and replacement choices are never eliminated.
    \item \textbf{Deterministic sharing.} Identical deterministic definitions (same constructor, parameters, and dependency references) are shared, whereas independent choices, arrangements, and groupings remain distinct.
    \item \textbf{Cardinality folding.} Known exact sizes substitute $a_j|v_j|$ by $a_jK_{v_j}$ in linear comparisons whenever the facts enforcing $K_{v_j}$ remain present in object definitions or retained constraints; a fully evaluated true comparison may then be dropped, while a false one signals a contradiction.
\end{itemize}

\subsection{Contradictions and Supported-Input Checks}
\label{app:sanity_checks}

Preprocessing returns $\bot$ whenever it certifies a contradiction, including negative or conflicting exact sizes, an exact size exceeding its upper bound, insufficient source capacity for a choice, an infeasible LP or empty integer size interval, a false fully evaluated comparison, conflicting unary evidence, or an out-of-range positive tuple-index assertion within a fixed-length branch.
These checks are sound but incomplete; remaining unsatisfiability is caught during counting and decoding.

Before execution, the compiler enforces the supported-fragment conditions of \cref{def:supported_fragment} together with the part-wise syntax restrictions of \cref{sec:grouped_objects,app:partwise_semantics}.
Specifically, part-wise constraints must occur conjunctively with purely deterministic atomic bodies, every ordered object length and symbolic bag multiplicity must have a certified finite upper bound, and every split component may contain at most one sequence.
Any violation raises an unsupported-program error rather than returning zero, distinguishing unsupported inputs from unsatisfiable programs.

\section{WFOMC Encoding and Decoding Details}
\label{app:encoding_details}

This section provides the auxiliary functionality axioms, the summary of size-atom translations, and the decoding pseudocode referenced in \cref{sec:encoding_rules}.

\subsection{Auxiliary Function Relations}
\label{app:function_relations}

The inverse relation $F_T^{-1}/2$, image predicate $Img_{F_T}/1$, and strengthened functionality axioms used in \cref{sub:enc_tuples} are expressed in $\ctwo{}$ with cardinality constraints as follows.
\begin{itemize}
    \item \textbf{Inverse.} $\forall x\forall y: F_T^{-1}(y,x) \leftrightarrow F_T(x,y)$.
    \item \textbf{Image.} $\forall y: Img_{F_T}(y) \leftrightarrow \exists x\, F_T(x,y)$.
    \item \textbf{Injectivity.} $Func_{inj}(F_T, Dom, CoDom) \equiv Func(F_T, Dom, CoDom) \land (|F_T| - |Img_{F_T}| = 0)$.
    \item \textbf{Surjectivity.} $Func_{surj}(F_T, Dom, CoDom) \equiv Func(F_T, Dom, CoDom) \land \forall y\,(CoDom(y)\!\to\!\exists x\,F_T(x,y))$.
    \item \textbf{Bijectivity.} $Func_{bij}(F_T, Dom, CoDom) \equiv Func_{inj}(F_T, Dom, CoDom) \land Func_{surj}(F_T, Dom, CoDom)$.
\end{itemize}
For a bag-sourced tuple and entity $e$, the unary fibre predicate $(F_T^{-1})_e/1$ marks all positions mapped to $e$ via $\forall x:(F_T^{-1})_e(x)\leftrightarrow\exists y\,(F_T(x,y)\land U_e(y))$, with positive weight $x^{\mathrm{occ}}_{T,e}$.
For \code{choose_replace_tuple(u, k)} over a set $u$, the encoder uses $Func(F_T,I_T,S_u)$ without injectivity or surjectivity.

\subsection{Size Atoms and Decoding Algorithm}
\label{app:decoder_algorithm}

\cref{tab:size_atom_encoding} summarizes the exponent expressions used in \cref{sub:enc_size_constraints}, where $\mathbf d[\xi]$ denotes the exponent of indeterminate $\xi$ in an exponent vector $\mathbf d$.
In the ordinary bag encoding, $\mu_{v,e}$ is the declared constant for an initialized bag, $\mathbf d[x_{v,e}]$ for a derived bag with its own multiplicity symbols, and zero outside $\pent(v)$; eligible $0/1$-bounded singleton entities (\cref{app:singleton_encoding}) contribute $1$ when $B_v(e)$ holds and $0$ otherwise (tracked in aggregate via the shared singleton cardinality symbol $x_{v,\mathrm{sing}}$), while aliases reuse their source representation.
For a full ordered object over a dynamic or derived source $u$ in a length-$k$ branch, the materialized constraint \code{|u| == k} enters $A$ through the corresponding set or bag cardinality row of \cref{tab:size_atom_encoding}.

\begin{table}[tbp]
\small
\centering
\caption{Size atoms as exponent expressions in $A$. Here $F_T$ represents tuple \code{T}, $Img_{F_T}$ is its image, and $R_A$ represents membership in argument \code{A}. Auxiliaries $C^{\mathrm{cnt}}_{T,A}$, $C^{\mathrm{dist}}_{T,A}$, and $C^{\mathrm{pat}}_{Q,p}$ carry positive weights $x^{\mathrm{cnt}}_{T,A}$, $x^{\mathrm{dist}}_{T,A}$, and $x^{\mathrm{pat}}_{Q,p}$, respectively. Free logical variables are universally quantified.}
\label{tab:size_atom_encoding}
\begin{tabular}{@{}lll@{}}
\toprule
Atom & Representation & Expression in $A$ \\
\midrule
Set size \code{|v|} & Set predicate $S_v$, with $\weight(S_v)=x_v$ & $\mathbf d[x_v]$ \\
Bag size \code{|v|} & Multiplicity terms & $\sum_e\mu_{v,e}$ \\
\code{v.count(e)} (bag) & Multiplicity term & $\mu_{v,e}$ \\
\code{T.count(A)} & $C^{\mathrm{cnt}}_{T,A}(x)\leftrightarrow\exists y\,(F_T(x,y)\land R_A(y))$ & $\mathbf d[x^{\mathrm{cnt}}_{T,A}]$ \\
\code{T.dedup_count(A)} & $C^{\mathrm{dist}}_{T,A}(x)\leftrightarrow Img_{F_T}(x)\land R_A(x)$ & $\mathbf d[x^{\mathrm{dist}}_{T,A}]$ \\
\code{Q.count(p)} & Adjacency predicate $C^{\mathrm{pat}}_{Q,p}$ (\cref{sub:enc_size_constraints}) & $\mathbf d[x^{\mathrm{pat}}_{Q,p}]$ \\
\bottomrule
\end{tabular}
\end{table}

\cref{alg:decode} implements \eqref{eq:decoder_coefficients}.
For each monomial in $f$ whose exponent vector satisfies $A$, it computes the set-partition quotient $q_\Sigma(\mathbf d)$ from the positive part-size multiplicities across $\Sigma$, accumulates the corrected coefficients in exact rational arithmetic, and divides the total by $\overcount$.

\begin{algorithm}[tbp]
    \DontPrintSemicolon
    \SetNoFillComment
    \caption{$\mathbf{decode}(f,A,\overcount,\Sigma)$}
    \label{alg:decode}
    \KwIn{\wfomc{} polynomial $f$; exponent constraints $A$; global factor $\overcount$; set-partition size groups $\Sigma$}
    \KwOut{The decoded count of the component}
    \lIf{$f=0$}{\Return $0$}
    $ans \gets 0$\;
    \ForEach{monomial $c\cdot m$ in $f$}{
        $\mathbf d \gets$ exponents of $m$, with $\mathbf d[\xi]=0$ for absent variables $\xi$\;
        \lIf{$\mathbf d$ does not satisfy $A$}{\textbf{continue}}
        $quotient \gets 1$\;
        \ForEach{set-partition size group $(x_{g_1},\dots,x_{g_k})\in \Sigma$}{
            \ForEach{distinct positive size $s$ with frequency $r$ in $(\mathbf d[x_{g_1}],\ldots,\mathbf d[x_{g_k}])$}{
                $quotient \gets quotient\cdot r!$\;
            }
        }
        $ans \gets ans + c/quotient$\;
    }
    \Return $ans/\overcount$\;
\end{algorithm}

\section{Correctness Proofs}
\label{app:correctness_proofs}

This appendix supplies only the proofs and auxiliary definitions deferred from \cref{sub:enc_tuples,sec:count_preservation,sec:bag_factorization}.

\subsection{Functionality}
\label{app:proof_func_axiom}

\begin{proof}[Proof of \cref{prop:func_axiom}]
Fix an interpretation $\world$ over the finite domain $\domain$.
If $Func(F,Dom,CoDom)$ holds, every element of $Dom^{\world}$ has a unique $F^{\world}$-successor in $CoDom^{\world}$ and no pair in $F^{\world}$ starts outside $Dom^{\world}$, so the first-coordinate projection $F^{\world}\to Dom^{\world}$ is a bijection and $|F^{\world}|=|Dom^{\world}|$; totality and typing supply the remaining conjuncts of $\Psi$.
Conversely, if $\Psi$ holds, typing gives $F^{\world}\subseteq Dom^{\world}\times CoDom^{\world}$ and totality makes the first-coordinate projection $F^{\world}\to Dom^{\world}$ surjective; since both finite sets have equal cardinality, the projection is bijective, so every element of $Dom^{\world}$ has a unique successor in $CoDom^{\world}$.
Conjoining any $\Phi$ preserves this equivalence.
\end{proof}

\subsection{Proofs for Section~\ref{sec:count_preservation}}
\label{app:proof_compiler_correctness}

We first record the part-relabelling property invoked in \cref{sec:grouped_objects,sec:count_preservation}, followed by the formal readout map and the proofs of \cref{lem:coefficients,lem:encoding_exactness,lem:label_forgetting}.

\begin{proposition}[Relabelling invariance of part-wise conditions]
    \label{prop:part_relabelling}
    Let $P$ be a valid \ourlang{} program and fix a branch of its Boolean and ordered-length expansion.
    Before canonical part-ordering selectors are imposed, every permutation of a set partition's temporary part labels, extended by reevaluating the deterministic expressions in its part-wise bodies, maps satisfying labelled assignments bijectively onto satisfying labelled assignments of the same branch while fixing original entities and all declared object values.
    Permutations within equal-size blocks also preserve the part-size ordering selector.
\end{proposition}
\begin{proof}
Fix the shared declared values $\mathbf s$ and enumerate the partition parts as $X_1,\ldots,X_k$.
Each part-wise condition evaluates a conjunctive atomic test $\bigwedge_{i=1}^{k}\varphi(X_i;\mathbf s)$ over deterministic local expressions $E_j(X_i;\mathbf s)$.
For any permutation $\sigma$, setting $X'_i=X_{\sigma^{-1}(i)}$ and reevaluating $E_j(X'_i;\mathbf s)$ preserves coverage, disjointness, and the conjunction above without introducing new choices.
Because partitions cannot be indexed, part bindings cannot escape their scope, part-wise conditions appear only conjunctively, and all ordered objects are declared outside part-wise bodies, neither declared objects nor global Boolean or length-branch guards can observe temporary part labels.
Hence the permutation stays within the same branch, applying $\sigma^{-1}$ gives its inverse, and permuting only equally sized parts preserves the size-ordering vector.
\end{proof}

\begin{definition}[Readout]
    \label{def:readout}
    For an accepted weighted representation $r=(\world,\tau)$ under the ordinary encodings of \cref{sec:encoding_rules}, its \emph{readout} $\rho(r)$ assigns values to objects of $Q$ as follows (and analogously for partial representations).
    \begin{itemize}
        \item For a set $v$, $\rho(r)(v)=\{e\in D\mid\world\models S_v(e)\}$.
        \item For an initialized bag $v$, $\rho(r)(v)$ takes its declared multiplicities; for a derived bag $v$, the multiplicity of $e\in D$ is $0$ if $e\notin\pent(v)$ or $\world\not\models B_v(e)$. Otherwise, when $e$ uses the singleton support encoding (\cref{app:singleton_encoding}), its multiplicity is $1$; in the general case, its multiplicity is $j$, where $x_{v,e}^j$ is the term selected by $\tau$ for the true literal $B_{v,e}(e)$.
        \item For a tuple $T$ of length $k$, $\rho(r)(T)=[e_0,\ldots,e_{k-1}]$, where $e_i$ is the unique entity with $\world\models F_T(idx_{T,i+1},e_i)$ for $i\in[k]_0$.
        \item For a sequence $v$ over a set source $u$, $\rho(r)(v)$ lists $\rho(r)(u)$ in $L$-order; for a bag source of length $k$, let $p_0,\ldots,p_{k-1}$ list the elements of $I_v$ in $L$-order and set $\rho(r)(v)=[e_0,\ldots,e_{k-1}]$, where $\world\models Pos_{v,e_i}(p_i)$.
        \item For a partition or composition $g$ with slots $g_1,\ldots,g_k$, $\rho(r)(g)=(\rho(r)(g_1),\ldots,\rho(r)(g_k))$.
    \end{itemize}
    Part references and composition-index aliases inherit their slot values, and deterministic part-wise expressions are read by their type rules at that slot.
\end{definition}

\begin{proof}[Proof of \cref{lem:coefficients}]
By definition, $f=\sum_{\world}\prod_{\ell\in L(\world)}\bigl(\sum_{j=1}^{n_\ell}m_{\ell,j}\bigr)$ over models $\world$ of $\sentence$ on $\domain$.
Because the domain and multiplicity bounds are finite, each weight is a finite sum of terms, and expanding the product by distributivity produces exactly one summand $\mathbf x^{\mathbf d(r)}$ per representation $r=(\world,\tau)$.
Collecting equal monomials yields $c_{\mathbf d}=|\{r:\mathbf d(r)=\mathbf d\}|$.
\end{proof}

\begin{proof}[Proof of \cref{lem:encoding_exactness}]
Write $\widehat f_v$ for the constructor action of object $v$ on labelled values, coinciding with $f_v$ of \cref{app:object_semantics} for ordinary objects and admitting selector-respecting slot sequences $\mathbf q$ with $\operatorname{forget}(\mathbf q)\in f_g(a)$ for each partition $g$.
Process rule units $v_1,\ldots,v_n$ in dependency order (each grouped object together with its slots forming one unit), followed by the constraints.
We first verify that, given any fixed partial representation of a unit's dependencies, its fresh predicates and weight terms admit exactly one valid extension per value in $\widehat f_{v_i}$ within the analysis bounds, except that a sequence rule admits $\overcount$ extensions per list.
\begin{itemize}
    \item \emph{Sets and bags.} Closed-world evidence fixes initialized sets and bag supports (with constant multiplicities), and Boolean equivalences uniquely fix derived set operations; without-replacement set choices admit exactly the subsets of their realized source.
    For derived bags, each multiplicity in $\{0,\ldots,\mult_v(e)\}$ corresponds uniquely to a support atom and selected exponent (with $\mult_v(e)=0$ fixing the support atom to false), so the element-wise equations in \cref{tab:bag_encoding_rules} and the source comparisons in $A$ represent each permitted bag once.
    For eligible singleton entities (\cref{app:singleton_encoding}), support membership determines the $0/1$ multiplicity, and checking eligibility on structural bounds before constraint tightening guarantees that at most one operand of an additive union can contain an eligible singleton, so support disjunction equals the multiplicity sum.
    \item \emph{Tuples.} By \cref{prop:func_axiom}, $F_T$ assigns a unique image to each named position; bijectivity, injectivity, or unrestricted functionality with fibre-cardinality equations $x^{\mathrm{occ}}_{T,e}=\mu_{u,e}$ match the exact list-constructor multiplicity conditions, and each permitted list uniquely fixes $F_T$, $Img_{F_T}$, and $(F_T^{-1})_e$.
    \item \emph{Sequences.} A permitted list of length $k$ uniquely determines the $L$-order on its prefix.
    For a set source, every total order on the $|\domain|-k$ non-source elements is admissible, giving $(|\domain|-k)!$ extensions.
    For a bag or replacement source, any of $k!$ assignments of the distinct position constants $idx_{Q,1},\ldots,idx_{Q,k}$ to prefix positions combines with any of $(|\domain|-k)!$ tail orders, giving $k!\,(|\domain|-k)!$ extensions.
    Because no other rule constrains $L$ outside the prefix and each component contains at most one sequence, every permitted list has exactly $\overcount$ extensions.
    \item \emph{Compositions and partitions.} Coverage together with disjointness (for sets) or per-entity multiplicity conservation (for bags, including the exactly-one-part singleton rule) admits exactly the slot sequences summing to the source, while the selectors in $A$ restrict partitions to size- or signature-nondecreasing sequences.
    Each slot sequence uniquely determines all part predicates and multiplicity terms, and composition indexing adds no choice.
    \item \emph{Constraints and size atoms.} Basic support, multiplicity, and tuple-index constraints test the read-out values directly, with negated support conditions using an equivalence-defined violator predicate that introduces no choice.
    On a sequence prefix $z_0,\ldots,z_{k-1}$, $L(z_i,z_j)$ encodes $i\leq j$ and $Pred(z_i,z_j)$ encodes $j=i+1$, so the formulas in \cref{sub:enc_sequences} match \cref{app:ordered_constraint_semantics} independently of tail orders and position-name permutations.
    Each auxiliary count predicate in \cref{tab:size_atom_encoding} is defined by equivalence from already fixed predicates, so its extension is unique and its exponent equals its semantic count.
\end{itemize}
By induction along $v_1,\ldots,v_n$, each rule touches only its fresh vocabulary and that of its dependencies, so after all $n$ units the partial representations map onto all labelled assignments respecting every declaration within the bounds, each with $\overcount$ preimages (or $1$ if no sequence occurs).
Conjoining the constraint rules retains precisely the preimages of assignments satisfying every constraint without altering preimage counts.
Because sound analysis bounds admit every satisfying labelled assignment, the readout image is exactly $\mathsf{Lab}(Q)$.
\end{proof}

\begin{proof}[Proof of \cref{lem:label_forgetting}]
Fix $s\in\mathsf{Sol}(Q)$ and let $r_{g,m}(s)$ count the parts of size $m$ in set partition $g$.
A labelled structure over $s$ is uniquely determined by the slot sequence of each partition (with local part variables and deterministic expressions following the slots).
In a set partition, nonempty parts are pairwise disjoint and hence distinct, so a size-nondecreasing ordering places the $r_{g,m}(s)$ parts of each positive size $m$ in a contiguous slot block in any of $r_{g,m}(s)!$ orders, whereas empty parts are identical and yield a single sequence.
In a bag partition, signature sorting fixes a unique nondecreasing sequence even when parts coincide, and compositions are already ordered.
Hence $s$ admits exactly $\prod_g\prod_{m>0}r_{g,m}(s)!$ selector-respecting slot sequences.
By \cref{prop:part_relabelling} and soundness of the analysis bounds before selectors are imposed, all of these belong to $\mathsf{Lab}(Q)$, so every $s\in\mathsf{Sol}(Q)$ has an admissible labelling.
For any accepted representation $r$ with $\pi(r)=s$, the part-size exponents record the same multiplicities $r_{g,m}(s)$, giving $q_\Sigma(\mathbf d(r))=\prod_g\prod_{m>0}r_{g,m}(s)!$.
\end{proof}

\subsection{Correctness of the Compilation Pipeline}
\label{app:proof_pipeline_correctness}

\begin{proof}[Proof of \cref{thm:compiler_correctness}]
By \cref{lem:analysis_soundness} and \cref{app:sanity_checks}, structural bounds and exact dual LP certificates are sound over-approximations on labelled assignments prior to canonical partition selectors, and returning $\bot$ certifies unsatisfiability.
Each preprocessing rewrite maintains the invariant that every semantic constraint implied by eliminated objects or folded terms remains enforced by surviving definitions or retained constraints.
In particular, constant folding and deterministic sharing replace deterministic expressions with equivalent ones; full-choice elimination replaces a without-replacement choice $v$ of size $k$ from source $u$ only after ensuring that $|u|=k$ is enforced directly by $u$'s surviving definition or by an explicit constraint \code{|u| == k} attached to the final surviving source of any substitution chain; and cardinality folding substitutes $K_w$ into arithmetic constraints only while the definitions or constraints enforcing $|w|=K_w$ remain intact.
Every rewrite therefore induces a bijection on satisfying assignments up to uniquely recoverable deterministic values.
Normalization instantiates part-wise bodies using deterministic auxiliary variables without adding choices (\cref{app:partwise_semantics,app:denotational_semantics}).
Size-range expansion over finite sound upper bounds and Shannon expansion over complete satisfying truth assignments form a disjoint cover of all solutions, with each length-$k$ branch for a full ordered object over a dynamic or derived source $u$ explicitly materializing \code{|u| == k} when $u$'s constructor does not already fix its size; because global Boolean and ordered-length guards are independent of unordered part labels, \cref{prop:part_relabelling} keeps part relabelling inside each branch.
Within each branch, \textbf{split} partitions objects and constraints into disconnected components whose solution space is their Cartesian product, carrying every attached size constraint alongside its source object, and local preprocessing preserves each component's count under the same rewrite invariants while yielding atomic constraints and fixed ordered lengths.
Each surviving component therefore satisfies \cref{prop:component_encoding_correctness}, so \cref{line:solve} returns its exact semantic count, and \cref{alg:compile} computes the sum-of-products formula \eqref{eq:compiler_sum_product}, returning $\comcount{D,O,C}$.
\end{proof}

\subsection{Proofs for Section~\ref{sec:bag_factorization}}
\label{app:proof_factorization}

To support Step~(i) of the proof of \cref{prop:factorization_correctness} in the main text, we establish the representation bijection for factorized classes.
In the factorized encoding, a true grounding $L_G(e)$ with weight $f_G$ carries one term per complete local state $\mathbf q\in\Omega_G$ (labelled by $\mathbf q$ as proof bookkeeping so distinct states with identical monomials remain distinct terms), and readout recovers the set memberships and bag multiplicities of $e$ across $O$ directly from that state label.

\begin{lemma}[Factorized set/bag classes]
    \label{lem:exact_factorization}
    Let $Q$ be a preprocessed set/bag component satisfying the assumptions of \cref{sec:bag_factorization}, and let $G$ be an eligible exchangeable class satisfying \cref{def:bag_factorization_applicability}.
    Writing $\mathrm{ord}$ and $\mathrm{fac}$ for the encodings before and after factorizing $G$, there is a bijection $\Phi:\mathsf{Rep}^{\mathrm{ord}}_{A_{\mathrm{ord}}}(Q)\to\mathsf{Rep}^{\mathrm{fac}}_{A_{\mathrm{fac}}}(Q)$ satisfying $\rho_{\mathrm{fac}}\circ\Phi=\rho_{\mathrm{ord}}$.
    Moreover, $\overcount=1$ and $\Sigma=\emptyset$ ($q_\Sigma\equiv 1$) in both encodings, and the same holds for any collection of pairwise disjoint eligible classes.
\end{lemma}
\begin{proof}
Fix the representation of all entities outside $G$.
Every supported operation in \cref{tab:bag_factor_local_rules} and every positive containment or equality constraint in $C$ relates only coordinates of the same entity, and every $e\in G$ shares the same initialized profile $(\mu_{R,e})_{R\in I}$ and inferred bounds $(\mult_v(e))_{v\in O_+}$.
Before aggregate size filtering, the admissible ordinary support and multiplicity-term choices on $G$ (including singleton variants) therefore correspond bijectively to $\prod_{e\in G}\Omega_G$.
Mapping each entity $e\in G$'s ordinary choices to the term of $L_G(e)$ labelled by its complete local state $\mathbf q\in\Omega_G$, while keeping all predicates and terms outside $G$ unchanged, defines a bijection on representations.
In the factorized instance (\cref{app:lifted_bag_factorization}), closed-world evidence fixes $L_G$ to $G$, unused dynamic support atoms on $G$ are fixed to false, ordinary support relations and positive comparisons are guarded to apply outside $G$, and factorized \code{supp} predicates are separated from initialized source supports so no unassigned or conflicting atoms arise; hence the inverse map uniquely recovers every ordinary atom and term with identical readout ($\rho_{\mathrm{fac}}\circ\Phi=\rho_{\mathrm{ord}}$).

For every tracked object $v\in O_c$, the exponent of $x_{v,G}$ sums $q_v$ over $e\in G$, so combining it with the unchanged contributions outside $G$ preserves every size expression and every named-entity count outside $G$.
Because local operations and positive comparisons on $G$ are enforced inside $\Omega_G$ and outside $G$ by the retained ordinary encoding, $A_{\mathrm{ord}}$ holds if and only if $A_{\mathrm{fac}}$ holds, restricting $\Phi$ to a bijection between accepted representations.
Since $Q$ contains no sequences or set partitions, $\overcount=1$ and $\Sigma=\emptyset$ in both encodings; composing these bijections over pairwise disjoint classes yields the general case.
\end{proof}

\section{Implementation of Bag Factorization}
\label{app:lifted_bag_factorization}

This section details how the \wfomc{} backend implements the exchangeable bag factorization of \cref{sec:bag_factorization} through component analysis and exact propositional weighted counting (\cref{app:bag_factorization_algorithm}).
It also describes the singleton support encoding used to compress eligible $0/1$-bounded bag coordinates that remain in the ordinary representation (\cref{app:singleton_encoding}).

\subsection{Factorization Analysis and Local Counting}
\label{app:bag_factorization_conditions}
\label{app:bag_factorization_algorithm}

Given a preprocessed component, the backend first checks whether it qualifies as an eligible set/bag component under \cref{sub:bag_applicability}.
A component is eligible when every object in it is a set or bag built from initializers and the supported operations of \cref{tab:bag_factor_local_rules}, and every constraint is either a positive local comparison (\code{subset} or \code{==}) or a size constraint over object cardinalities and \code{B.count(e)} atoms (excluding components containing tuples, sequences, groupings, negated comparisons, or membership queries on factorized entities).

Within an eligible component with initialized objects $I$ and dynamic objects $O_+$, each entity $e\in D_I=\bigcup_{R\in I}\support{R}$ with finite non-negative bounds is grouped by its joint initialized profile $\mathbf m_e=(\mu_{R,e})_{R\in I}$ (taking $0$ for initializers omitting $e$, and $1$ for set initializers containing $e$) together with its dynamic bound vector $(\mult_v(e))_{v\in O_+}$ (with dynamic sets bounded by $1$ when $e\in\pent(v)$ and $\maxsize(v)>0$, and $0$ otherwise).
Operands derived from distinct initializers can therefore enter the same local intersection, union, difference, or positive comparison.
Any entity named by a \code{B.count(e)} term with $B\in O$ is excluded from candidate classes and retains its ordinary coordinates throughout the component.
\cref{alg:bag_factorization} accepts each candidate class $G$ with $|G|>1$ that passes a cumulative compilation budget bounding the estimated binary-circuit work plus the product $\prod_{v\in O_c}(\mult_v+1)$ of tracked domain sizes; ineligible components and rejected classes fall back cleanly to the ordinary encoding.

For each accepted class $G$, the backend encodes $\Omega_G$ as a functional propositional CNF circuit.
Initialized coordinates are substituted directly as constant bit vectors $q_R=\mu_{R,e}$ from the common profile.
Each dynamic coordinate with upper bound $u$ uses $0$ variables when $u=0$ and $\lceil\log_2(u+1)\rceil$ Boolean variables otherwise, with an unsigned upper-bound clause $q_v\leq u$ whenever $u+1$ is not a power of two.
Unsigned comparators implement choice, containment, minimum (intersection), and maximum (union); bitwise equalities enforce positive equality; full-width adders retaining the carry bit implement additive union and truncated difference without wraparound; disjunction over source bits implements \code{supp}; and \code{choose_replace} adds clauses requiring each output bit to imply the source set bit.
Because every auxiliary variable is defined functionally, each local state $\mathbf q\in\Omega_G$ corresponds to a unique satisfying assignment.
For each tracked object $v\in O_c$, bit $j$ ($j\geq 0$) of $q_v$ receives positive weight $x_{v,G}^{2^j}$ and negative weight $1$, while all other weights are $1$.
If a comparison between fixed coordinates yields an empty clause, the local factor evaluates immediately to $0$; otherwise exact weighted model counting via symbolic Ganak sums over untracked coordinates and returns $f_G(\mathbf x)$ in \eqref{eq:lifted_bag_factor}.

The following two-initializer example illustrates multi-initializer factorization.
Let \code{A = bag(a: 2, b: 2)} and \code{B = bag(a: 3, b: 3)}, with \code{X = choose(A)}, \code{Y = choose(B)}, \code{Z = X & Y}, and \code{|Z| == 2}.
Both entities share initialized profile $(2,3)$, and only $|Z|$ belongs to $O_c$.
Among the twelve local pairs $(q_X,q_Y)\in\{0,1,2\}\times\{0,1,2,3\}$ for one entity, six have minimum $0$, four have minimum $1$, and two have minimum $2$, giving $f_G(z)=6+4z+2z^2$ and $[z^2]f_G(z)^2=40$ satisfying structures.

To integrate accepted classes into the outer \wfomc{} instance, the encoder runs a preparation pass before ordinary encoding and a finalization pass afterwards (\cref{alg:bag_factorization}).
Preparation introduces a fresh unary predicate $L_G/1$ fixed to $G$ by closed-world evidence and reserves the size symbols $x_{v,G}$ for $v\in O_c$.
Ordinary encoding retains initialized evidence, omits per-entity multiplicity predicates on factorized dynamic bags, gives any factorized \code{supp(B)} its own support predicate, guards ordinary support relations and positive comparisons outside the classes whose local factors already enforce them, and sums the shared symbols $x_{v,G}$ with remaining ordinary contributions in $A$.
Finalization evaluates each local factor $f_G$, sets $\weight(L_G)=f_G$ and $\negweight(L_G)=1$, and adds negative unary evidence fixing unused dynamic support atoms of $G$ in $O_+$ to false so the outer solver does not re-enumerate choices already summed into $f_G$.

\begin{algorithm}[tbp]
    \DontPrintSemicolon
    \SetNoFillComment
    \caption{Encoding a supported set/bag component with exchangeable bag factorization}
    \label{alg:bag_factorization}
    \KwIn{An eligible set/bag component with object set $O$, its analysis record, and the component budget}
    \KwOut{Factorized predicates, local weights, and ordinary encoding contributions for the component}
    $\mathcal{G}\gets\emptyset$\;
    Let $I$ be all initialized objects in $O$ and set $O_+\gets O\setminus I$\;
    \lIf{$I=\emptyset$ or $O_+=\emptyset$}{\Return the ordinary encoding}
    Set $D_I\gets\bigcup_{R\in I}\support{R}$ and group entities with finite non-negative bounds by $((\mu_{R,e})_{R\in I},(\mult_v(e))_{v\in O_+})$\;
    \ForEach{candidate group $G$}{
        Remove from $G$ any entity named by a \code{B.count(e)} term with $B\in O$\;
        \lIf{$|G|\leq 1$}{\textbf{continue}}
        Determine tracked objects $O_c\subseteq O_+$ as in \cref{sub:bag_local_factor} and estimate compilation work\;
        \lIf{the estimated work exceeds the remaining budget}{\textbf{continue}}
        Deduct the estimated work from the remaining budget and add $G$ to $\mathcal{G}$\;
    }
    \lIf{$\mathcal{G}=\emptyset$}{\Return the ordinary encoding}
    For each $G\in\mathcal{G}$, introduce $L_G/1$ fixed to $G$ by closed-world evidence and reserve $x_{v,G}$ for $v\in O_c$\;
    Encode objects with support relations guarded outside factorized classes and sum shared size contributions in $A$\;
    Encode positive comparisons only outside common recorded factors, retaining all other constraints\;
    \ForEach{$G\in\mathcal{G}$}{
        $f_G\gets$ exact weighted model count of the local CNF for $\Omega_G$ with binary bit weights\;
        Assign weights $(f_G,1)$ to $L_G/1$ and fix unused dynamic support atoms of $G$ in $O_+$ to false\;
    }
    \Return the factorized and ordinary encoding contributions\;
\end{algorithm}

\subsection{Singleton Encoding}
\label{app:singleton_encoding}

Within the ordinary bag encoding, any \emph{singleton entity} whose structural multiplicity bound (before constraint-based tightening) is at most $1$ in every bag of the component and that is not named by a \code{B.count(e)} term is tracked purely through support predicates rather than per-entity multiplicity symbols.
Closed-world evidence fixes $Singleton/1$ to this eligible set $H$.

Whenever the cardinality of a dynamic bag $v$ is needed, an auxiliary predicate isolates its remaining ordinarily encoded singleton members,
\begin{equation*}
    \forall t:B_{v,\mathrm{sing}}(t)
        \leftrightarrow\bigl(B_v(t)\land Singleton(t)\bigr),
    \qquad
    \weight(B_{v,\mathrm{sing}})=x_{v,\mathrm{sing}},\quad
    \negweight(B_{v,\mathrm{sing}})=1.
\end{equation*}
Because unused factorized support atoms are fixed to false, the exponent of $x_{v,\mathrm{sing}}$ equals $\sum_{e\in H_v}\mu_{v,e}$, where $H_v\subseteq H$ denotes the singleton entities retained in the ordinary encoding at $v$.

Because supports remain explicit, entity-specific membership tests (such as \code{a in X}) and bag operations act directly on support predicates over $H_v$.
Choice, intersection, and union translate to support implication, conjunction, and disjunction, respectively, and bag difference uses $\forall t\in H_v: B_v(t)\leftrightarrow(B_{u_1}(t)\land\neg B_{u_2}(t))$ when both operands obey the $0/1$ structural bound (additive union retains ordinary multiplicity equations whenever both operands may contain the same entity).
Positive subbag and equality constraints reduce to support implication and equivalence on $H_v$, while their negations count violating singleton entities via an auxiliary predicate.
For grouped bags with source support $S$ and part predicates $P_1,\ldots,P_k$, multiplicity conservation on $H$ reduces to
\begin{equation*}
    \forall x:(Singleton(x)\land S(x))\to
        \biggl(\bigvee_{i=1}^{k}P_i(x)\land\bigwedge_{1\leq i<j\leq k}\neg(P_i(x)\land P_j(x))\biggr).
\end{equation*}
Entities outside $H$, as well as coordinates required by bag-partition symmetry breaking or ordered objects, retain their ordinary multiplicity encoding.

\section{ASP and \textsc{Essence} Encodings}
\label{app:asp_essence_encodings}

The ASP--Clingo and ESSENCE--Conjure baselines in \cref{sec:experiments} translate \ourlang{} programs into models for general-purpose declarative solvers.
Both share parsing, entity analysis, case expansion, and component decomposition with the \wfomc{} pipeline, so each encoder receives an independent component with finite bounds and fixed ordered-object lengths; component counts are multiplied within branches and summed across branches as in \cref{sub:encoding}.
Both baselines count by exhaustive enumeration and apply explicit symmetry breaking to partitions, avoiding decoding quotients at the cost of enumerating every solution.

\subsection{ASP Encoding}
\label{app:asp_encoding}

The ASP encoding generates candidate objects via choice rules and filters invalid structures with integrity constraints, replacing entity and object names with integer identifiers and encoding multiplicities with \code{mult(r,e,k)}.

\begin{example}\label{ex:asp_encoding}
Choosing two of three entities (\code{S = set(a, b, c); T = choose(S, 2)}) compiles to
\begin{backendcode}{CofolaASP}
entity(0..2).
mult(0,E,1) :- entity(E).
1 { mult(1,E,0); mult(1,E,1) } 1 :- entity(E).
:- mult(1,E,1), mult(0,E,0).
:- #sum { K,E : mult(1,E,K) } != 2.
\end{backendcode}
\noindent whose three stable models correspond to the three two-element subsets.
\end{example}

For bags, each object--entity pair chooses one multiplicity up to its inferred bound; without-replacement choices bound it by the source multiplicity, replacement choices require only membership in the source support, and deterministic operations enforce pointwise maximum, minimum, sum, truncated subtraction, or support indicator rules.

For an ordered object \code{r} of fixed length $n$, the encoder introduces positions $0,\ldots,n-1$ and rules
\begin{backendcode}{CofolaASP}
can(r,E) :- entity(E), mult(src,E,N), N > 0.
1 { at(r,P,E) : can(r,E) } 1 :- pos(r,P).
mult(r,E,N) :- entity(E), num(N),
               N = #count { P : at(r,P,E) }.
\end{backendcode}
\noindent requiring the resulting occurrence counts to equal the source multiplicities for full arrangements, not exceed them for without-replacement choices, or satisfy only source support under replacement.

Grouped objects allocate $k$ part objects whose per-entity multiplicities sum to the source multiplicity, allowing empty parts.
Compositions leave part order unrestricted, whereas partitions order parts in non-increasing lexicographic order of their multiplicity vectors over a fixed entity order to break symmetry.
Basic, positional, aggregate, and sequence-pattern constraints translate into integrity constraints over \code{mult} and \code{at} atoms, with \code{next_to} combining both orientations disjunctively at each adjacent pair so overlapping matches count once.

\subsection{\textup{\textsc{Essence}} Encoding}
\label{app:essence_encoding}

The \textsc{Essence} encoding represents objects using native \code{set}, \code{mset}, and \code{sequence} variables bounded by the inferred size and multiplicity limits.

\begin{example}
The program in \cref{ex:asp_encoding} translates to
\begin{backendcode}{CofolaEssence}
letting Entity be domain int(0..2)
find o_0 : set (size 3) of Entity
find o_1 : set (size 2) of Entity
such that
  o_0 = {0, 1, 2},
  o_1 subsetEq o_0
\end{backendcode}
\noindent whose three solutions correspond to the three subsets.
\end{example}

Set operations use native operators, while bag operations constrain \code{freq(r,e)} pointwise using the same rules as in \cref{app:asp_encoding}.
Ordered objects use $1$-based \code{sequence} variables following the same three source-usage modes, and grouped objects use one collection variable per part with multiplicity conservation and lexicographic symmetry breaking for partitions.
The backend invokes \code{conjure solve --number-of-solutions=all}, refining via Conjure and Savile Row and enumerating solutions with Minion.

\section{Supplementary Experimental Analysis}
\label{app:supplementary_experiments}

\cref{app:unsupported} classifies the unmodelled MATH questions. \cref{app:benchmark_parameters,app:benchmark_diagnostics} describe the benchmarks and report supplementary results. \cref{app:factor_applicability_inventory} gives the bag-factorization measurements and scaling references.

\subsection{Analysis of Unmodelled Questions}
\label{app:unsupported}

\cref{tab:unmodelled_categories} classifies the $145$ candidate MATH questions not modelled in \ourlang{} into eight mutually exclusive primary categories.
Arithmetic over numerical attributes ($89$) and geometric symmetry groups ($33$) together account for $84.1\%$ of all unmodelled questions; three representative examples below illustrate the dominant missing primitives.

\begin{table}[tbp]
\centering
\small
\caption{Primary categories of the $145$ MATH candidate questions not directly modelled in \ourlang{}.}
\label{tab:unmodelled_categories}
\setlength{\tabcolsep}{5pt}
\renewcommand{\arraystretch}{1.12}
\begin{tabularx}{\linewidth}{@{}>{\raggedright\arraybackslash}p{0.34\linewidth}r>{\raggedright\arraybackslash}X@{}}
\toprule
Category & Count & Missing language primitive \\
\midrule
Arithmetic and numerical outcomes & 89 & Numerical entity attributes (sums, products, divisibility) or counting distinct numerical scores \\
Circular and geometric symmetries & 33 & Quotienting arrangements under rotations or reflections (necklaces, polyhedra) \\
Windows, ordered values, and runs & 9 & Sliding-window constraints, monotone value comparisons, or maximal-run parity \\
Palindromes and position equality & 7 & Direct equality comparison between two tuple positions \\
Grid and spatial constraints & 4 & Two-dimensional row, column, or lattice-path relations \\
Comparison with a reference order & 1 & Counting position matches or swap distance against a reference tuple \\
Joint attribute conditions & 1 & Simultaneous all-equal or all-different conditions across multiple attributes \\
History-dependent stack operations & 1 & Counting distinct outputs reachable via interleaved push--pop histories \\
\midrule
Total & 145 & \\
\bottomrule
\end{tabularx}
\end{table}

\begin{problem}\label{prob:unmodelled_winning_scores}
In a cross-country meet between two teams of five runners each, a runner who finishes in the $n$th position contributes $n$ to their team's score, and the team with the lower total score wins. If there are no ties, how many different winning scores are possible?
\end{problem}
\noindent Here the target is the number of distinct attainable numerical scores rather than finishing orders; \ourlang{} treats entities as names without numerical attributes or projection onto derived numerical outcomes.

\begin{problem}\label{prob:unmodelled_cube_beads}
How many ways are there to put eight beads of different colours on the vertices of a cube, if rotations of the cube, but not reflections, are considered the same?
\end{problem}
\noindent Identifying arrangements under an arbitrary geometric symmetry group lies outside the built-in symmetry conventions of sets, bags, and partitions.

\begin{problem}\label{prob:unmodelled_palindromes}
Four-digit integers are formed using the digits $2,3,4,5$, and each digit may be used any number of times. How many such integers are palindromes, reading the same forwards and backwards?
\end{problem}
\noindent Although \ourlang{} tests \code{T[i] == e} against named entities, it lacks direct position-to-position equality tests such as \code{T[0] == T[3]}.
Because our modelling study counts only direct models using native primitives, all seven questions in the ``Palindromes and position equality'' row of \cref{tab:unmodelled_categories} are excluded from the $247$ real models.
Nevertheless, each of these seven questions admits an indirect count-preserving encoding via auxiliary singleton sets.
In \cref{prob:unmodelled_palindromes}, for instance, declaring a length-four replacement tuple \code{T} together with two singletons \code{X = choose(digits, 1)} and \code{Y = choose(digits, 1)} and requiring \code{T[0] in X}, \code{T[3] in X}, \code{T[1] in Y}, and \code{T[2] in Y} uniquely determines \code{X} and \code{Y} for every palindrome, preserving the exact count $16$.
Across all seven questions, such singleton-witness encodings return the exact MATH answer under ASP--Clingo ($7/7$), whereas ordinary \ourlang{}-\wfomc{} solves $3/7$ within the $100$-second limit and times out on the remaining $4/7$ because the auxiliary singletons and tuple-position constraints enlarge the compiled instance.
This confirms both that direct coverage ($63.0\%$) is a lower bound on expressiveness and that native position-equality primitives remain valuable for lifted solving.

\subsection{Benchmark Parameters}
\label{app:benchmark_parameters}

\cref{tab:growing_parameters} specifies the eight growing families, where $n\in\{5,10,\ldots,100\}$ denotes the cardinality of the base set or total size of the base bag (including repeated copies).

\begin{table}[tbp]
\centering
\small
\setlength{\tabcolsep}{4pt}
\renewcommand{\arraystretch}{1.14}
\caption{Exact parameterizations of the eight growing families.}
\label{tab:growing_parameters}
\begin{tabular}{@{}p{0.20\linewidth}p{0.73\linewidth}@{}}
\toprule
Family & Construction at base size $n$ \\
\midrule
TVs & $d=\max(2,\lfloor n/4\rfloor)$ defective TVs and $n-d$ working TVs; choose five with at least two defective. \\
Workers & Three labelled groups of sizes $\lfloor n/2\rfloor$, $\lfloor n/3\rfloor$, and their remainder. \\
BANANA & Multiplicities $\max(1,\lfloor n/6\rfloor)$ for B, $\max(1,\lfloor n/2\rfloor)$ for A, and the remainder for N; permute the full bag. \\
MATHCOUNTS & $\lfloor n/3\rfloor$ distinct vowels, two copies of T, and $n-\lfloor n/3\rfloor-2$ other distinct consonants; choose four with at most one vowel. \\
Four-letter & Four distinct letters from $n$; fix the first to one named letter and require a second named letter to occur. \\
Nested choice & Choose four of $n$ people, then one captain from that team; $4\binom n4$ structures. \\
Selected pairs & Choose four of $n$ entities and partition them into two unordered pairs; $3\binom n4$ structures. \\
Selected sequence & Choose and order five of $n$ entities, including two specified adjacent entities; $2\cdot4!\binom{n-2}{3}$ structures. \\
\bottomrule
\end{tabular}
\end{table}

The synthetic benchmark uses six base sizes, $n\in\{4,8,12,16,24,32\}$. At each size, the eight families in \cref{tab:synthetic_parameters} contribute three SAT/UNSAT pairs each, giving $144$ pairs in total.
For sets, $n$ is the number of entities in the base set. For bags, $n$ counts distinct entity types, with the stock of each type specified separately.
Each SAT program admits at least two solutions. Its UNSAT companion changes one statement.
We checked the reference counts independently using combinatorial formulas, enumeration, or integer dynamic programs for the corresponding family.

\begin{table}[tbp]
\centering
\small
\setlength{\tabcolsep}{4pt}
\renewcommand{\arraystretch}{1.10}
\caption{Construction of the paired synthetic programs. Each family contains $18$ SAT/UNSAT pairs. The final column specifies the one change that makes each companion unsatisfiable.}
\label{tab:synthetic_parameters}
\begin{tabularx}{\linewidth}{@{}>{\raggedright\arraybackslash}p{.20\linewidth}>{\raggedright\arraybackslash}X>{\raggedright\arraybackslash}X@{}}
\toprule
Family & SAT construction & Change in the UNSAT companion \\
\midrule
\multicolumn{3}{@{}l}{\emph{A. One object drawn from a fixed source}}\\
Set choice & Choose a subset with quotas on three pairwise unions of disjoint colour classes. & Increase one quota by one. Their sum must be even since each selected member of the three classes contributes twice. \\
Bag choice & Choose a subbag of fixed size with a quota on one class of types. & Reduce the quota by one, so the demand on the other class exceeds its stock. \\
Ordered selection & Choose an injective tuple with quotas on two complementary classes. & Increase one quota by one, so the two quotas exceed the tuple length. \\
Set partition & Partition a set into a fixed number of unordered blocks of a given minimum size. & Raise the minimum until the blocks require more elements than the set contains. \\
\midrule
\multicolumn{3}{@{}l}{\emph{B. Joint choices or objects drawn from a previous choice}}\\
Disjoint committees & Choose a team and two labelled disjoint committees within it. & Enlarge the second committee by one member, so the committees exceed the team size. \\
Joint bag selection & Choose two subbags from different stocks with a prescribed intersection size. & Reduce the intersection by one, so their union exceeds the total of the per-type stock maxima. \\
Dependent order & Choose a team and order its members subject to precedences among specified members. & Replace one precedence to close a directed cycle. \\
Dependent partition & Choose a subset and partition it into unordered blocks of a given minimum size. & Raise the minimum until the blocks require more elements than were chosen. \\
\bottomrule
\end{tabularx}
\end{table}

Both bag families use the same stock and quota rules at all six sizes.
The set and order families round size proportions at $n=12$ and use explicit small configurations at $n=4$.
The partition families adjust the number of blocks or the size bounds where necessary. For example, the third dependent-partition configuration at $n=8$ selects four entities and divides them into two nonempty blocks.
The artifact records the parameter values and exact reference count for every program.

\begin{samepage}
In one joint bag selection at $n=4$, the two stocks have multiplicities $(3,3,2,2)$ and $(2,2,3,3)$, and each chosen bag must contain eight copies.
An intersection of size four forces their union to attain its maximum size of twelve, with three copies of each type.
Each bag must take all six copies of its two high-stock types. It can distribute its remaining two copies across the other two types in three ways, giving $3^2=9$ joint selections.
Changing the intersection size to three would require a union of size thirteen.\par
\end{samepage}

\subsection{Supplementary Benchmark Results}
\label{app:benchmark_diagnostics}

For set choices and ordered selections in group A, we verified that the CoLa and \ourlang{} constraints agree on every feasible vector of colour-class counts.
For bag choices we verified the per-type stocks, the choice size, and the marked-class quota, and for partitions we verified the source set, the number of unordered blocks, and their common minimum size.
These checks cover $72$ SAT and $70$ UNSAT models, while the remaining two UNSAT programs reduced to zero during preprocessing.
In every timed run on these $142$ models, we confirmed that the input passed to the official CoSo solver matched the verified model byte for byte.

To complement \cref{fig:matched_inference_costs} on non-sequence structures, \cref{tab:matched_nonseq_inference} compares lifted and propositional inference (Ganak) on identical ordinary compiled inputs (with bag factorization disabled) across all $20$ scales ($n=5,10,\ldots,100$) of Nested choice (sets), MATHCOUNTS (bags), and Selected pairs (set partitions).
Both methods solve all $60$ instances across three repetitions each ($360$ exact runs), with propositional inference faster on $46/60$ smaller instances and flat lifted inference overtaking it by $n=100$ on all three families.

\begin{table}[tbp]
\centering\small
\setlength{\tabcolsep}{4pt}
\renewcommand{\arraystretch}{1.12}
\caption{Matched-encoding inference comparison on three non-sequence growing families ($20$ scales each, $n=5,\ldots,100$, factorization disabled). Ratio is the geometric mean of propositional-to-lifted inference time (values below $1$ favour propositional inference); median inference times (in seconds) are shown as lifted / propositional at $n=5$, $50$, and $100$.}
\label{tab:matched_nonseq_inference}
\begin{tabular}{@{}lrrrrrrr@{}}\toprule
Family & Solved & Ratio & \makecell{Prop\\faster} & \makecell{CNF at $n=100$\\(vars / clauses)} & $n=5$ & $n=50$ & $n=100$ \\ \midrule
Nested choice & 20/20 & 0.597 & 16/20 & 501 / 902 & 0.117 / 0.024 & 0.121 / 0.073 & 0.124 / 0.170 \\
MATHCOUNTS & 20/20 & 0.492 & 18/20 & 1{,}783 / 3{,}962 & 0.128 / 0.019 & 0.131 / 0.066 & 0.131 / 0.153 \\
Selected pairs & 20/20 & 0.789 & 12/20 & 901 / 1{,}902 & 0.119 / 0.029 & 0.121 / 0.098 & 0.124 / 0.233 \\
\bottomrule\end{tabular}
\end{table}

\subsection{Bag Factorization Measurements}
\label{app:factor_applicability_inventory}

\cref{tab:factor_applicability_inventory} summarizes the applicability of bag factorization to the $695$ programs in the main benchmarks; structural exclusions occur when a component contains ordered objects, groupings, negated comparisons, or constraints outside \cref{def:bag_factorization_applicability}.
On the synthetic programs, the three runs agreed on which classes were selected for every input. Two UNSAT set-choice programs reduced to zero before compilation.
On the $60$ selected real MATH programs, median three-run runtimes were $1.128$ seconds without factorization and $1.145$ seconds with factorization. The geometric mean of the per-program off/on runtime ratios was $0.986$, with median $0.983$ and interquartile range $0.964$--$0.999$. Peak resident memory was $104.07$ and $104.13$\,MiB, respectively.

\begin{table}[tbp]\centering\small
\setlength{\tabcolsep}{5pt}
\renewcommand{\arraystretch}{1.12}
\caption{Factorization analysis outcomes by program in the main evaluation runs. The selected column counts programs with at least one exchangeable class; every selected program ($186/186$) is an eligible set/bag component within the scope of \cref{prop:factorization_correctness}.}
\label{tab:factor_applicability_inventory}
\begin{tabular}{@{}lrrrr@{}}\toprule
Population & Inputs & Selected & Structural exclusion & No candidate \\ \midrule
Real & 247 & 60 & 70 & 117 \\
Synthetic SAT & 144 & 54 & 52 & 38 \\
Synthetic UNSAT & 144 & 52 & 52 & 40 \\
Growing & 160 & 20 & 79 & 61 \\
All inputs & 695 & 186 & 253 & 256 \\
\bottomrule\end{tabular}\end{table}

To construct the type-scaling study of \cref{sub:experiments_bag_lifting}, we inspected all $52$ MATH programs that initialized a bag or selected a bag with replacement.
Five yielded an exchangeable bag factor at original scale (IDs 21, 70, 132, 165, and 188 in \cref{tab:bag_scaling_references}), and three more did so once entity types were doubled (IDs 2, 93, and 234, which stock only a single type per category at original scale); the only other selected bag-bearing program (ID 275) factorized a separate set choice rather than its bag object and was therefore excluded from bag-type scaling.
For all eight sources, multiplying the number of distinct types by $k\in\{2,4,8,16\}$ while keeping per-type stocks and selection rules fixed produced $32$ expanded instances alongside the eight originals.
\cref{tab:bag_scaling_references} lists their independent reference counts, where $F(a,b,q)=[x^q](1+x)^a(1+x+x^2)^b$ counts selections of $q$ magnets from $a$ single-copy and $b$ double-copy types, and in ID 2 tier $j$ uses denominations $(1,5,25)\,64^j$ so that base-$64$ uniqueness makes all $24^k-1$ nonempty selections yield distinct sums.
Across all $40$ programs ($120$ runs per setting), factorization solved all $120$ runs, whereas the ordinary encoding solved $63$ runs, timed out on $12$, and ran out of memory on $45$.

\begin{table}[tbp]\centering\small
\caption{Eight MATH sources and their reference counts at type multiplier $k$. Copies of the same type are indistinguishable.}
\label{tab:bag_scaling_references}
\setlength{\tabcolsep}{4pt}
\renewcommand{\arraystretch}{1.08}
\begin{tabularx}{\linewidth}{Xl@{}}
\toprule
Selection at scale $k$ & Reference count \\
\midrule
 Four scoops from $3k$ ice cream flavours, with replacement. & $\binom{3k+3}{4}$ \\
Six cookies from $3k$ types, with replacement. & $\binom{3k+5}{6}$ \\
Two vowels and four consonants with the stocks in \textsc{Mathematics} replicated $k$ times across distinct types. & $F(2k,k,2)\,F(3k,2k,4)$ \\
 Two marbles from $3k$ colours with one marble and $k$ colours with three. & $\binom{4k}{2}+k$ \\
Three items in total. One buyer may take each of $5k$ cookie and $3k$ milk types once; the other may repeat cookie types. & $[x^3](1+x)^{8k}(1-x)^{-5k}$ \\
A nonzero payment from $k$ denomination tiers with stocks $(3,2,1)$ per tier. & $24^k-1$ \\
A nonempty, proper subbag of a library stocking six copies of each of $k$ titles. & $7^k-2$ \\
A nonempty basket from $k$ apple types (five each) and $k$ orange types (ten each). & $66^k-1$ \\
\bottomrule\end{tabularx}\end{table}

\cref{tab:joint_bag_diagnostics} details the staged failure diagnostics for the synthetic joint bag-selection family ($36$ programs, $108$ factorized runs) discussed in \cref{sub:experiments_synthetic}.
Every run constructs two exchangeable class factors (each of size $n/2$, tracking $|X|$, $|Y|$, and $|Z|$, with $12$, $20$, or $30$ nonzero local monomials across the three stock configurations) in about $0.05$ seconds, holding the outer symbol count fixed at six.
In an unfactorized control on the lowest-stock SAT configuration, the ordinary encoding already expands to $12$ symbols and $104{,}976$ monomials at $n=4$ (compared with $6$ symbols and $2{,}500$ monomials with factorization) and exhausts the $32$\,GiB memory limit at $n=8$ and $12$, where factorization solves all six programs.
At $n\geq 16$, however, the six-variable outer polynomial itself reaches tens to hundreds of millions of nonzero monomials, causing $29$ decode timeouts, $1$ decode memory exhaustion, and $18$ outer \wfomc{} memory exhaustions across the $48$ failed runs.

\begin{table}[tbp]\centering\small
\setlength{\tabcolsep}{4pt}
\renewcommand{\arraystretch}{1.12}
\caption{Staged diagnostics on the $36$ synthetic joint bag-selection programs ($108$ runs with factorization enabled). Medians are aggregated first over the three repetitions per program and then across programs at each size, counting time until termination for runs hitting a limit within a stage; outer monomial ranges cover programs whose outer \wfomc{} call returned a polynomial. Failed runs list decoding (dec) and outer \wfomc{} (out) failures.}
\label{tab:joint_bag_diagnostics}
\begin{tabular}{@{}rrrrrrrrr@{}}\toprule
$n$ & Solved & \makecell{Outer\\poly} & \makecell{Outer nonzero\\monomials} & \makecell{Local\\build (s)} & \makecell{Outer\\infer (s)} & \makecell{Decode\\(s)} & \makecell{Peak RSS\\(MiB)} & \makecell{Failed runs\\(of $18$)} \\ \midrule
4 & 6/6 & 6/6 & $2.5\times 10^3$--$3.6\times 10^4$ & 0.048 & 0.119 & 0.078 & 107 & 0 \\
8 & 6/6 & 6/6 & $7.3\times 10^4$--$1.6\times 10^6$ & 0.048 & 0.171 & 2.259 & 251 & 0 \\
12 & 6/6 & 6/6 & $6.3\times 10^5$--$1.6\times 10^7$ & 0.047 & 0.590 & 21.086 & 1{,}461 & 0 \\
16 & 2/6 & 6/6 & $3.0\times 10^6$--$8.6\times 10^7$ & 0.049 & 2.541 & 88.599 & 7{,}118 & 12 dec \\
24 & 0/6 & 4/6 & $3.0\times 10^7$--$2.1\times 10^8$ & 0.048 & 36.165 & 79.731 & 29{,}039 & 12 dec, 6 out \\
32 & 0/6 & 2/6 & $1.5\times 10^8$ & 0.050 & 91.392 & 67.517 & 32{,}053 & 6 dec, 12 out \\
\bottomrule\end{tabular}\end{table}

\end{document}